%% file: main.tex
\pdfoutput=1

\documentclass{article}
\usepackage{template_files/jfrExamplee}
\usepackage{graphicx}
\usepackage{setspace}
\usepackage{amsmath}
\usepackage{amssymb}
\usepackage[dvipsnames]{xcolor}
\usepackage{colortbl}
\usepackage{multirow}
\usepackage{paralist}
\usepackage{enumitem}
\usepackage{hyperref}
\usepackage{comment}
\hypersetup{
    colorlinks = true,
    allcolors = {blue},
}

\usepackage{subcaption}
\usepackage{algorithm,algorithmic}
\usepackage{titlesec}

\usepackage{tocloft}

\usepackage[framemethod=tikz]{mdframed}
\usepackage{lipsum}

\definecolor{mycolor}{rgb}{0.122, 0.435, 0.698}

\newmdenv[innerlinewidth=0.5pt, roundcorner=4pt,linecolor=mycolor,innerleftmargin=6pt,
innerrightmargin=6pt,innertopmargin=6pt,innerbottommargin=6pt]{mybox}

\usepackage{booktabs}
\usepackage{tikzscale}
\usepackage{pgfplots}
\pgfplotsset{compat=1.16}
\usepackage{tikz}

\usepackage[smaller,nohyperlinks]{acronym}

\newcommand{\ph}[1]{{\textbf{#1}:}}
\newcommand{\pr}[1]{\textbf{#1:}} %

\newcommand{\rev}[1]{#1}
\newcommand{\revv}[1]{#1}

\newcommand{\argmax}{\mathop{\mathrm{argmax}}}

\graphicspath{{../}{./}{./figures/}}

\title{\large An Addendum to NeBula: \\
Towards Extending TEAM CoSTAR's Solution to Larger Scale Environments}

\author{
\small \normalfont
    \parbox{\linewidth}{\centering
Ali Agha$^*$\textsuperscript{1},
Kyohei Otsu$^*$\textsuperscript{1}, 
Benjamin Morrell$^*$\textsuperscript{1}, 
David D. Fan\textsuperscript{1}, 
Sung-Kyun Kim\textsuperscript{1}, 
Muhammad Fadhil Ginting\textsuperscript{1},
Xianmei Lei\textsuperscript{1}, 
Jeffrey Edlund\textsuperscript{1},
Seyed Fakoorian\textsuperscript{1},
Amanda Bouman\textsuperscript{2}, 
Fernando Chavez\textsuperscript{1}, 
Taeyeon Kim\textsuperscript{1},
Gustavo J. Correa\textsuperscript{1},
Maira Saboia\textsuperscript{1},
Angel Santamaria-Navarro\textsuperscript{3},
Brett Lopez\textsuperscript{1}, 
Boseong Kim\textsuperscript{4},
Chanyoung Jung\textsuperscript{4},
Mamoru Sobue\textsuperscript{1},
Oriana Claudia Peltzer\textsuperscript{6},
Joshua Ott\textsuperscript{6},
Robert Trybula\textsuperscript{1},
Thomas Touma\textsuperscript{2},
Marcel Kaufmann\textsuperscript{1},
Tiago Stegun Vaquero\textsuperscript{1}, 
Torkom Pailevanian\textsuperscript{1},
Matteo Palieri\textsuperscript{10},
Yun Chang\textsuperscript{5},
Andrzej Reinke\textsuperscript{1},
Matthew Anderson\textsuperscript{2},
Frederik E.T. Schöller\textsuperscript{1},
Patrick Spieler\textsuperscript{1},
Lillian M. Clark\textsuperscript{1},
Avak Archanian\textsuperscript{1},
Kenny Chen\textsuperscript{1},
Hovhannes Melikyan\textsuperscript{1},
Anushri Dixit\textsuperscript{2},
Harrison Delecki\textsuperscript{6},
Daniel Pastor\textsuperscript{2},
Barry Ridge\textsuperscript{1},
Nicolas Marchal\textsuperscript{1},
Jose Uribe\textsuperscript{1},
Sharmita Dey\textsuperscript{1},
Kamak Ebadi\textsuperscript{1},
Kyle Coble\textsuperscript{1},
Alexander Nikitas Dimopoulos\textsuperscript{1},
Vivek Thangavelu\textsuperscript{11},
Vivek S. Varadharajan\textsuperscript{9},
Nicholas Palomo\textsuperscript{1},
Antoni Rosinol\textsuperscript{5},
Arghya Chatterjee\textsuperscript{1},
Christoforos Kanellakis\textsuperscript{8},
Bjorn Lindqvist\textsuperscript{8},
Micah Corah\textsuperscript{1},
Kyle Strickland\textsuperscript{7},
Ryan Stonebraker\textsuperscript{1},
Michael Milano\textsuperscript{1},
Christopher E. Denniston\textsuperscript{1},
Sami Sahnoune\textsuperscript{1},
Thomas Claudet\textsuperscript{1},
Seungwook Lee\textsuperscript{4},
Gautam Salhotra\textsuperscript{1},
Edward Terry\textsuperscript{1},
Rithvik Musuku\textsuperscript{2},
Robin Schmid\textsuperscript{1},
Tony Tran\textsuperscript{1},
Ara Kourchians\textsuperscript{1},
Justin Schachter\textsuperscript{1},
Hector Azpurua\textsuperscript{1},
Levi Resende\textsuperscript{1},
Arash Kalantari\textsuperscript{1},
Jeremy Nash\textsuperscript{1},
Josh Lee\textsuperscript{2},
Christopher Patterson\textsuperscript{12},
Jennifer G. Blank\textsuperscript{13},
Kartik Patath\textsuperscript{1},
Yuki Kubo\textsuperscript{1},
Ryan Alimo\textsuperscript{1},
Yasin Almalioglu\textsuperscript{1},
Aaron Curtis\textsuperscript{1},
Jacqueline Sly\textsuperscript{1},
Tesla Wells\textsuperscript{1},
Nhut T. Ho\textsuperscript{7},
Mykel Kochenderfer\textsuperscript{6},
Giovanni Beltrame\textsuperscript{9},
George Nikolakopoulos\textsuperscript{8}, 
David Shim\textsuperscript{9},
Luca Carlone\textsuperscript{5}, 
Joel Burdick\textsuperscript{2} %
    }%
  \\\\
\textsuperscript{1} NASA Jet Propulsion Laboratory, California Institute of Technology\\
\textsuperscript{2} California Institute of Technology\\
\textsuperscript{3} Institut de Robòtica i Informàtica Industrial, CSIC-UPC\\
\textsuperscript{4} Korea Advanced Institute of Science and Technology \\
\textsuperscript{5} Massachusetts Institute of Technology\\
\textsuperscript{6} Stanford University \\
\textsuperscript{7} California State University, Northridge \\
\textsuperscript{8} Luleå University of Technology \\
\textsuperscript{9} École Polytechnique de Montréal \\
\textsuperscript{10} Polytechnic University of Bari \\
\textsuperscript{11} Cornell University \\
\textsuperscript{12} McGill University \\
\textsuperscript{13} Blue Marble Space Institute of Science \\
}

\begin{document}
\include{mathsym}

\maketitle
\begin{abstract}
This paper presents an appendix to the original NeBula autonomy solution \cite{agha2021nebula} developed by the TEAM CoSTAR (Collaborative SubTerranean Autonomous Robots), participating in the DARPA Subterranean Challenge. Specifically, this paper presents extensions to NeBula's hardware, software, and algorithmic components that focus on increasing the range and scale of the exploration environment. From the algorithmic perspective, we discuss the following extensions to the original NeBula framework: (i) large-scale geometric and semantic environment mapping; (ii) an adaptive positioning system; (iii) probabilistic traversability analysis and local planning; (iv) large-scale POMDP-based global motion planning and exploration behavior; (v) large-scale networking and decentralized reasoning; (vi) communication-aware mission planning; and (vii) multi-modal ground-aerial exploration solutions. We demonstrate the application and deployment of the presented systems and solutions in various large-scale underground environments, including limestone mine exploration scenarios as well as deployment in the DARPA Subterranean challenge.

\end{abstract}

\newpage
\tableofcontents

\input{sections/1.introduction}

\input{sections/2.estimation}
\input{sections/3.slam}

\input{sections/4.artifacts}

\input{sections/5.motion_planning}

\input{sections/6.global_planning}
\input{sections/7.comms}
\input{sections/8.mission_planning}

\input{sections/9.hardware}

\input{sections/10.drone_hw_and_sw}

\input{sections/11.field}
\input{sections/12.conclusions}

\subsubsection*{Acknowledgments}
The work is partially supported by the Jet Propulsion Laboratory, California Institute of Technology, under a contract with the National Aeronautics and Space Administration (80NM0018D0004), and Defense Advanced Research Projects Agency (DARPA). Further, it was partially supported by the Spanish Ministry of Science and Innovation under the project AUDEL (TED2021-131759A-I00, funded by MCIN/ AEI /10.13039/501100011033 and by the "European Union NextGenerationEU/PRTR"), and by the Consolidated Research Group RAIG (2021 SGR 00510) of the Departament de Recerca i Universitats de la Generalitat de Catalunya.
\footnote{© 2022. All rights reserved.}

We also would like to thank our collaborators and sponsors including Boston Dynamics, JPL office of strategic investments, Intel, Clearpath Robotics, Velodyne, Telerob USA, Markforged, Silvus Technologies, ARCS California State University Northridge, Vectornav, Redara Labs, ARCH Mine, Mine Safety and Health Administration (MSHA), National Institute for Occupational Safety and Health, Polytechnique Montreal, and West Virginia University. 

We would like to thank the rest of Team CoSTAR, our collaborators, and advisors for their support, fruitful discussions and their contributions to the project. In particular, we would like to thank Rohan Thakker, Michael Wolf, Andrea Tagliabue, Jesus Torres, Hanseob Lee, Henry Leopold, Gene Merewether, Jairo Maldonado-Contreras, Sunggoo Jung, Leon Kim, Eric Heiden, Thomas Lew, Abhishek Cauligi, Tristan Heywood, Andrew Kramer, Gustavo Pessin, Nils Napp, Giulio Autelitano, Nicholas Ohi, Sahand Sabet, Abhishek Thakur, Kevin Lu, Chuck Bergh, Sandro Berchier, Meriem Miled, Joseph Bowkett, Micah Feras, Navid Nasiri, Tomoki Emmei, Rianna M. Jitosho, Cagri Kilic, Gita Temelkova, Leonardo Forero, Daniel Tikhomirov, Olivier Toupet, Aaron Ames, Issa Nesnas, Fernando Mier-hicks, Shreyansh Daftry, Peng Nicholas, Jack Bush, Terry Suh, Mike Paton, Jared Beard, Jennifer Nguyen, Michelle Tan, Andrew J Haeffner, Pradyumna Vyshnav, Marcus Abate, Alexandra Stutt, Brett Kennedy, Larry Matthies, Jeff Hall, Joe Bautista, Paul Backes, Marco Tempest, Richard French, Andila Wijekulasuriya, Kayla Mesina, Erica Bettencourt, Amiel Gitai Hartman, Jeff Delaune, Darmindra Arumugam, Jacopo Villa, Soojean Han, David Chan, Andrew Bieler, Slawek Kurdziel, Steve Zhao, Reza Radmanesh, Jose Mendez, Roberto Mendez, Jack Dunkle, Jong Tai Jang, Petter Nilson, Filip Claeson, Emil Fresk, Ransalu Senanayake, Carmen, Mohammad Javad Khojasteh.

The content is not endorsed by and does not necessarily reflect the position or policy of the government or our sponsors.

\input{sections/A.acronym_glossary}

\bibliographystyle{template_files/apalike}
\bibliography{main}

\end{document}

%% file: mathsym.tex

\def\tr{^{\rm T}}
\def\trd{{}^{\rm T}}
\def\Atan{\mathrm{Atan2}}
\def\Acos{\mathrm{Acos}}
\def\sgn{\mathrm{sgn}}
\def\de{\mathrm{d}}
\def\diag{\mathrm{diag}}

\def\zero{\hbox{\bf 0}}
\def\Zero{{\mbox{\boldmath $O$}}}

\def\bfa{{\mbox{\boldmath $a$}}}
\def\bfb{{\mbox{\boldmath $b$}}}
\def\bfc{{\mbox{\boldmath $c$}}}
\def\bfd{{\mbox{\boldmath $d$}}}
\def\bfe{{\mbox{\boldmath $e$}}}
\def\bff{{\mbox{\boldmath $f$}}}
\def\bfg{{\mbox{\boldmath $g$}}}
\def\bfh{{\mbox{\boldmath $h$}}}
\def\bfi{{\mbox{\boldmath $i$}}}
\def\bfj{{\mbox{\boldmath $j$}}}
\def\bfk{{\mbox{\boldmath $k$}}}
\def\bfl{{\mbox{\boldmath $l$}}}
\def\bfm{{\mbox{\boldmath $m$}}}
\def\bfn{{\mbox{\boldmath $n$}}}
\def\bfo{{\mbox{\boldmath $o$}}}
\def\bfp{{\mbox{\boldmath $p$}}}
\def\bfq{{\mbox{\boldmath $q$}}}
\def\bfr{{\mbox{\boldmath $r$}}}
\def\bfs{{\mbox{\boldmath $s$}}}
\def\bft{{\mbox{\boldmath $t$}}}
\def\bfu{{\mbox{\boldmath $u$}}}
\def\bfv{{\mbox{\boldmath $v$}}}
\def\bfw{{\mbox{\boldmath $w$}}}
\def\bfx{{\mbox{\boldmath $x$}}}
\def\bfy{{\mbox{\boldmath $y$}}}
\def\bfz{{\mbox{\boldmath $z$}}}
\def\bfA{{\mbox{\boldmath $A$}}}
\def\bfB{{\mbox{\boldmath $B$}}}
\def\bfC{{\mbox{\boldmath $C$}}}
\def\bfD{{\mbox{\boldmath $D$}}}
\def\bfE{{\mbox{\boldmath $E$}}}
\def\bfF{{\mbox{\boldmath $F$}}}
\def\bfG{{\mbox{\boldmath $G$}}}
\def\bfH{{\mbox{\boldmath $H$}}}
\def\bfI{{\mbox{\boldmath $I$}}}
\def\bfJ{{\mbox{\boldmath $J$}}}
\def\bfK{{\mbox{\boldmath $K$}}}
\def\bfL{{\mbox{\boldmath $L$}}}
\def\bfM{{\mbox{\boldmath $M$}}}
\def\bfN{{\mbox{\boldmath $N$}}}
\def\bfO{{\mbox{\boldmath $O$}}}
\def\bfP{{\mbox{\boldmath $P$}}}
\def\bfQ{{\mbox{\boldmath $Q$}}}
\def\bfR{{\mbox{\boldmath $R$}}}
\def\bfS{{\mbox{\boldmath $S$}}}
\def\bfT{{\mbox{\boldmath $T$}}}
\def\bfU{{\mbox{\boldmath $U$}}}
\def\bfV{{\mbox{\boldmath $V$}}}
\def\bfW{{\mbox{\boldmath $W$}}}
\def\bfX{{\mbox{\boldmath $X$}}}
\def\bfY{{\mbox{\boldmath $Y$}}}
\def\bfZ{{\mbox{\boldmath $Z$}}}

\def\bfGamma{{\mbox{\boldmath $\Gamma$}}}
\def\bfDelta{{\mbox{\boldmath $\Delta$}}}
\def\bfTheta{{\mbox{\boldmath $\Theta$}}}
\def\bfLambda{{\mbox{\boldmath $\Lambda$}}}
\def\bfXi{{\mbox{\boldmath $\Xi$}}}
\def\bfPi{{\mbox{\boldmath $\Pi$}}}
\def\bfSigma{{\mbox{\boldmath $\Sigma$}}}
\def\bfUpsilon{{\mbox{\boldmath $\Upsilon$}}}
\def\bfPhi{{\mbox{\boldmath $\Phi$}}}
\def\bfPsi{{\mbox{\boldmath $\Psi$}}}
\def\bfOmega{{\mbox{\boldmath $\Omega$}}}
\def\bfalpha{{\mbox{\boldmath $\alpha$}}}
\def\bfbeta{{\mbox{\boldmath $\beta$}}}
\def\bfgamma{{\mbox{\boldmath $\gamma$}}}
\def\bfdelta{{\mbox{\boldmath $\delta$}}}
\def\bfepsilon{{\mbox{\boldmath $\epsilon$}}}
\def\bfzeta{{\mbox{\boldmath $\zeta$}}}
\def\bfeta{{\mbox{\boldmath $\eta$}}}
\def\bftheta{{\mbox{\boldmath $\theta$}}}
\def\bfiota{{\mbox{\boldmath $\iota$}}}
\def\bfkappa{{\mbox{\boldmath $\kappa$}}}
\def\bflambda{{\mbox{\boldmath $\lambda$}}}
\def\bfmu{{\mbox{\boldmath $\mu$}}}
\def\bfnu{{\mbox{\boldmath $\nu$}}}
\def\bfxi{{\mbox{\boldmath $\xi$}}}
\def\bfpi{{\mbox{\boldmath $\pi$}}}
\def\bfrho{{\mbox{\boldmath $\rho$}}}
\def\bfsigma{{\mbox{\boldmath $\sigma$}}}
\def\bftau{{\mbox{\boldmath $\tau$}}}
\def\bfupsilon{{\mbox{\boldmath $\upsilon$}}}
\def\bfphi{{\mbox{\boldmath $\phi$}}}
\def\bfchi{{\mbox{\boldmath $\chi$}}}
\def\bfpsi{{\mbox{\boldmath $\psi$}}}
\def\bfomega{{\mbox{\boldmath $\omega$}}}
\def\bfvarepsilon{{\mbox{\boldmath $\varepsilon$}}}
\def\bfvartheta{{\mbox{\boldmath $\vartheta$}}}
\def\bfvarpi{{\mbox{\boldmath $\varpi$}}}
\def\bfvarrho{{\mbox{\boldmath $\varrho$}}}
\def\bfvarsigma{{\mbox{\boldmath $\varsigma$}}}
\def\bfvarphi{{\mbox{\boldmath $\varphi$}}}
\def\bfimath{{\mbox{\boldmath $\imath$}}}
\def\bfjmath{{\mbox{\boldmath $\jmath$}}}

%% file: sections/1.introduction.tex
\clearpage
\section{Introduction}

Exploring large-scale environments with multi-robot systems will enable many robotic applications in both terrestrial \cite{balta2017integrated,santana2007sustainable,best2022resilient,kulkarni2021autonomous,kanellakis2021towards} and planetary \cite{sasaki2020map,autonomyformarsrovers,fordlunarpits,stamenkovic2019next,thomas-agu,blank2021autonomous,morrell2020robotic} settings. In particular, many search and rescue missions require exploration and traversal of large areas within limited time budgets, which can benefit from parallel and coordinated search methods with multi-robot systems.

\ph{DARPA SubT Challenge} The DARPA Subterranean or “SubT” Challenge ~\cite{subt_challenge} is a robotic competition that seeks novel approaches to such search and rescue missions in various subterranean environments, including mines, industrial complexes, and natural caves~\cite{subt_challenge,subt_rules,agha2021nebula}. The primary scenario of interest for the competition is providing autonomous and rapid situational awareness in challenging subterranean environments, where the layout of the environment is unknown. Potential representative scenarios range from planetary cave exploration to rescue efforts in collapsed mines, post-earthquake search and rescue missions in urban underground settings, and cave rescue operations for injured or lost spelunkers. Additional scenarios include missions where teams of autonomous robotic systems are sent to perform rapid search and mapping in support of follow-on operations in advance of service experts, such as astronauts and rescue personnel.

\begin{figure}[ht]
  \centering
  \includegraphics[trim={0 10cm 0 2cm},clip,width=\linewidth]{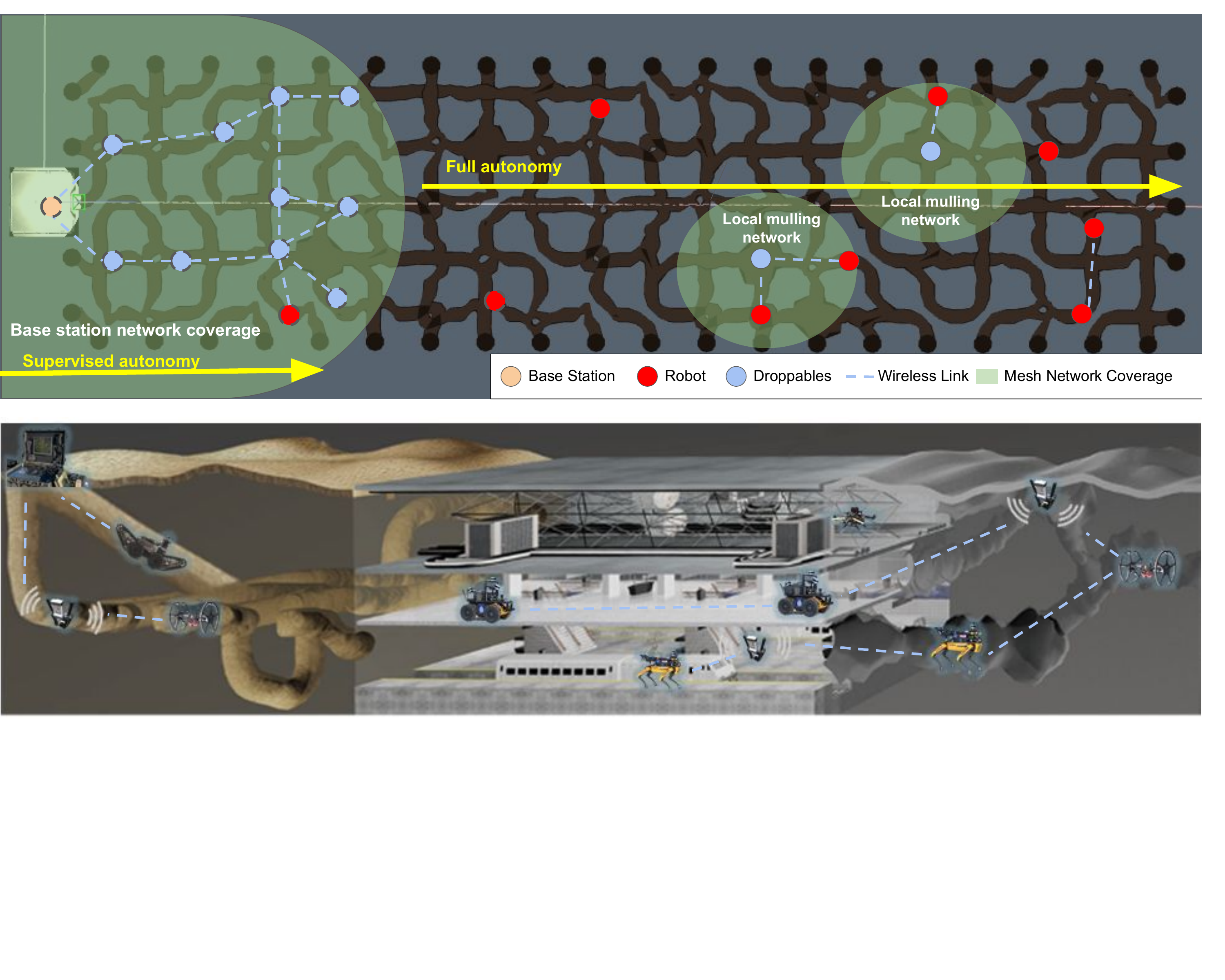}
  \caption{NeBula's Concept of Operation (Figure from \cite{agha2021nebula}). Top: Bird’s eye view of autonomy in a maze-like underground environment. Bottom: Perspective view with our robots in different environments.}
  \label{fig:conops_diagram}
\end{figure}

\begin{figure}[t]
  \centering
  \includegraphics[trim={0 3cm 0 2cm},clip,width=0.9\linewidth]{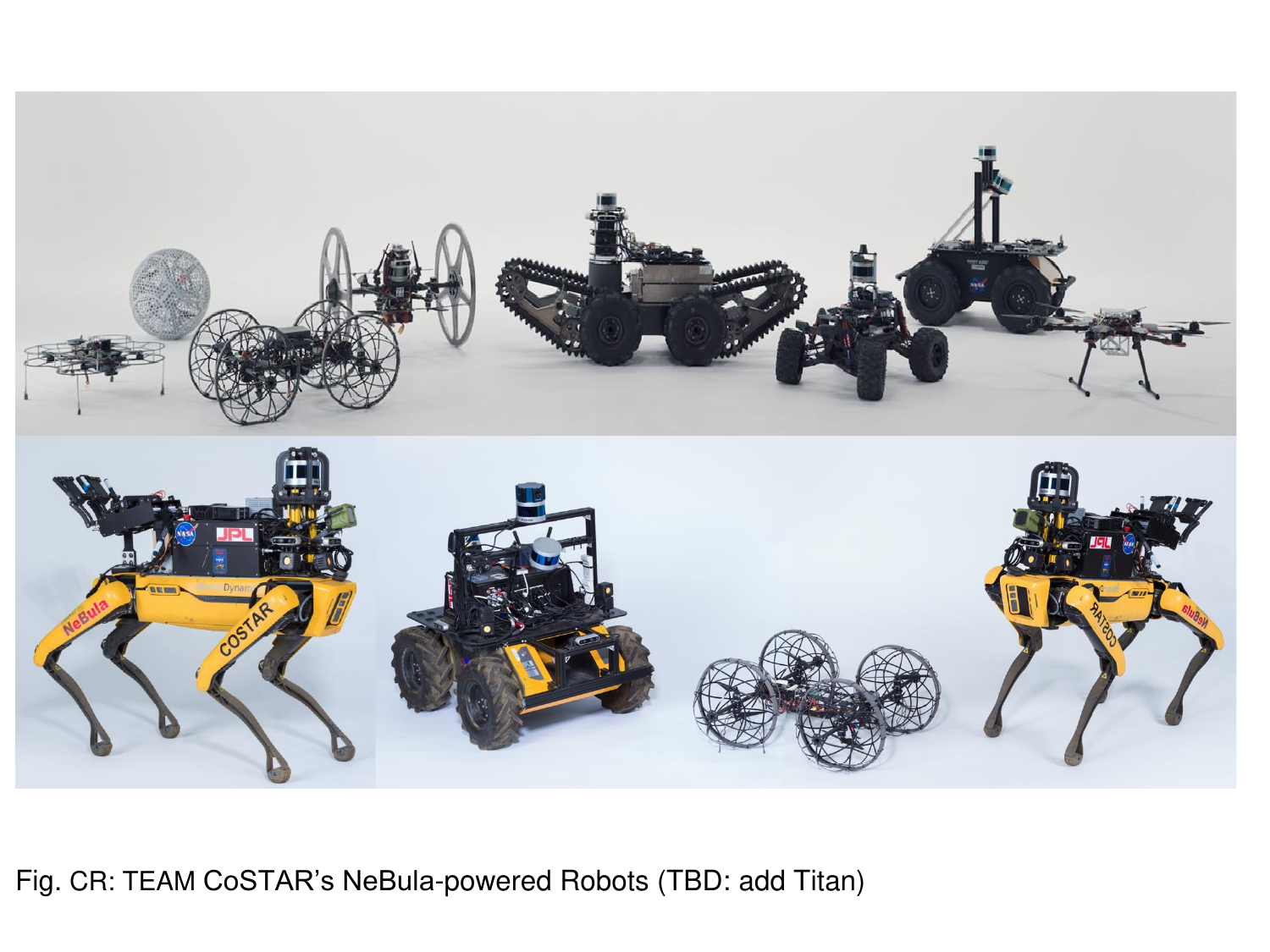}
  \caption{Team CoSTAR's NeBula-powered robots. ~\cite{agha2021nebula}}
  \label{fig:nebula_robots}
\end{figure}

\ph{ConOps (Concept of operation)} In this paper, we discuss a robotic autonomy solution deployed by Team CoSTAR at the DARPA SubT challenge. The concept of operation relies on a team of heterogeneous robots with complementary capabilities, simultaneously exploring the environment and addressing challenges associated with various unknown terrain elements~\cite{agha2021nebula}. In short, robots have several activities and roles, including (1) \textit{Vanguard Operations} to explore the map frontier, (2) \textit{Mesh Network Expansion} to create and extend a wireless mesh network, (3) \textit{Communication-free full autonomy} to operate beyond the communication (comm) network range, (4) \textit{Return to comm} behavior to guide the robots to the rendezvous points or to the mesh network to exchange information with the base station. Fig.~\ref{fig:conops_diagram} and Fig.~\ref{fig:nebula_robots} show this high-level concept and some of the robots.

\ph{NeBula autonomy} The CoSTAR team's robotic fleet is controlled by an autonomy solution, referred to as NeBula (Networked Belief-aware perceptual Autonomy). Details about NeBula's architectural principles, features, and examples can be found in the original NeBula paper ~\cite{agha2021nebula}. NeBula is a field-focused autonomy solution that aims to model uncertainty in decision making to account for discrepancies between computational models and the real-world. It aims at enabling a mobility-agnostic solution and provide modularity to facilitate integration with different robot types.

\begin{figure*}[!t]
  \centering
  \includegraphics[width=0.8\textwidth,angle=-90]{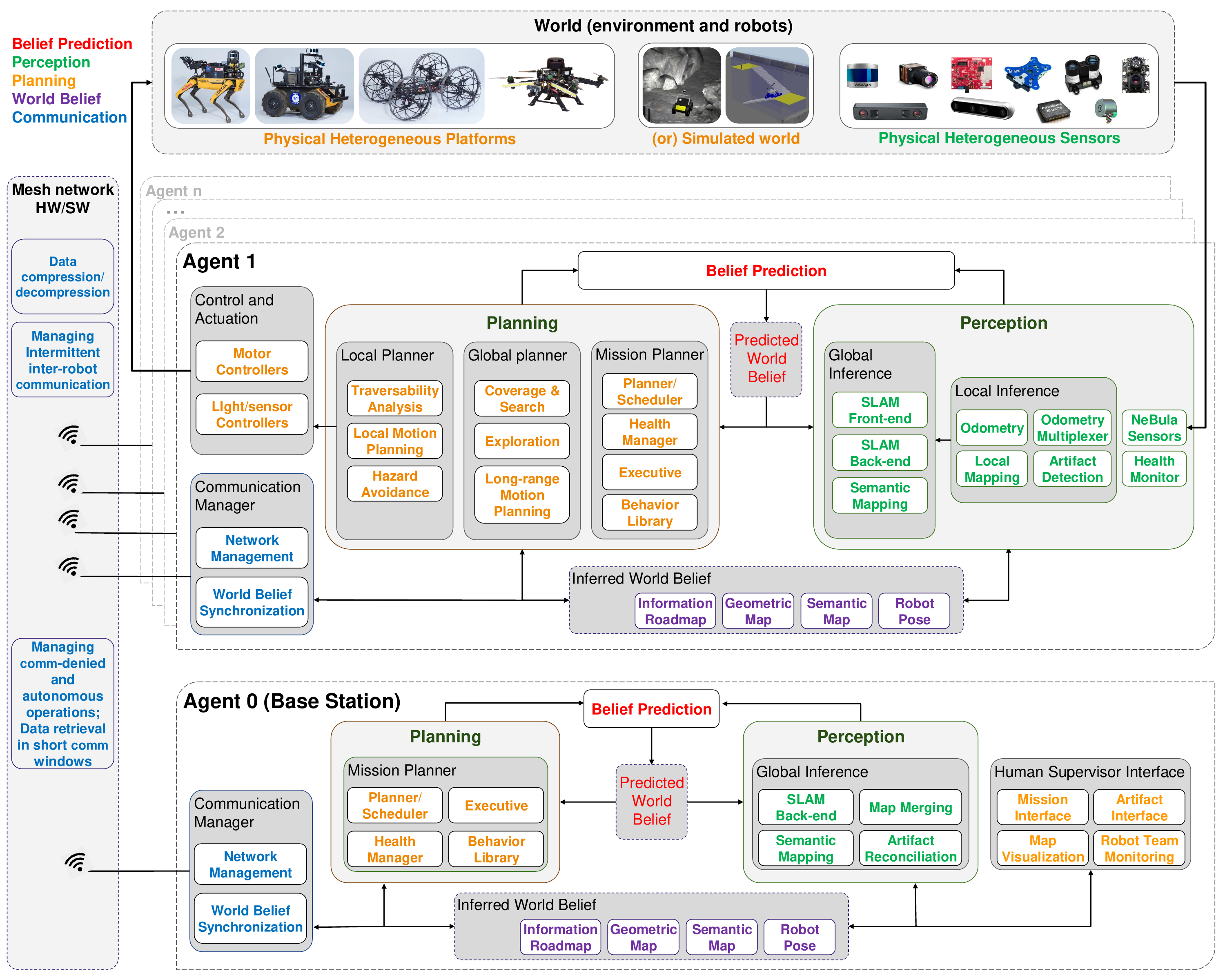}
  \caption{NeBula functional block diagram. ~\cite{agha2021nebula}}
  \label{fig:nebula_architecture}
\end{figure*}

\ph{NeBula architecture}
Quoting from ~\cite{agha2021nebula}, ``Fig.~\ref{fig:nebula_architecture} illustrates the NeBula’s high-level functional block diagram, and will serve as a visual outline of the sections of this paper. The system is composed of multiple assets: mobile robots, stationary comm nodes, and a base station, each of which owns different computational and sensing capabilities. The base station acts as the central component to collect data from multiple robots and distribute tasks, if and when a communication link to the base station is established. In the absence of the communication links the multi-asset system performs fully autonomously. The fundamental blocks include: \textit{(i) Perception} modules, responsible to process the sensory data and create a world model belief, \textit{(ii) Planning} modules that will make onboard decisions based on the current world belief, \textit{(iii) World Belief} blocks that include probabilistic models of the world and mission state, \textit{(iv) Communications} modules, which (when possible), synchronize the shared world models across the robots and the base station, and \textit{(v) Operations} modules, which aid the human supervisor to effectively monitor the system performance and interact with it if and when communication links are established."

\ph{Outline}
We will discuss the new additions and extensions of each module in this improved NeBula architecture (NeBula 2.0), compared to the original NeBula 1.0 paper ~\cite{agha2021nebula}. Specifically, we will discuss advances in the following fronts:
\begin{enumerate}[topsep=0pt,itemsep=-1ex,partopsep=1ex,parsep=1ex]
    \item State estimation in perceptually-degraded environments (\autoref{sec:state_estimation})
    \item Large scale positioning and 3D mapping (\autoref{sec:lamp})
    \item Semantic understanding and artifact detection (\autoref{sec:artifacts})
    \item Risk-aware traversability analysis (\autoref{sec:traversability})
    \item Global motion planning and coverage/search behaviors (\autoref{sec:global_planning})
    \item Multi-robot networking (\autoref{sec:multirobot_networking})
    \item Mission planning and system health management (\autoref{sec:mission_planning})
    \item Mobility systems and hardware integration (\autoref{sec:hardware}).
\end{enumerate}

%% file: sections/2.estimation.tex
\section{State Estimation}\label{sec:state_estimation}
One of the fundamental components of the NeBula 2.0 architecture is the reliable state estimation under perceptually-degraded conditions. This includes environments with large variations in lighting, obscurants (e.g., dust, fog, and smoke), self-similar scenes, reflective surfaces, and featureless/feature-poor surfaces. Furthermore, the rough terrain of many underground environments causes high vibrations, bumps, or slippery motions that degrades the performance of various sensors.

\ph{ARQ and difference with global positioning}
The state estimation of Nebula 2.0 considers a commonly accepted architecture where the estimations with local and global consistencies are processed separately, joining them in a global tree of 6D transformations. In particular, the local consistency is kept during the estimation of a single robot's ego-motion, namely odometry, which provides the 6D transformation between the initial and current position of the robot's base link. This odometry estimation is meant to be at a reasonable frequency, continuous and smooth, so it can be fed to the control module, although it will suffer from global positioning drift. Instead, the global consistency is only kept by the large-scale positioning system which updates the 6D transformation between the global map and the odometry frame of a particular robot. This global positioning system can be of a lower rate, non-continuous or non-smooth but provide precise global positioning. In this section, we introduce the methods that provide a robust and resilient odometry estimation, locally consistent, while the globally consistent state estimation, considering the multi-robot architecture, is described in Section~\ref{sec:lamp}. For more details on our particular odometry estimation methods, please see~\cite{hero2019isrr,palieri2020locus,RIO2020,SeyedMCCKF20,SeyedAMCCKF21,LION,lew2019contact, chen2022dlo} and to understand the context of challenges, strengths and weaknesses see~\cite{past_present_future_of_slam_in_extreme_underground_environments}. %

\ph{Contributions w.r.t. the previous JFR}
We based our Nebula 2.0 odometry estimation approach on our field-tested previous work, Nebula~\cite{agha2021nebula}.  However, the new NEBULA 2.0 version incorporates important novelties, for instance, new estimation approaches (e.g., \cite{palieri2020locus} or the ones described in Section~\ref{sec:drone_hw_&_sw}) or the inclusion of a sensor fusion approach (instead of multiplexing between the redundant odometry sources) based on our Adaptive Maximum Correntropy Criterion Kalman Filter~\cite{SeyedAMCCKF21}. In the following, for the sake of completeness, we briefly describe our heterogeneous and resilient odometry estimation scheme (HeRO)~\cite{hero2019isrr} with particularities of the estimation engines. Later on, we provide a deep description of the adaptive behaviors that better represent our contributions.

\begin{figure}[!ht]
  \centering
  \subfloat[Main HeRO modules and interconnections]{\includegraphics[width=0.75\linewidth]{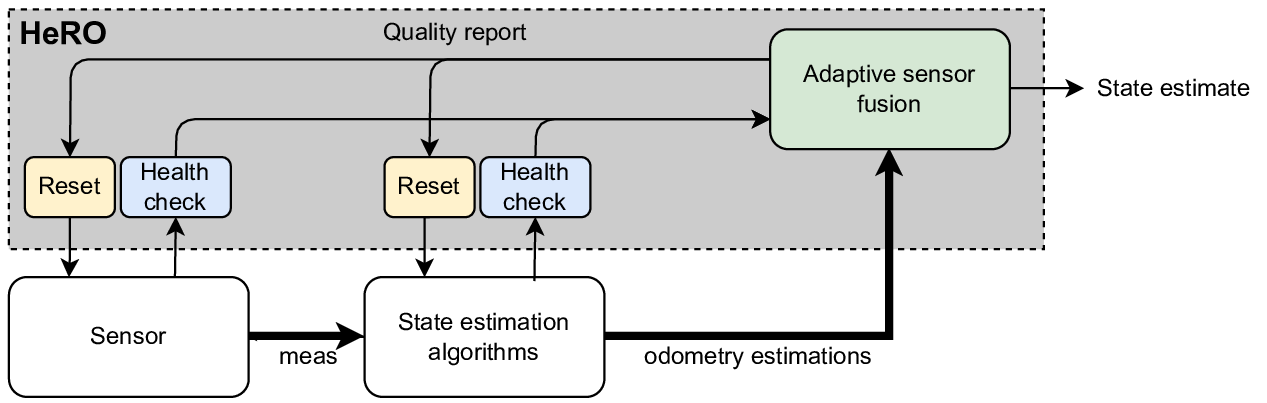}}\\
  \subfloat[Examples of HeRO sensors, algorithms (\emph{e.g.}, LIO: LiDAR-inertial, VIO: Visual-inertial, TIO: thermal-inertial, RIO: RaDAR-inertial, OF: Optical flow, CIO: contact-inertial or WIO: Wheel-inertial odometries) and observable state components]{\includegraphics[width=\textwidth]{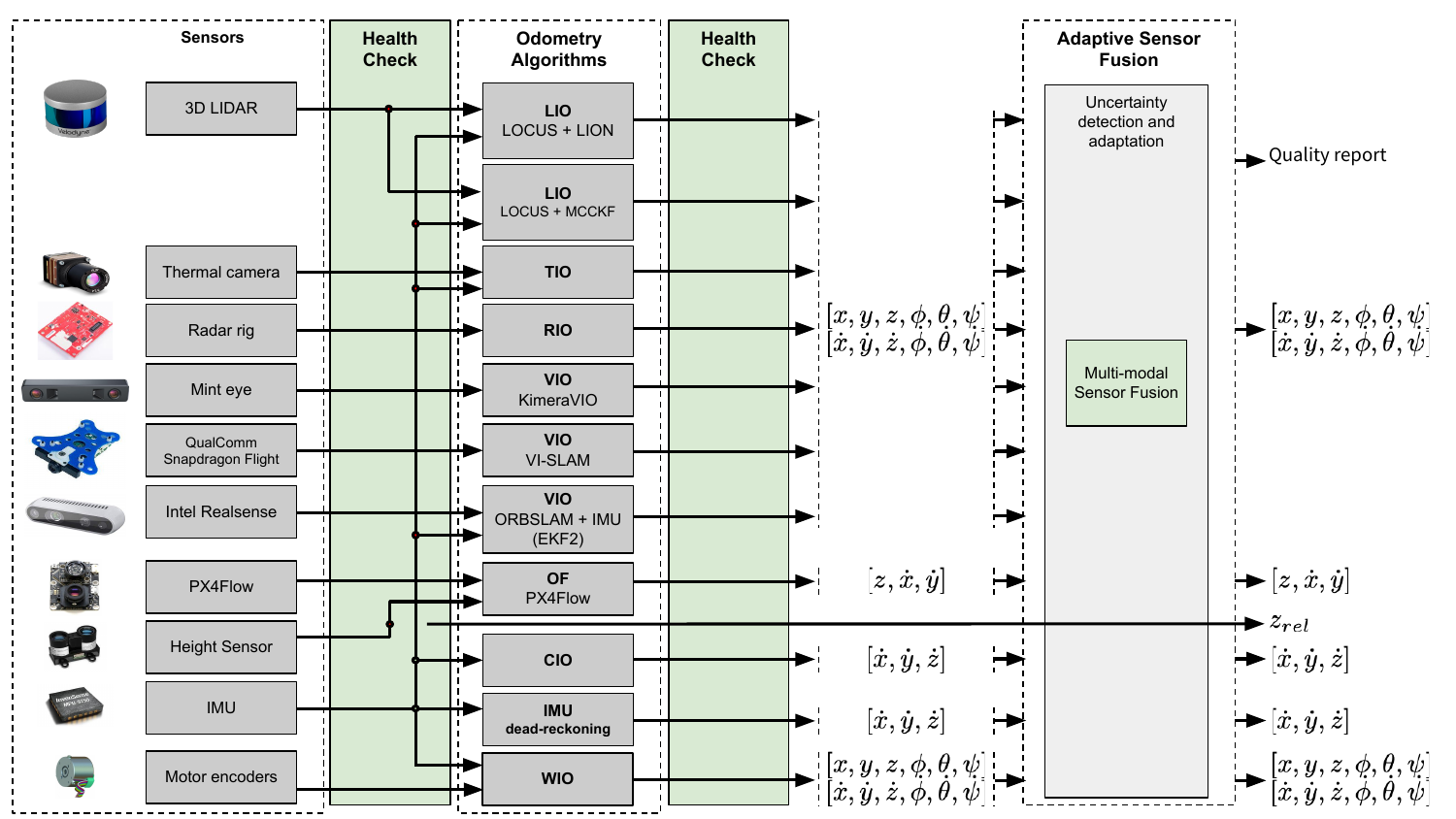}}
  \caption{NeBula's state estimation architecture
  }
  \label{fig:hero}
\end{figure}

\ph{HeRO (Heterogeneous and Resilient Odometry Estimation)}
The proposed HeRO~\cite{hero2019isrr} sensor fusion architecture, shown in Fig.~\ref{fig:hero}, considers resiliency and heterogeneity in both sensing and estimation algorithms.
It is designed to expect and detect failures while adapting the behavior of the system to ensure safety thanks to a framework of estimation algorithms running in parallel supervised by a resiliency logic. The HeRO has three main functions: a) perform confidence tests in data quality (measurements and individual odometry estimation methods) and check health of sensors and algorithms; b) re-initialize those algorithms that might be malfunctioning; c) generate a smooth state estimate by fusing the inputs based on their quality and the error of the expectations (see below how the adaptive behaviors work).
The output of the fusion block includes the \emph{state-estimate} and a \emph{quality-report} for each single estimation method and sensor input.  When failures are detected, the quality-report, which is based on our adaption strategy, resets each sensor or estimating method. The \emph{state-estimate} provides a full estimate of robot's states by fusing different modalities which is then used by the rest of the Nebula stack.  

\ph{Heterogeneous complementary algorithms}  In addition to selecting heterogeneous sensing modalities, HeRO uses heterogeneous odometry algorithms, \emph{e.g.}, LiDAR-inertial (LIO), visual-inertial (VIO), thermal-inertial (TIO), kinematic-inertial (KIO), contact-inertial (CIO) or RaDAR-inertial (RIO), running in parallel to decrease the probability of a state estimation failure.
The key idea behind HeRO is that any single state estimation source can carry errors, either due to failures in sensor measurements, algorithms or both, but having a complete failure becomes increasingly rare as the number of heterogeneous parallel approaches increases.
HeRO is front-end agnostic, accepting a various algorithmic solutions and with the ability to incorporate either tightly or loosely coupled approaches. 
However, to take advantage of all possible mobility modes, there is a need for estimating position, orientation, velocity and, ideally, acceleration.
HeRO is tailored to incorporate a vast variety of estimation algorithms.
\rev{Our solution considers software-synchronized sensors (common clock synchronization after initialization), with extrinsic calibrations roughly obtained from the robot model designs and fine tuned used optimization approaches such as Kalibr (camera-imu)~\cite{FurgaleRehderEtAl2013,RehderNikolicEtAl2016} or LiDAR-align (LiDAR-LiDAR)~\cite{ethz_lidar_align}; or by aligning with the robot manufacturer frames.}
We rely on a variety of in-house and open-source algorithms for sensor fusion. A few examples are as follows: WIO uses an Extended Kalman Filter (EKF) to fuse the measurements of the wheel encoders and those from an IMU. 
CIO also takes advantage of an EKF but this time including the modelling of the contacts~\cite{lew2019contact}.
The optical flow approach (OF), VIO, and thermal imagery fusion (TIO) leverage a combination of opensource and commercial solutions, including \revv{PX4Flow \cite{px4Flow}, ORBSLAM \cite{murORB2}, Qualcomm VI-SLAM~\cite{qualcommtechnologiesinc.MachineVisionSDK}, and the MIT KimeraVIO \cite{Rosinol20icra-Kimera}, ROTIO~\cite{khattak19}}, among others.
The RIO algorithm is our own development presented in~\cite{RIO2020}. In the following we describe the main characteristics of our LiDAR-based odometry method (LOCUS 2.0), used in our ground vehicles. The estimation approaches running on the aerial vehicles are described in section~\ref{sec:drone_hw_&_sw}.

\newpage


\subsection{LOCUS 2.0} \label{LOCUS}
LiDAR odometry has attracted considerable attention as a robust localization method for autonomous robotic operation in complex GPS-denied environments. However, achieving reliable performance on heterogeneous robotic teams operating in large-scale settings under computation and memory constraints remains an open challenge. While several LiDAR odometry algorithms can achieve remarkable accuracy, their computational cost can still be prohibitive for real-time functioning on computationally restricted platforms. Furthermore, the majority of existing approaches maintain in memory a global map for localization purposes, rendering them unsuitable for large-scale explorations on memory-constrained robots.

\begin{figure} [h]
	\centering     
	\includegraphics[width = 0.8\linewidth]{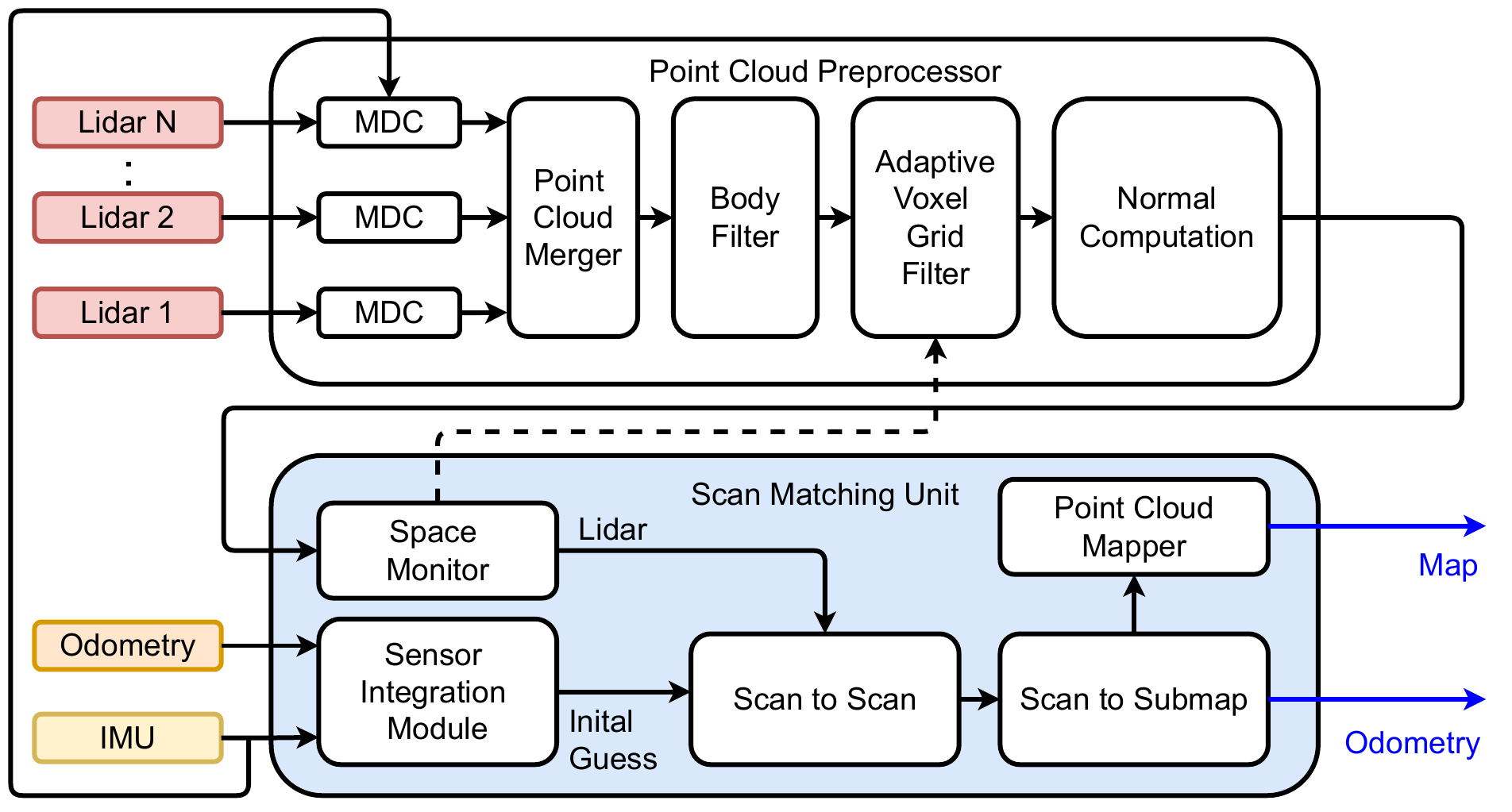}
	\caption{\footnotesize{LOCUS 2.0 architecture}}
	\label{fig:locus_2_architecture}
\end{figure}

To overcome these challenges, we update our LiDAR odometry solution LOCUS \cite{palieri2020locus} and present LOCUS 2.0 \cite{reinke2022locus}, a multi-sensor LiDAR-centric solution for high-definition odometry and 3D mapping in real-time with enhanced computational efficiency and decreased memory demand. While maintaining the overall architecture of its predecessor, LOCUS 2.0 introduces different algorithmic and system-level improvements that decrease the computation load and memory usage, enabling real-time ego-motion estimation and mapping over large-scale explorations under computation and memory constraints. The current proposed architecture reported in Fig.~\ref{fig:locus_2_architecture} is composed of three main components: \textit{i)} point cloud preprocessor, \textit{ii)} scan matching unit, \textit{iii)} sensor integration module.

The \textit{point cloud preprocessor} is responsible for the management of multiple-input LiDAR streams (e.g. motion distortion correction, merging, filtering) to produce a unified 3D data product that can be efficiently processed by the \textit{scan matching unit}. A first main novelty of LOCUS 2.0 is the introduction in the \textit{point cloud preprocessor} of an \textit{adaptive voxel grid filter}: this maintains the number of voxelized points to a fixed value to ensure near-constant runtime of the LiDAR registration stage regardless of the environment geometry and number of LiDARs used by the robot. A second main novelty of LOCUS 2.0 is the integration in the \textit{point cloud preprocessor} of a \textit{normal computation} module: this computes normals from the instantaneous voxelized point cloud and forwards this information to the scan-matching unit to enhance the computational efficiency in the LiDAR registration stage as described below. 

The \textit{scan matching unit} performs a Generalized Iterative Closest Point (GICP) \cite{segal2009generalized} based scan-to-scan and scan-to-submap registration to estimate the 6-DOF motion of the robot. As explained in \cite{palieri2020locus}, in robots with multi-modal sensing, if available, we use an initial guess from a non-LiDAR source provided by the \textit{sensor integration module}. This eases the convergence of the GICP in the scan-to-scan matching stage, by initializing the optimization with a near-optimal seed that improves accuracy and reduces computation, enhancing real-time performance. A third main novelty of LOCUS 2.0 is the introduction of a novel normals-based GICP formulation which leverages the point cloud normals information provided by the \textit{point cloud preprocessor} to approximate the point covariance calculation and reduce the computation time of point cloud alignment. In comparison to its predecessor, this allows LOCUS 2.0 to not recalculate covariances for each point in the input scan and in the input submap at every registration, reducing computation time and enhancing the overall computational efficiency. As the first version of LOCUS maintained the global octree map in memory, a fourth main novelty of LOCUS 2.0 is the introduction of two selectable map sliding window approaches: \textit{i)} a multi-threaded octree mapper, and a \textit{ii)} incremental k-dtree mapper \cite{cai2021ikd} (ikd-Tree). These are responsible to maintain in memory only a robot-centered submap of the environment, to bound memory consumption and enable efficient large-scale explorations on memory constrained robots. A detailed description of LOCUS 2.0 can be found in \cite{reinke2022locus}. 


\subsection{Adaptive Sensor Fusion} 
There has been a lot of work done on single-modal odometry estimation, such as LiDAR-based methods~\cite{palieri2020locus, shan2020lio, xu2021fast}, vision-based methods~\cite{Rosinol20icra-Kimera}, and so on. However, each single-modal odometry estimation method has a specific characteristic. As an example, when the environment is well-structured, LiDAR-based odometry,  can provide extremely accurate state estimate. However, they suffer from low-rate and high-delay state estimation due to computational overhead.  In addition, relying on a single-modal odometry estimation reduces the robustness of the state estimation pipeline to sensor or algorithm failures. In recent years, multi-modal sensor fusion has received a great amount of attention, and a variety of approaches have been developed. They can be categorized into two main approaches: loosely- and tightly-coupled methods. Loosely-coupled methods (e.g.,~\cite{zhang2018laser,khattak2020complementary}) have gained more popularity due to their simplicity, extendibility, and low computational efforts. Tightly-coupled approaches (e.g.,~\cite{shao2019stereo}) are known for more accuracy and robust performance. They are, however, computationally expensive and difficult to expand to other modalities in order to be more compatible with varied environments. A new method proposed in~\cite{zhao2021super} uses the combination of loosely and tightly approaches to fuse different modalities.   

Most of the multi-modal fusion methods use constant matrices  (i.e., static noise covariance matrices) to quantify the uncertainty for each measurement. These matrices are usually obtained by trial and error approaches.  However, in practice, the uncertainty is a function of time and environment, and it can fluctuate over time. Despite the fact that numerous studies have been done on estimating noise statistics using adaptive Kalman filtering (AKF) approaches, they are either not in a fusion framework or only valid when the noise distribution is Gaussian.   
Here, we propose a fusion framework in a loosely-coupled scheme that adaptively detects the noise covariance matrices. The approach, called Adaptive Maximum Correntropy Criterion Kalman Filtering (AMCCKF), combines the benefits of the AKF~\cite{mohamed1999adaptive, huang2020slide} and MCCKF~\cite{fakoorian2019robust} to achieve robust and accurate odometry estimation of mobile robots. Although the AMCCKF adaptively estimates the noise covariance matrices (similar to AKF), thanks to MCC property, it is robust when the measurement or process noise is not Gaussian, e.g., contains jumps. The proposed filter not only improves the accuracy of state estimates, but also reduces the error in hand-tuning the noise covariance matrices.
Two variants of the AMCCKF are derived: a VB-based AMCCKF and a residual-based AMCCKF. The two designs can be chosen based on required accuracy and onboard computation capabilities. The detailed derivation of AMCCKF can be found in~\cite{SeyedAMCCKF21}.

\subsection{Experimental Analysis}


\textbf{LOCUS 2.0:} We test the performance of LOCUS 2.0 gathering the data collected by wheeled rovers Husky robots from Clearpath Robotics and legged Spot robots from Boston Dynamics in a variety of real-world environments such as caves, tunnels, and abandoned factories which was released as the Nebula odometry dataset in \cite{reinke2022locus}. In the following, we present results on: \textit{i)} decreased computational load enbaled with GICP from normals, \textit{ii)} near-constant runtime for LiDAR registration enabled with the adaptive voxelization strategy, \textit{iii)} decreased memory usage over large-scale explorations enabled by the map sliding window approaches multi-threaded octree and ikd-Tree. 

\begin{figure}[th!]
	\centering     
	\includegraphics[width = 0.6\linewidth]{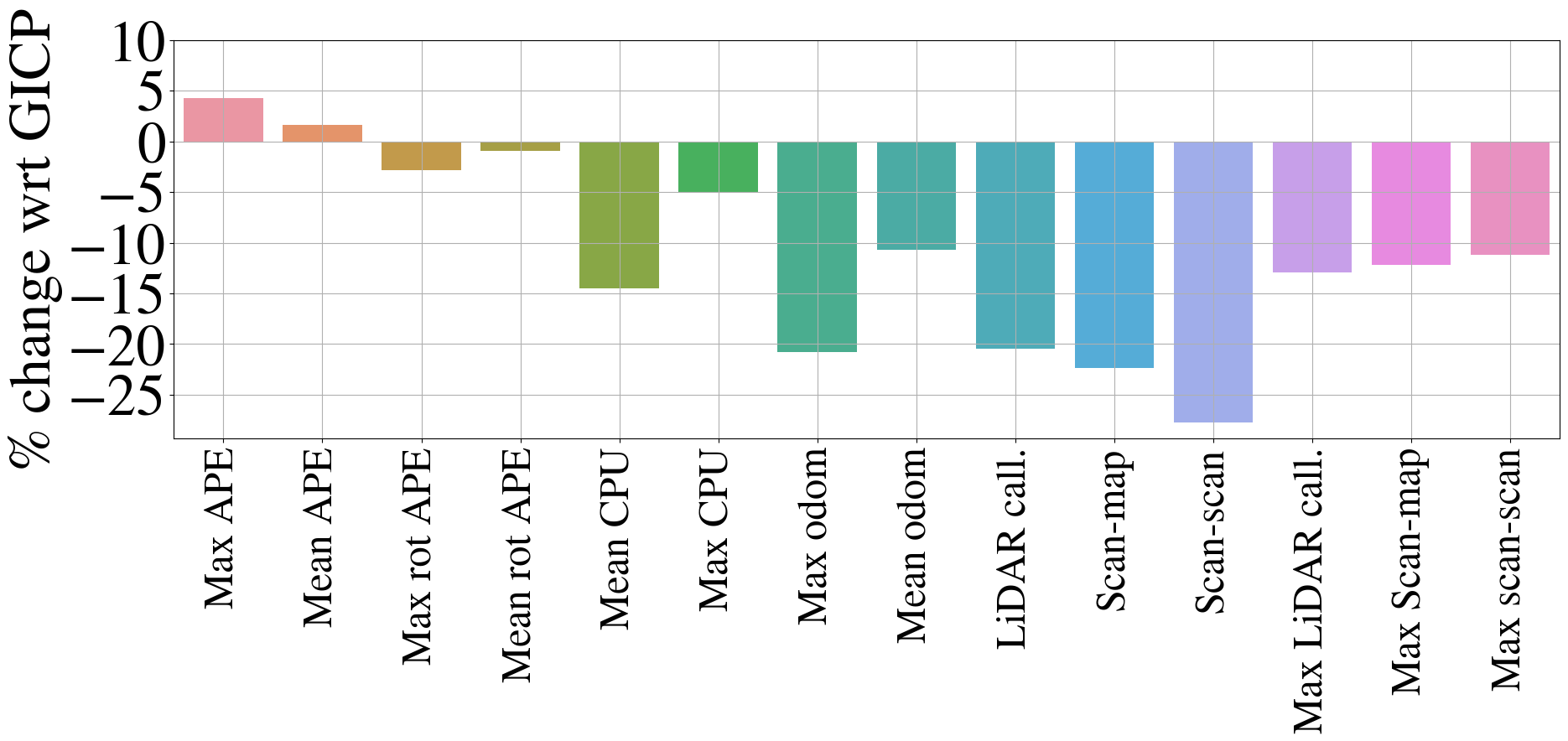}
	\caption{\footnotesize{Comparison of GICP from normals and GICP. Results show that GICP from normals decreases CPU load and computation time, while introducing minimal degradation in estimation accuracy.}}
	\label{fig:gicp_from_normals}
\end{figure}

First, we report in Fig.~\ref{fig:gicp_from_normals} results on computational efficiency enhancement provided with the proposed GICP from normals with respect to the standard GICP formulation. Experiments show that the proposed GICP from normals formulation results in significantly enhanced computation performance with respect to the standard GICP, with negligible decrease on localization accuracy. In particular, Fig.~\ref{fig:gicp_from_normals} shows the average percentage change across the NeBula odometry dataset for each metric with respect to the standard GICP. The statistics were calculated based on $5$ runs for each dataset. 
GICP from normals reduces all the computational metrics in LOCUS 2.0: mean and max CPU usage, mean and max odometry delay, scan-to-scan, scan-to-submap, LiDAR callback duration and their maximum times. The computation burden is reduced by $18.57\%$. This reduction results in an increase of the odometry update rate of $11.10\%$ which is beneficial for other parts of the system such as planning and control algorithms. One drawback of this method leads to a slight increase in the mean and max Absolute Position Error (APE). The reason is that normals are calculated from sparse point clouds and stored in the map without further recomputation. In the standard GICP approach instead covariances are recalculated from dense map point clouds. 

\begin{figure}[th!]
\begin{subfigure}{0.45\columnwidth}
  \centering
  \includegraphics[trim={0 0cm 0 0},clip, width=1\columnwidth]{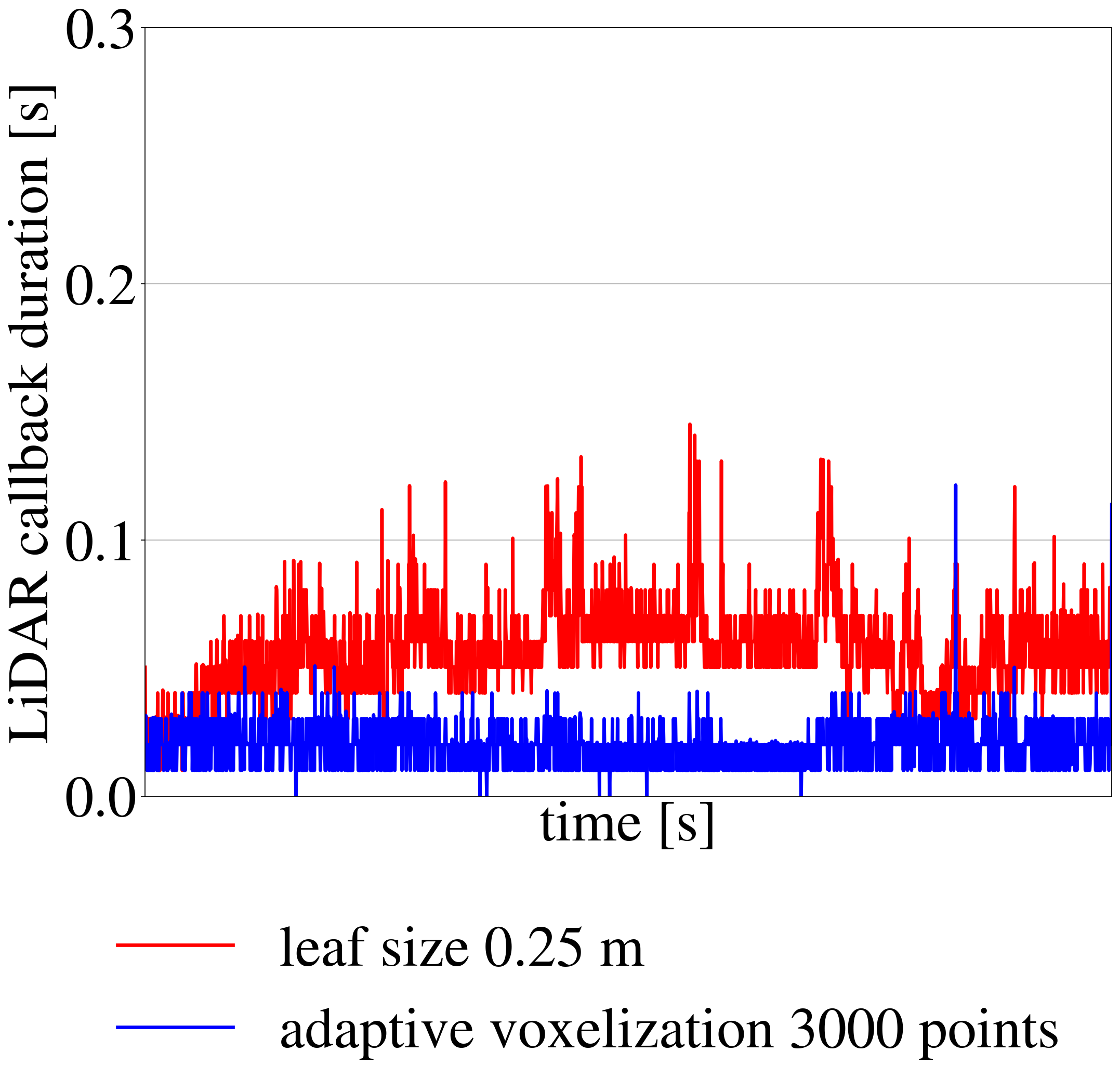}
  \caption{\textcolor{black}{LiDAR callback duration for Husky robot in urban environment.}}
\end{subfigure}
\begin{subfigure}{0.45\columnwidth}
  \centering
  \includegraphics[trim={0 0cm 0 0},clip, width=1\columnwidth]{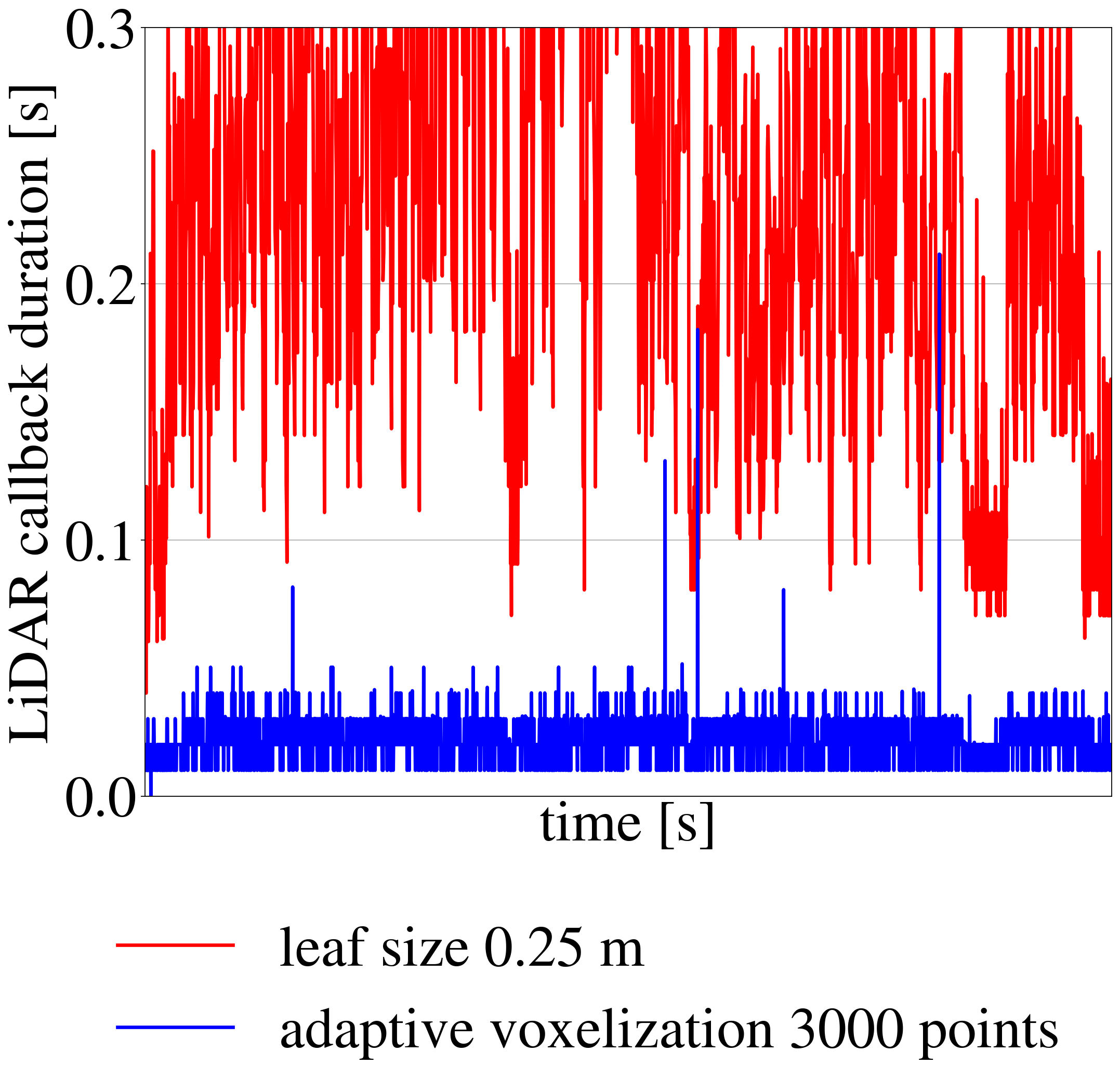}
  \caption{\textcolor{black}{LiDAR callback duration for Husky robot in tunnel/cave environment.}}
\end{subfigure}
  \caption{Computation time difference for adaptive
voxelization with 3000 points used in LOCUS 2.0 (blue) and 0.25 m constant leaf size used in LOCUS 1.0 (red). a) Urban environment using 2 LiDARs. b) Tunnel/Cave environment using 3 LiDARs. Results show that the adaptive voxelization strategy enables the average computation time to be bounded within real-time requirements (e.g. 0.1 s for 10 Hz lidar rate) regardless of the environment type and lidar configuration.}
  \label{fig:keeping_points_adaptive_voxelization}
\end{figure}

Second, we report in Fig.~\ref{fig:keeping_points_adaptive_voxelization} results on the near-constant runtime for LiDAR registration enabled with the adaptive voxelization strategy of LOCUS 2.0.  
The adaptive voxel grid filter keeps the number of points relatively consistent across a dataset, no matter what the input $N_{desired}$ is (see \cite{reinke2022locus} for details).
However, there is still some variability in the computation time across a dataset. Nonetheless, as shown in Fig.~\ref{fig:keeping_points_adaptive_voxelization}, the approach produces a consistent average computation time across different environments and sensor configurations without any large spike in computation time. This performance gives more predictable computational loads, regardless of robot or environment, as shown in  Fig.~\ref{fig:keeping_points_adaptive_voxelization}, where the LiDAR callback time is similar for two different datasets with different sensor configurations. The adaptive voxelization ensures that the computation time is on the same level. For example, as shown in Fig.~\ref{fig:keeping_points_adaptive_voxelization}, in the urban environment Husky is equipped with two LiDARs, in the tunnel environment Husky is equipped with three LiDARs, and still the LiDAR callback time is maintained at the same value. 

\begin{figure}[th!]
\begin{subfigure}{0.45\columnwidth}
  \centering
  \includegraphics[trim={0 0cm 0 0},clip, width=1\columnwidth]{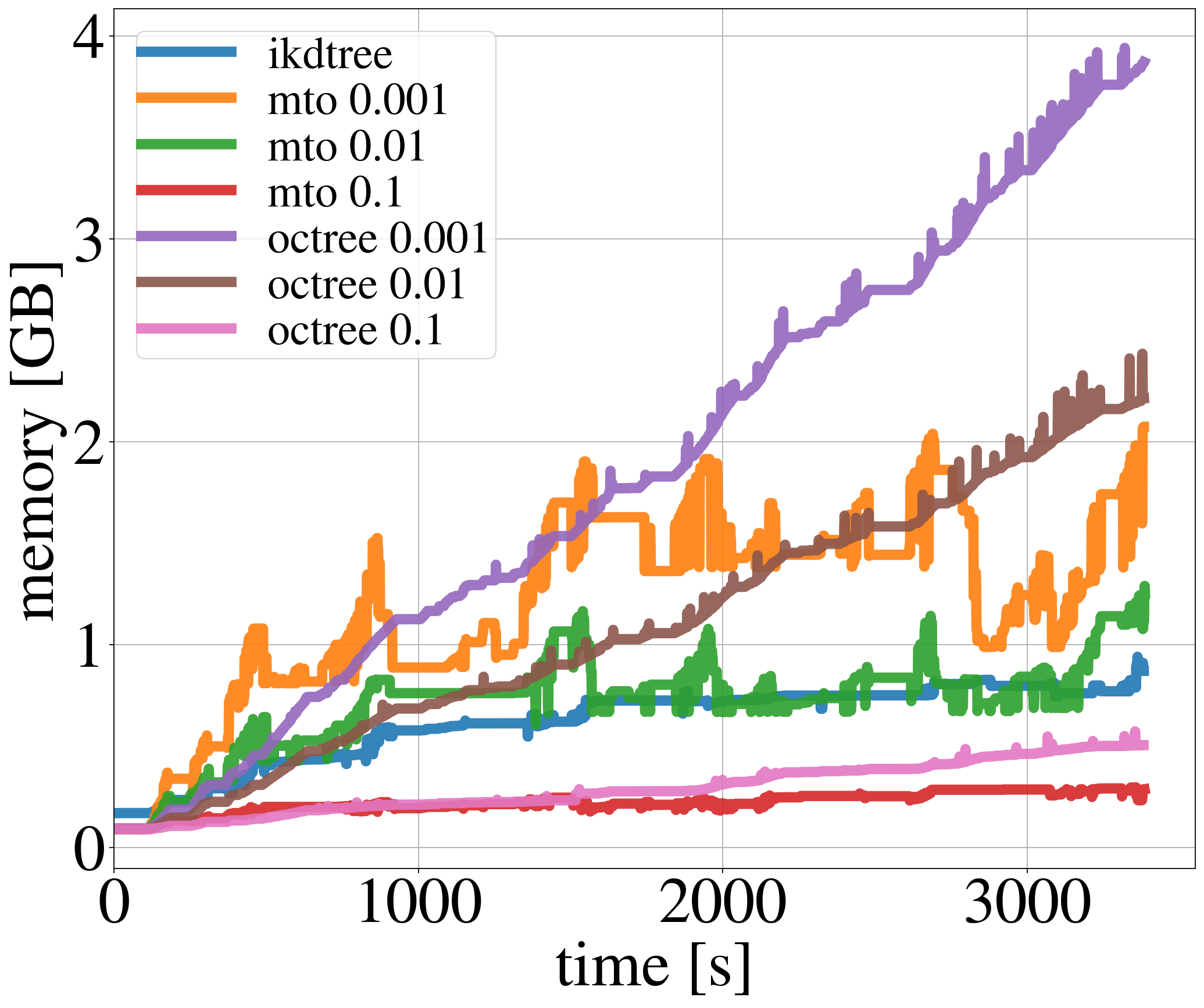}
  \caption{\textcolor{black}{Memory usage over time for Spot in cave environment.}}
\end{subfigure}
\begin{subfigure}{0.45\columnwidth}
  \centering
  \includegraphics[trim={0 0cm 0 0},clip, width=1\columnwidth]{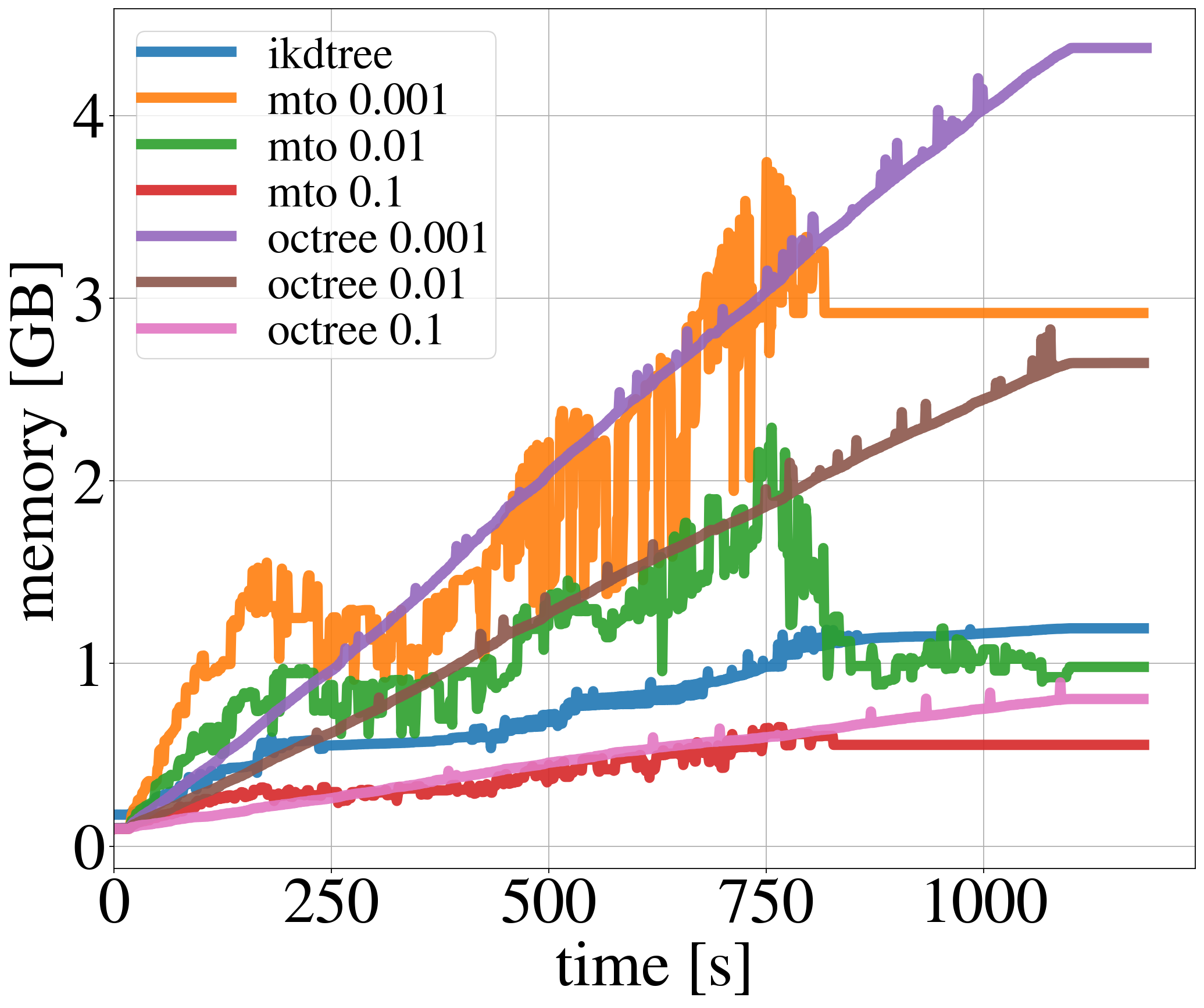}
  \caption{\textcolor{black}{Memory usage over time for Husky in tunnel environment.}}
\end{subfigure}
  \caption{Memory usage over time for urban and cave environments from NeBula odometry dataset \cite{reinke2022locus}. }
  \label{fig:exp:mapping_techniques}
\end{figure}

Finally, we report in Fig.~\ref{fig:exp:mapping_techniques} results on the benefits provided on memory consumption by the sliding-window map approaches compared to classic, static octree-based structures. In these experiments, LOCUS 2.0 uses: ikd-Ttree and multi-threaded octree (mto).
The octree with leaf size $0.001~m$ is the baseline used in LOCUS 1.0 to maintain the full map information. To evaluate the effect of different parameter configurations, we run the multi-threaded octree with leaf sizes of $0.1~m$, $0.01~m$, and $0.001~m$. For sliding-window approaches, the map size is kept at $50~m$. For the scan-to-scan and scan-to-submap registration stage, \textit{GICP from normals} is used with the parameters chosen based on previous experiments. Fig.~\ref{fig:exp:mapping_techniques} profiles the memory usage over time of the different approaches for one of the urban and cave datasets from the NeBula odometry dataset. The most significant memory occupancy is for octree and multi-threaded octree with $0.001~m$ leaf size. The ikd-Ttree achieves similar performance in terms of memory usage as the multi-threaded octree with leaf size $0.01m$. The multi-threaded octree with $0.1m$ leaf size provides instead the smallest memory usage, at the cost of a slightly decreased localization accuracy (see Table \ref{tab:map_comparison}).

\begin{table}[t]
\caption{\footnotesize{Relative memory and CPU change.
	}}
	\label{tab:map_comparison}
	\vspace{-0.3cm}
	\begin{center}
			\begin{tabular}{c c c c c c } \hline
				 & ikdtree & 	mto $0.001$ & mto $0.01$ & mto $0.1$ & octree $0.001$ \\ 
				\toprule
    Max APE   & $+15.89\%$ & $+7.74\%$ & $+13.29\%$ & $+15.41\%$  & X \\ 
				\midrule
    Mean APE  & $+3.86\%$ & $-1.31\%$ & $+2.84\%$ & $+6.04\%$  & X \\ 
				\midrule
				Memory  & $-68.09\%$ & $-38.88\%$ & $-62.15\%$ & $-87.76\%$  & X \\ 
				\midrule
				CPU & $9.36\%$ & $50.42\%$ & $44.36\%$ & $19.61\%$ & X \\ 
				\bottomrule
			\end{tabular}
	\end{center}
	\vspace{-0.3cm}
\end{table}

\begin{figure}[th!]
\begin{subfigure}{\columnwidth}
  \centering
  \includegraphics[trim={0 0cm 0 0},clip, width=0.75\columnwidth]{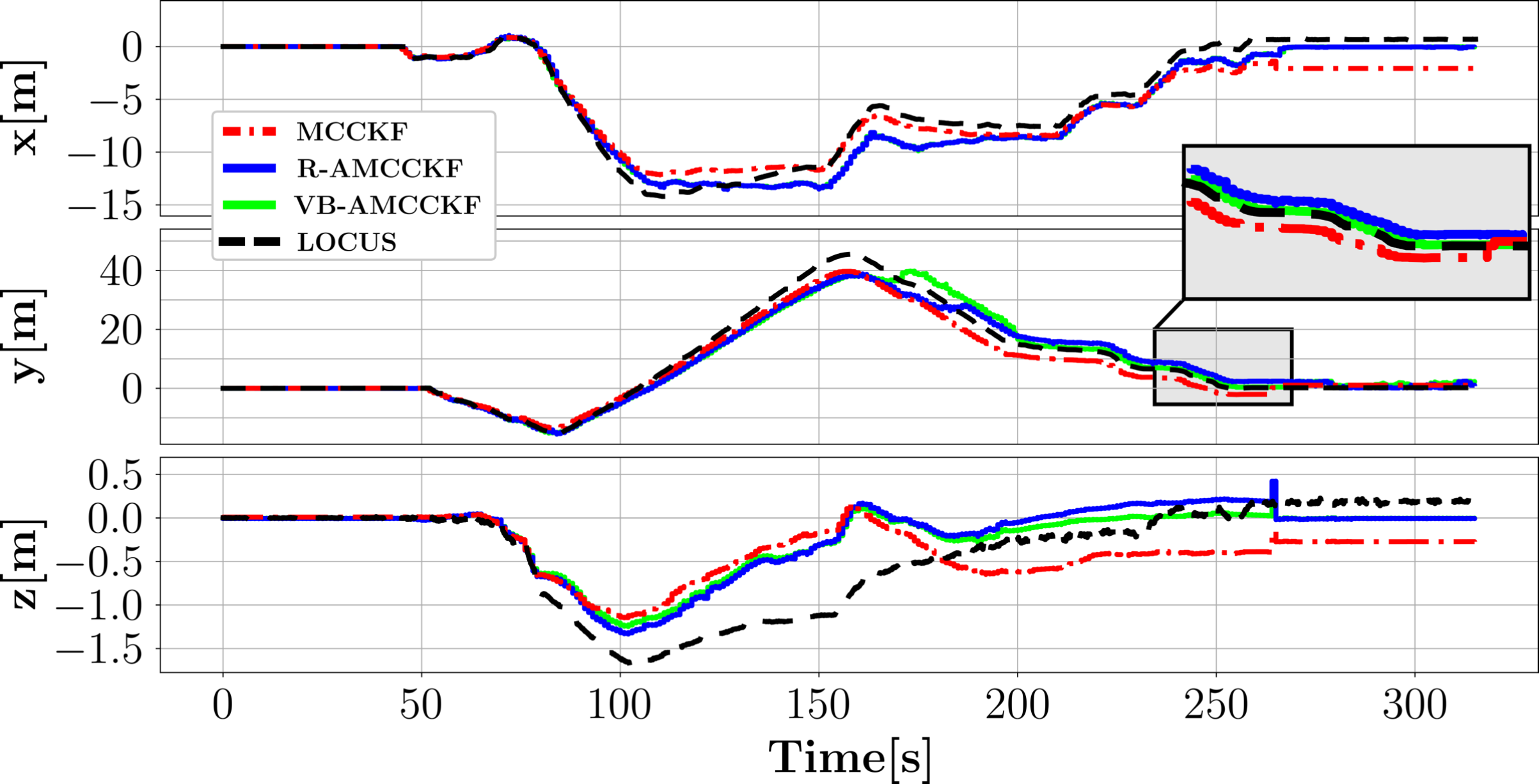}
  \caption{\textcolor{black}{Husky position estimations.}}
  \label{fig:sub-first_ammcckf}
\end{subfigure}

\begin{subfigure}{\columnwidth}
  \centering
  \includegraphics[trim={0 0cm 0 0},clip, width=0.75\columnwidth]{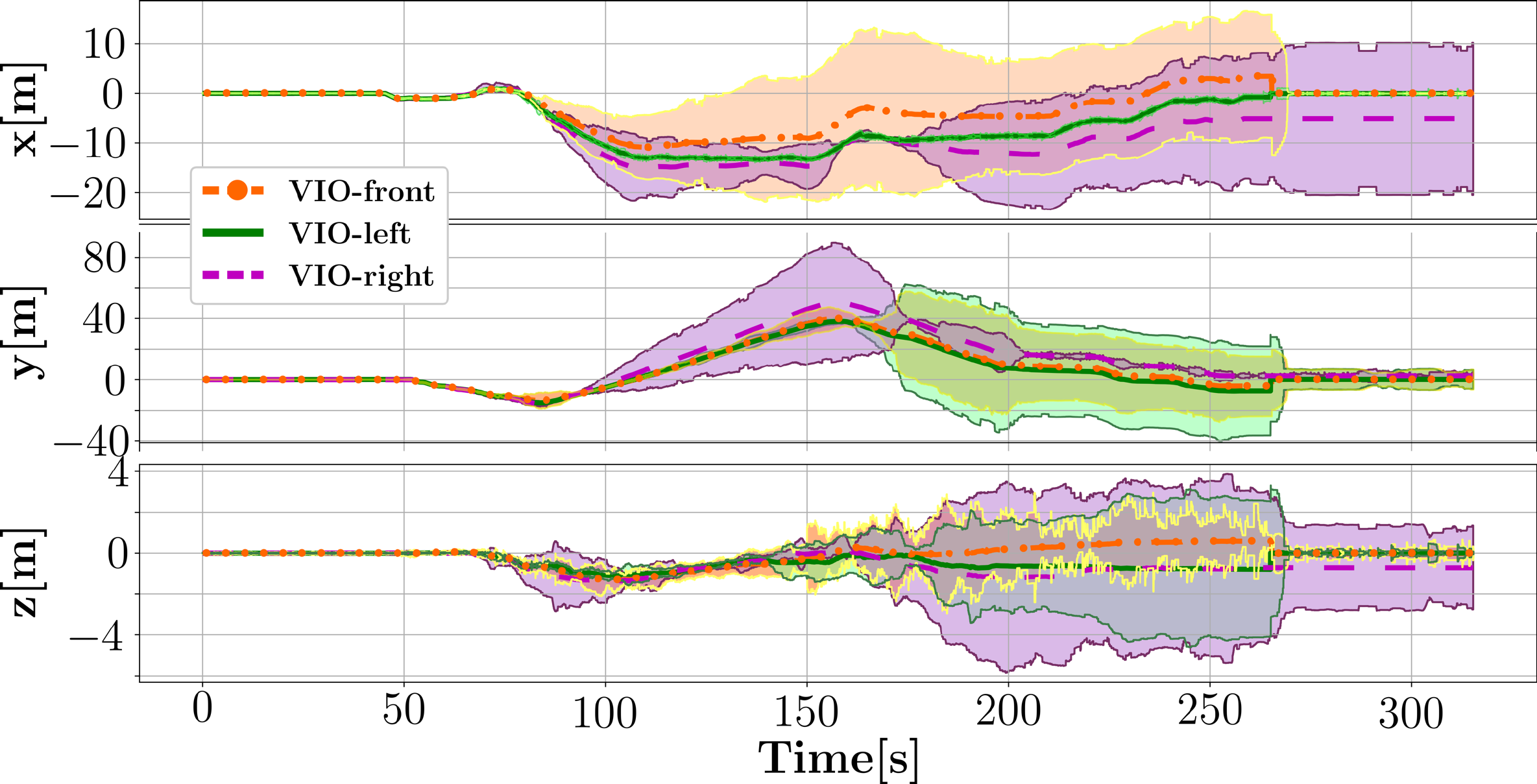}
  \caption{\textcolor{black}{The VIO estimations with their \emph{3-sigma} bands from the measurement noise covariance.}}
  \label{fig:sub-second_ammcckf}
\end{subfigure}
  \caption{The VIO sensors are fused by VB-AMCCKF and R-AMCCKF. The position of the ground robot is shown in (a) and $\pm 3$ times of the square root of estimated covariance of each VIO sensors is shown in (b). Note that we only show the position of the robot here for the purpose of illustration, however same results hold for rest of the states. }
  \label{fig:pos_cov_husky}
\end{figure}

\textbf{AMCCKF:}
We also present experimental results of our proposed adaptive fusion method in various environments. Here, we provide one of the experiments on our wheel-robot, Husky, which is equipped with different types of exteroceptive and proprioceptive sensors including IMU, LiDAR, RealSense T265 camera, and so on.  For this test, the AMCCKF fused three commercial-of-the-shelf VIO from T265 cameras with an IMU. Fig.~\ref{fig:sub-first_ammcckf} first compares the traditional \mbox{MCCKF}, the proposed VB-AMCCKF and R-AMCCKF with the LOCUS 2.0 that was discussed in Section~\ref{LOCUS}. This LOCUS is used to give ground-truth for comparison since it usually delivers high-accuracy state estimates.  The goal of AMCCKF is to achieve the same level of accuracy as LOCUS while providing a higher rate and lower latency.      
Here, the traditional MCCKF cannot reach the accuracy of the VB-AMCCKF or R-AMCCKF because the MCCKF lacks adaptive behavior; both VB-AMCCKF and R-AMCCKF generate similar estimates to LOCUS at the end of the trajectory. 
Fig.~\ref{fig:sub-second_ammcckf} shows how the R-AMCCKF adapts for each VIO sensor. The modification of individual dimensions on the covariance matrices allow the AMCCKF to take advantage of the ``best" available estimate for each individual dimension.
For instance, notice how VIO-left performs best in the estimation for $x$ position, while a combination of front and right cameras prevail for $y$, and VIO-front for $z$.

%% file: sections/3.slam.tex
\section{Large-Scale Positioning and 3D Mapping}\label{sec:lamp}
While NeBula's state estimates are accurate and robust through many challenging environments, there is still drift over large scales of exploration, and through especially challenging perceptual degradation. Incorporating loop closures in a Simultaneous Localization and Mapping (SLAM) system is the common way to constrain drift, yet using multi-robot loop closures provides an opportunity to even further increase accuracy. This is the approach taken in our multi-robot Simultaneous Localization and Mapping (SLAM) module called LAMP 2.0 (Large-scale Autonomous Mapping and Positioning)~\cite{chang2022LAMP2}. LAMP 2.0 is designed specifically to accommodate to systems of over 8 heterogeneous agents exploring over many kilometers of distance in a scalable, prioritized and real-time manner. We use a centralized architecture, leveraging the fact that search and rescue scenarios need to get their data back to the operator, just as in the SubT Challenge. In this section we outline how LAMP 2.0 is designed to accommodate heterogenous robot teams for robust large-scale exploration. We then show in results from field demonstrations the benefit of performing multi-robot SLAM, and the effectiveness of LAMP 2.0, including errors below 2 m over multiple kilometers of traverse.



\subsection{Lamp 2.0 System Design} 
\begin{figure*}[t!]
    \centering
    \includegraphics[trim={0cm 0cm 0cm 1cm},clip, width=0.8\textwidth]{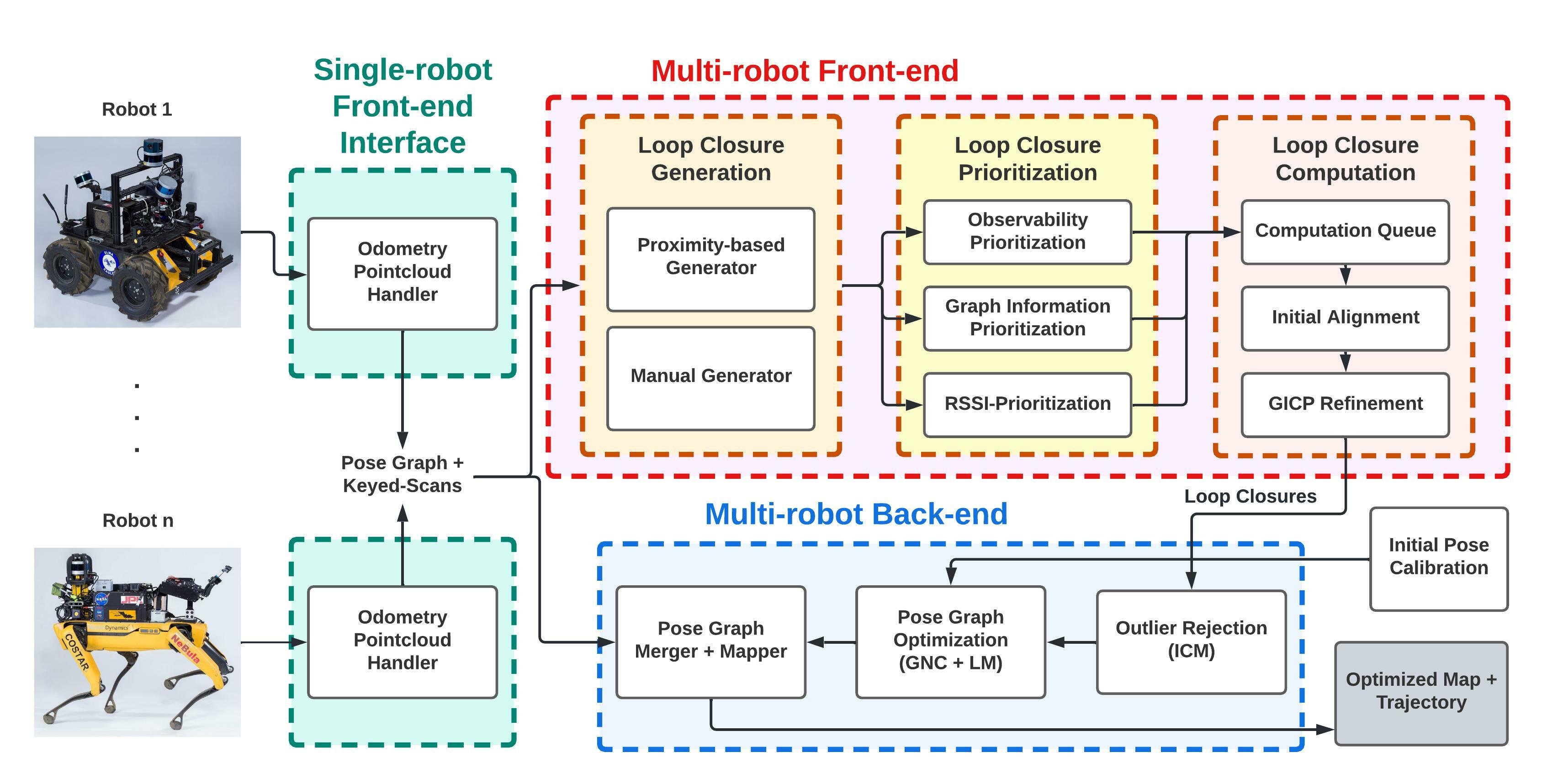}
    \caption{Architecture of LAMP 2.0, our multi-robot localization and mapping system,
    which consists of a single-robot front-end interface to produce pose graph and keyed scans input (adaptable to different odometry sources and lidar configurations) on each robot, a multi-robot front-end for loop closure detection and computation, and a multi-robot back-end for outlier-resilient pose graph optimization and map generation. ~\cite{chang2022LAMP2}.}
    \label{fig:architecture}
\end{figure*}

LAMP 2.0 is a computationally efficient and outlier-resilient centralized multi-robot SLAM system that is adaptable to different input odometry sources. 
Fig.~\ref{fig:architecture} provides an overview of our system architecture.
The system consists of (i) a \emph{single-robot front-end} on each robot that takes in any odometry input (e.g. HeRO) to build a pose-graph consisting of odometry edges and nodes. Each robot then sends each new edge and node, along with an associated point cloud (called a keyed scan) to the base station, where these factors are combined to make the multi-robot pose-graph. 
(ii) a \emph{multi-robot front-end}, running on a centralized base station, which uses the multi-robot graph and keyed scans to perform multi-robot loop closure detection, and (iii) a \emph{multi-robot back-end}, that adds the intra- and inter-robot loop closures from the front-end to the multi-robot graph to perform a robust pose graph optimization for the whole team of robots.

LAMP 2.0 is an evolution from LAMP 1.0, with the main enhancements being:
\begin{enumerate}
    \item A robust and scalable loop closure detection module that is able to handle and prioritize a rapidly growing number of loop closure candidates and includes modern 3D registration techniques to improve the accuracy and robustness of the detected loop closures.
    \item An outlier-robust back-end based on Graduated Non-Convexity~\cite{yang2020graduated} for pose graph optimization.
\end{enumerate}

In the rest of this section, we highlight the core components of LAMP 2.0 that enable large-scale and robust, multi-robot localization and mapping in challenging underground environments. 

  The goal of initial pose calibration is to establish a common reference frame for all robots in the team and we achieve this goal by calibrating sensors onboard each robot before operation (both intrinsic and extrinsic calibration).  We refer the reader to LAMP section of ~\cite{agha2021nebula} for a better understanding of this process in detail.




\subsubsection{Single-Robot Front-End}\label{sec:front-end-interface}
The single-robot front-end of LAMP 2.0 is designed to be adaptable to any odometry input in the appropriate frame. This makes it simple to adapt to accept robust fusions of odometry sources (such as HeRO~\cite{hero2019isrr}), or adapt to different odometry sources, for different robots, such as from Hovermap~\cite{jones2020hovermap} or LOCUS 2.0 ~\cite{reinke2022locus}.

LAMP 2.0 utilizes a pose graph formulation~\cite{cadena2016slam} to represent the estimated robot trajectory. Each node in the graph represents an estimated robot pose and each edge linking two nodes in the graph represents the relative motion or position measurement between the pair of nodes. 
As each robot has limited memory and computation onboard, it is important to prevent the pose graph from growing too large. 
So, a sparse pose graph with new nodes are only added after the robot exceeds a certain translation threshold (2 m) and rotation threshold (30 degrees). 
Robot pose nodes are indicated as the \emph{key nodes} where each key node is associated with a \textit{keyed scan}, which is the pre-processed point cloud acquired at the corresponding time. 

Each time the graph is updated, the incremental changes are sent to the central base station. 
This system is designed, in conjunction with our communication system, to handle robots going out of and back into communications. When out of communications, the keyed scans and pose graph updates are queued. This queue is then sent as a batch to the base station when communications is regained.



\begin{figure}[t]
    \centering
    \includegraphics[width=.8\columnwidth]{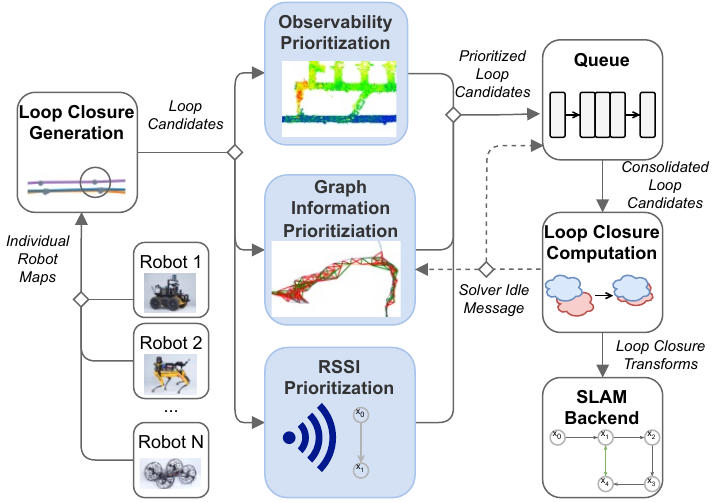}
    \caption{Diagram of SLAM multi-robot front-end and loop closure processing pipeline. Robot maps are merged on a base station and passed to the generation module, which performs proximity based analysis to generate candidate loop closures. These candidate loop closures are sorted by the prioritization modules, which feed into a queue. The loop closure solver computes the transformations between pairs of lidar scans which is given to the SLAM back-end which adds a loop closure factor.~\cite{denniston2022prioritize} \vspace{-7mm}}\label{fig:lcd_diagram}
\end{figure}

\subsubsection{Architecture of Multi-robot front-end}\label{sec:lidarLC}
The full multi-robot SLAM front-end is shown in Fig.~\ref{fig:lcd_diagram}, 
and consists of three main steps: loop closure generation, loop closure prioritization, 
and loop closure computation. These three steps are a the core part of enabling LAMP 2.0 to be scalable to many robots and large scales, both of which produce large graphs with many potential loop closures. For more details see~\cite{denniston2022prioritize}

\ph{Loop Closure Generation}

On receiving updates pose-graph and keyed scan data, the multi-robot front-end triggers a search for loop closure candidates - pairs of nodes with associated keyed scans. There are numerous ways to generate potential loop closures, such as place recognition~\cite{cattaneo2022lcdnet,schaupp2019OREOSOR,cop2018DelightAE}, junction detection~\cite{DARE-SLAM}, or artificial beacons~\cite{funabiki2020uwb}, or geometric proximity. While our architecture is adaptable to any of these approaches, we use adaptive proximity and artificial radio beacons in our experiments. The adaptive proximity approach generates loop candidates from nodes that lie withing a certain distance $d$ from the most recent node in the pose graph; $d$ is adaptive and is defined as 
$d = \alpha|n_{current} - n_{candidate}|$, which is dependent on the relative traversal between two nodes for the single-robot case and 
$d = \alpha n_{current}$, which is dependent on the absolute traversal for the multi-robot case, 
where $n_{current}$ and $n_{candidate}$ are the index of the current and candidate nodes respectively and $\alpha$ is a constant ($0.2m$).

For generating loop candidates from artificial radio beacons, we utilize the systems deployed communication nodes. These radio beacons are assumed static when deployed, and provide a Returned Signal Strength Indicator (RSSI) for all pairs in the network that can be used as a pseudo-range. If the RSSI between a robot and a deployed, static node exceeds a threshold indicating close proximity, that robot's nearby nodes are associated with that radio beacon. These nodes are then matched with those of any robot that passes the radio beacon in the future to produce a set of loop closure candidates. 


\ph{Loop Closure Prioritization} 

With many robots, large scales of exploration and the desire for high localization accuracy, there are many loop closure candidates generated. So many, that it is infeasible to perform GICP on them all within the mission time limit. Hence, there is a need to prioritize the candidates to maximize the localization benefit within a given time and computational allocation. To select the best loop closures to process we have three independent modules that orders the loop closures to process in batches.
The \emph{Graph Information Prioritization} module uses a graph neural network to determine which subset of loop closure candidates would decrease the error the most if they were added to the graph.
The \emph{Observability Prioritization} module prioritizes loop closure candidates with point clouds that have many geometric features and are particularly suitable for ICP-based alignment.
The \emph{RSSI Prioritization} module attempts to find loop closures which are close to a known radio beacon.

Each prioritization module sends a set of its top priority candidates, which are then merged into a round-robin queue.
Once these loop closure candidates are ordered, they are stored until the loop closure computation module is done computing the transforms of the previous set of loop closures. 
Each time the loop closure computation node is done processing the previous set, the queue sends another fixed-size set of loop closures for computation. 
This architecture allows the generation and computation rates to be independent, and for the loop closure computation module (the computation bottleneck) to be saturated as often as possible.

\ph{Loop Closure Computation} 

Every loop candidate is evaluated with a point cloud alignment process which both computes the transform and assesses the fitness of alignment. This alignment process proceeds in two stages. First we use SAmple Consensus Initial Alignment (SAC-IA)~\cite{rusu2009FPFH} or TEASER++~\cite{yang2020teaser} to first find an initial transformation. Then we perform Generalized Iterative Closest Points (GICP)~\cite{segal2009gicp} to refine the transformation.
We discard the loop closures that either have a mean error that exceeds some maximum threshold after SAC-IA ($32 m$ in our experiments) 
or have a mean error that exceeds some maximum threshold after GICP ($0.9 m$ in our experiments).
The remaining loop closures and associated transforms are then sent to the back-end,
and the computation module requests more loop closures from the prioritization queue.

To compute the loop closures in an efficient and scalable manner, we use an 
adaptable-size thread pool to perform computation in parallel across the current consolidated loop closure candidates.

The most computationally expensive step in this system is the loop closure computation step, hence the prioritization approach is designed to achieve maximum benefit from computation expended in loop computation, by only processing the loop closure candidates most likely to succeed (\textbf{Observability}, \textbf{RSSI}), and improve localization (\textbf{Graph Information}).
We refer the readers to~\cite{denniston2022prioritize} for more in depth presentation and analysis of this module.

\subsubsection{Multi-Robot Back-End}\label{sec:PGO}

The final, yet critical, part of LAMP 2.0 is the robust pose-graph optimization, to reliably generate a globally consistent and drift-free 3D map of the environment. This back-end combines the multi-robot pose-graph with the passing loop closure transforms and performs robust pose-graph optimization (PGO) to produce a final optimization trajectory. Robustness to outliers is essential in the PGO step, because otherwise there is the potential for just one false loop to cause corruption in the localization for all agents.

After optimizing the graph, the global map is generated by transforming the keyed scans to the global frame using the optimized trajectory.

\ph{Outlier-robust Pose Graph Optimization} 

To safeguard against erroneous loop closures, 
our multi-robot back-end includes two outlier rejection options:
Incremental Consistency Maximization (ICM)~\cite{LAMP},
which checks detected loop closures for consistency with each other and the odometry before they are added to the pose graph,
and Graduated Non-Convexity (GNC)~\cite{yang2020graduated}, 
which is used in conjunction with Levenberg-Marquardt 
to perform an outlier-robust pose graph optimization to obtain
both the trajectory estimates and inlier/outlier decisions on the loop closure not discarded by ICM.

\subsection{Representative Experiments}
\begin{figure}[t]
\centering
\subfloat[Loop closures verified over time]{\includegraphics[trim=20 0 30 35, clip, width=0.49\columnwidth]{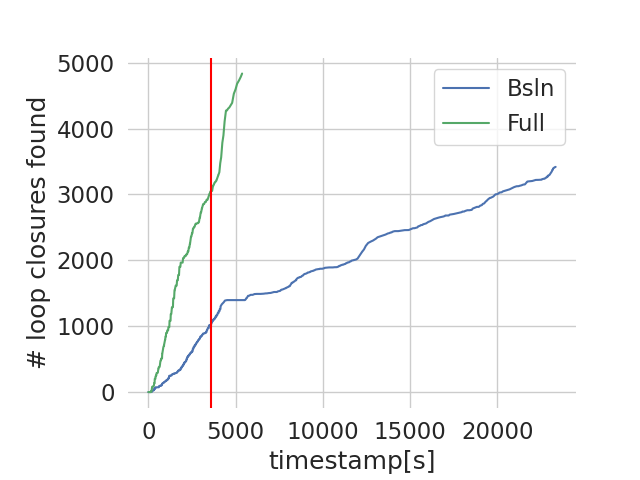}}
\hfill
\subfloat[Average trajectory error over time]{\includegraphics[trim=20 0 30 35, clip, width=0.49\columnwidth]{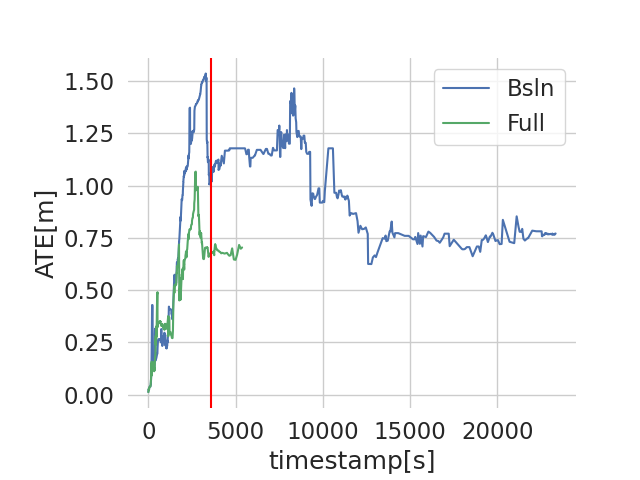}}
\caption{Direct comparison of with and without prioritization on the Tunnel dataset. The red line marks the end of the mission. The full system is able to find a similar number of inlier loop closures and a lower error at the end of the mission than the baseline can after computing all loop closures.}
\label{fig:prioritization-exp}
\end{figure}

\ph{Loop Closure Prioritization} The benefits of loop closure prioritization is demonstrated in Fig.~\ref{fig:prioritization-exp}.
For this experiment, we allowed the system to run past the duration of the mission (1 hour)
and plot the number of loop closures detected along with the trajectory error. With prioritization, we were able to detect more loop closures in less than one-quarter of
the time compared to without prioritization. 
The trajectory error is also reduced earlier due to the earlier detection of the loop closures, giving us a better trajectory estimate before the conclusion of the mission.

\begin{figure}
\centering
\subfloat[Tunnel ATE]{\includegraphics[trim={25 5 35 40}, clip, width=0.24\columnwidth]{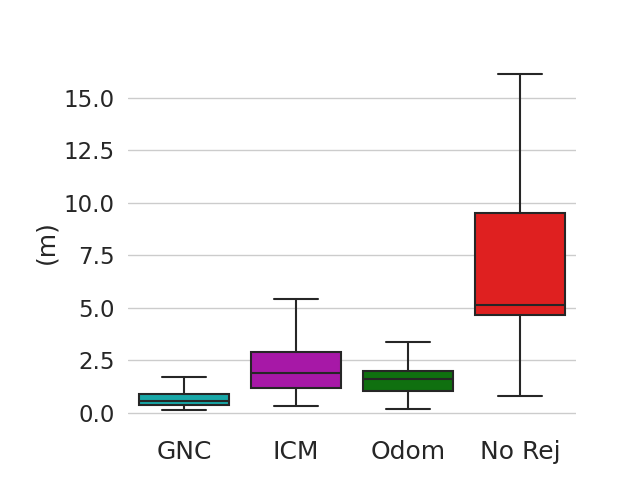}}
\hfill
\subfloat[Urban ATE]{\includegraphics[trim={25 5 35 40}, clip, width=0.24\columnwidth]{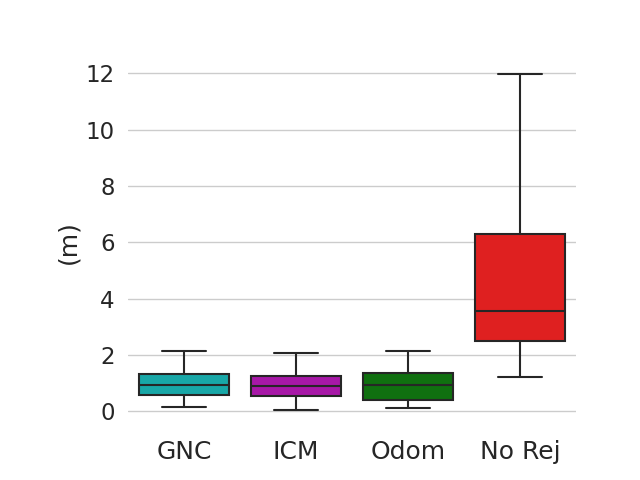}}
\hfill
\subfloat[Finals ATE]{\includegraphics[trim={25 5 35 40}, clip, width=0.24\columnwidth]{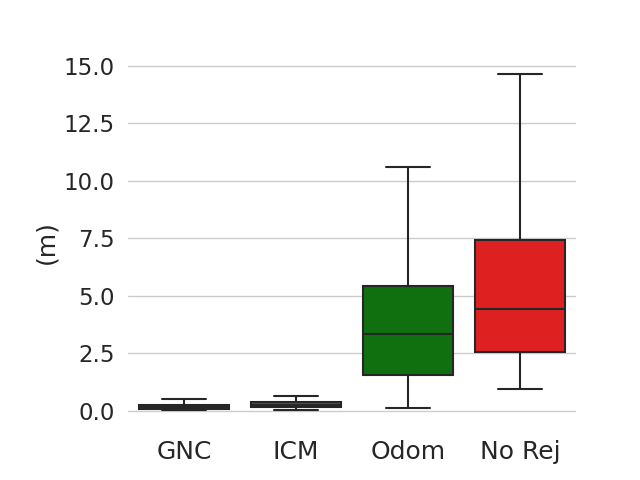}}
\hfill
\subfloat[Limestone Mine ATE]{\includegraphics[trim={25 5 35 40}, clip, width=0.24\columnwidth]{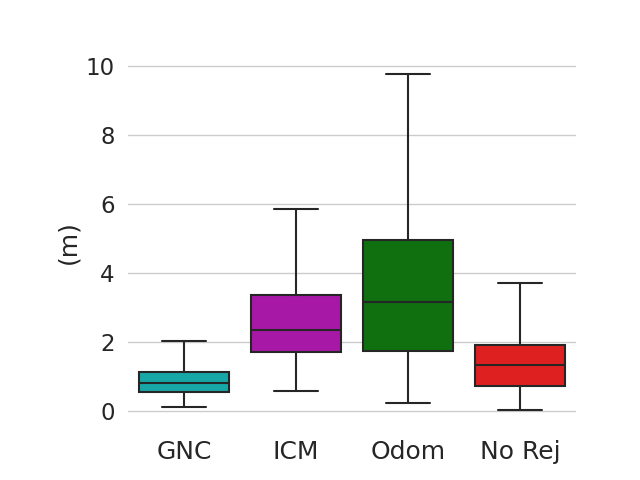}}
\caption{Comparison of trajectory ATE (across the multi-robot trajectory) for ICM and GNC compared to the cases with no loop closure or no outlier rejection across four different environments.}
\label{fig:or-ablation-blxplt}
\end{figure}

\pr{Outlier Rejection} The improvements in the capabilities of our outlier-robust pose graph optimization module is showcased with an ablation study in Fig.~\ref{fig:or-ablation-blxplt}. We compare the average final trajectory error
with GNC or ICM~\cite{LAMP} as an outlier rejection method; we also report trajectory errors for the case when no outlier rejection is performed (``No Rej'') and when no loop closure is detected (``Odom''). It is clear that no rejection mechanism leads to unacceptably large errors in almost all cases. While both GNC and ICM largely improve the trajectory estimate relative to the no loop closure or no outlier rejection case, GNC in general is more robust to outliers and gives better trajectory estimates, as especially evident in the longer Tunnel and Limestone Mine datasets where ICM failed to reject some of the outliers in the first and was too conservative in the other.

\begin{figure}[t]
\centering
\subfloat[Single robot input]{\includegraphics[trim={0 50 0 60}, clip, width=0.45\columnwidth]{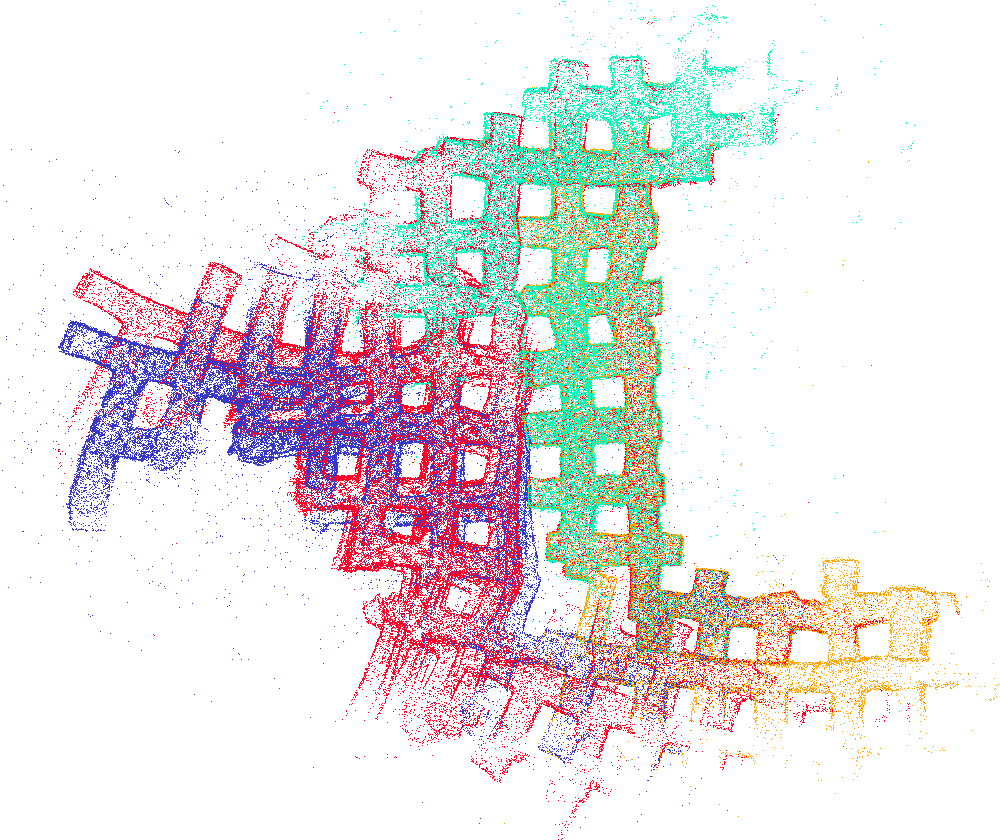}}
\hfill
\subfloat[LAMP 2.0 output]{\includegraphics[trim={0 50 0 60}, clip, width=0.45\columnwidth]{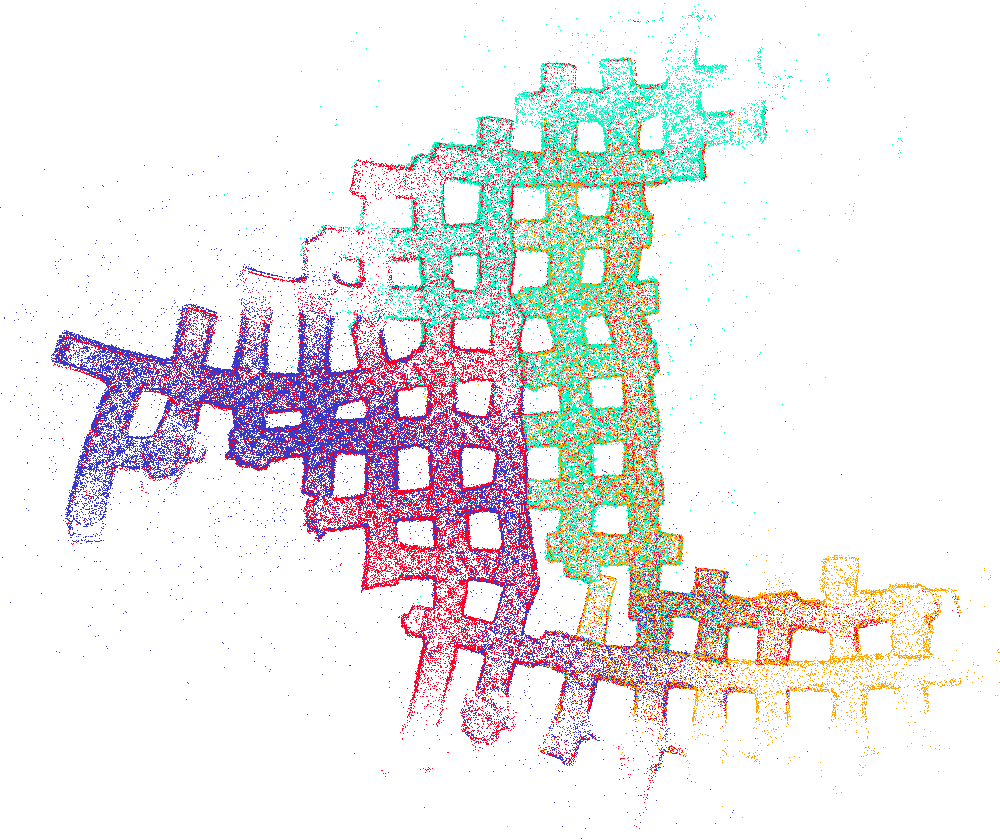}}
\caption{Single-robot map (no loop closures) compared with LAMP 2.0 map (with inter- and intra- robot loop closures) in the Kentucky Underground. }
\label{fig:ku_comparison}
\end{figure}

\pr{Multi-robot System Evaluation}
In order to demonstrate the end-to-end results of the full LAMP 2.0 system, we play back the data in real-time to the base-station and the data is processed in the same order and rate as it was generated in the field.
In Table~\ref{tab:full-comparison}, we show the improvement in average trajectory error 
of LAMP 2.0 compared to LAMP 1.0 for each robot on the four datasets,
LAMP 2.0 achieves minimal errors (below 2 m) 
with trajectory lengths of up to 2.2 km.

Fig.~\ref{fig:ku_comparison} illustrates the impact of inter- and intra- robot loop closures, showcasing how much localization benefit there is from combining measurements between a team of robots (in this case 4 robots).

\begin{figure*}[t!]
\centering
\includegraphics[trim=40 20 20 20, clip, width=1.0\textwidth]{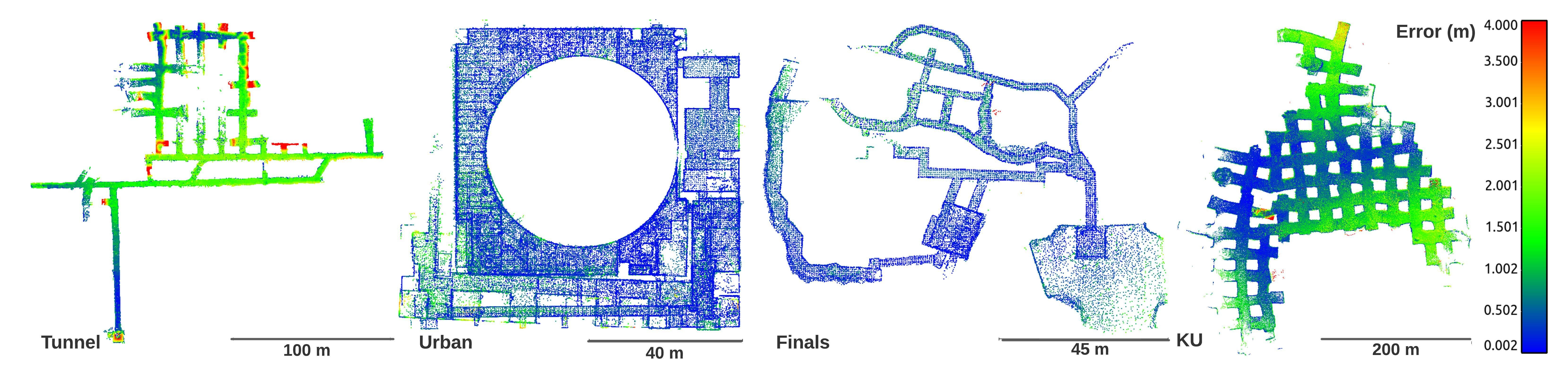}
\caption{Final map error in meters of the DARPA Subterranean Challenge finals course for the 
Tunnel, Urban, Finals, and Kentucky Underground (KU) datasets~\cite{chang2022LAMP2}. }
\label{fig:mapping-results}
\end{figure*}

Finally, we present the full multi-robot mapping results across four datasets through the LAMP 2.0 global point cloud map colored by the cloud-to-cloud error
against the ground truth map in Fig.~\ref{fig:mapping-results}. We are able to achieve map errors below 4 m in each of these large-scale, challenging environments with varying cross sections, and frequency of true loop closures.

\setlength{\tabcolsep}{2pt}
\begin{table}[t!]
\centering
\caption{Comparison of LAMP 2.0 against LAMP 1.0 }\label{tab:full-comparison}
\begin{tabular}{cl cccc}
\toprule
& \multirow{ 2}{*}{Robot} & Traversed  & LAMP 2.0 & Single Robot & LAMP 1.0  \\
&  &  [m] & ATE [m] & ATE [m] & ATE[m] \\
\midrule
\multirow{2}{*}{Tunnel}
& husky3 & 1194 & \textbf{0.65} & 0.83 & 1.07 \\
& husky4 & 1362 & 0.72 & \textbf{0.63} & 1.44 \\
\midrule
\multirow{3}{*}{Urban}
& husky1 & 612 & \textbf{0.79} & 0.87 & 0.95 \\
& husky4 & 416 & \textbf{0.76} & 0.79 & 0.78 \\
& spot1 & 502 & 1.31 & 1.46 & \textbf{0.99} \\
\midrule
\multirow{4}{*}{KU}
& husky1 & 2204 & 1.01 & \textbf{0.9} & 5.34 \\
& husky2 & 1526 & \textbf{0.71} & 0.75 & 2.85 \\
& husky3 & 1678 & \textbf{1.31} & 1.33 & 2.11 \\
& husky4 & 896 & \textbf{0.69} & 0.86 & 5.49 \\
\midrule
\multirow{4}{*}{Final}
& husky3 & 72 & \textbf{0.16} & 0.16 & 0.56 \\
& spot1 & 430 & \textbf{0.2} & 0.37 & 0.37 \\
& spot3 & 484 & \textbf{0.21} & 0.38 & 0.55 \\
& spot4 & 238 & \textbf{0.15} & 0.23 & 0.62 \\
\end{tabular}
\end{table}

%% file: sections/4.artifacts.tex
\usetikzlibrary{positioning}
\section{Semantic Understanding and Artifact Detection} \label{sec:artifacts}


Semantic object mapping (SOM), that is the detection and localization of objects of interest using robots' sensing systems in a given environment, is a critical component in the NeBula autonomy framework. The requirements of SOM and our prior approach to it in the SubT challenge is discussed in \cite{agha2021nebula}. Successful SOM within the context of the SubT challenge requires high recall to avoid missing true target objects and high precision to avoid wasting valuable operational time on false positives. Achieving such requirements demands that several constraints be adhered to, including (i) mission time, (ii) computational capacity, (iii) mesh network bandwidth, (iv) size, weight and power (SWaP) limitations, (v) human cognitive load. 
This section discusses NeBula's new approach to SOM - the Early Recall, Late Precision (EaRLaP) pipeline. We focus on static objects, with visual, thermal, or depth signatures. The pipeline and its performance are explained in more detail in ~\cite{lei2022earlap}. We discuss spatially diffuse signal localization in \ref{art:signal_loc}.

\subsection{EaRLaP System Design} 
\begin{figure*}[!ht]
\centering
    \resizebox{\textwidth}{!}{%
\input{figures/artifacts/earlap_func}
}
  \caption{\footnotesize Hierarchical decomposition of the semantic object mapping problem. $f$ is divided into three sub-functions and these sub functions are further divided to reach a composition of seven sub-functions, which is the EaRLaP semantic object mapping pipeline.}
  \label{fig:earlap_decomposition}
\end{figure*}
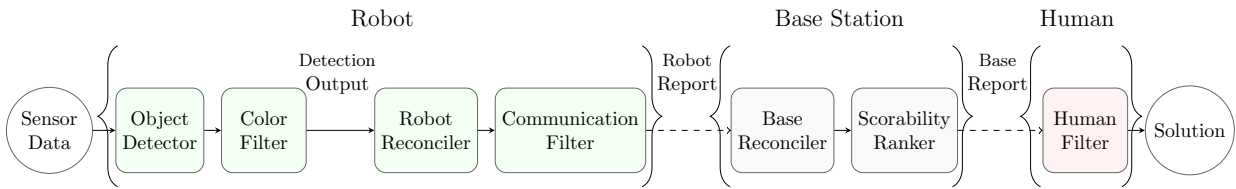

The detection and localization of a set of objects using visual color or thermal signatures are broken down into seven sub-functions, as shown in Fig.~\ref{fig:earlap_decomposition}.

\ph{Object Detector} NeBula robots are equipped with multiple cameras to cover a near $360^{\circ}$ field of view (FOV), generating multiple high-rate image streams. Each stream is processed by convolutional neural network (CNN) object detectors at a low rate on robot processors with limited computational power. As the object detectors cannot keep up with the high rate of the image streams, an image selection filter selects the least blurry images for detection to ensure that computational power is used effectively. A pruned YOLOv5m6 model handles high-resolution images to detect small objects.

\ph{Handcrafted-features Filter} To remove obvious false positive detections of objects with known distinctive features, such as color or size, we handcrafted basic filters to discard detections that do not respect such class-distinctive properties (i.e. a human survivor cannot be taller than 2.5 meters).

\ph{Robot Reconciler} The global position of a given robot relative to the calibration gate at its starting point in the environment is estimated by other system modules \cite{LAMP, palieri2020locus}, however the detected objects still need to be reconciled and localized relative to the robot. In order to initially filter out repeated detections, those which have accumulated during periods in which the robot has not moved beyond set translation and rotation thresholds are eliminated. The remaining detections must then be either reconciled with previously mapped object candidates using spatial proximity criteria or new such candidates must be created.  The primary means used for robot-relative object position localization are bearing estimates \cite{dellaert2012factor}, but the range from the robot is further refined by averaging LiDAR point distances within the bounding box in the camera frame. Once the object range has been estimated, more false positives can be filtered out by comparing the known object size to the estimated object size from the bounding box at that range. 

\ph{Communications Filter} To limit the quantity of reports sent from the robots to the base station, we only transmit reports in which multiple observations have been reconciled. Additionally, we favor first sending reports for which the median report confidence of all reconciled observations is high, and localization uncertainty is low. This maximizes our chances of sending true positive reports if the communication network is saturated and not all reports can be sent back to the base station.

\ph{Base Reconciler} When multiple robots communicate to the base station, they could report the same objects if they explore the same areas. We can merge these detections into a unique object. A report is grouped to an existing report cluster if it is spatially closer than $d^b_{\textrm{min}}$ to an existing report in that cluster of the same object class. Such distance-based reconciliation assumes that objects of interest are sparsely distributed in the environment. 

\ph{Scorability Ranker} In order to mitigate against a time-pressured operator missing true positives while evaluating the reconciled reports on the base station, we rank the reports in terms of their scoring potential. We introduce the concept of ``scorability" which combines metrics from YOLO detection, color and size likelihood, and recurrence of an object. The reports are ranked with respect to scorability in descending order at the base station when presented to the human operator. This maximizes the chance of the operator seeing and scoring all of the true positive reports, even if they cannot review all of the total reports given the time constraints.
 
\ph{Human Filter} A graphical user interface (GUI) was developed that allows the human operator to control robotic exploration behavior and interact with artifact reports. The operator can reject a report, edit its estimated location or class, submit it, or save it for later.  

\begin{figure}[!t]
  \centering
  \includegraphics[width=0.7\linewidth]{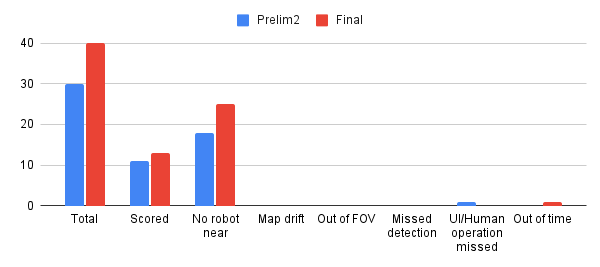}
  \caption{Scoring break down of the Final and Prelim Run2. Additional description: "No robot near": outside of the field-of-view (FOV) of the cameras. }
  \label{fig:art_result}
\end{figure}

\begin{figure}[!t]
  \centering
  \includegraphics[width=0.4\linewidth]{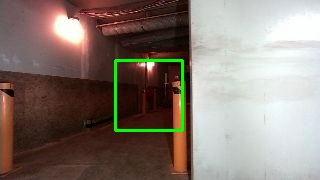}
  \caption{This detected backpack was reported in the base station but was rejected by the operator}
  \label{fig:missing_bp}
\end{figure}

\ph{Performance} EaRLaP maintained high recall under the constraints and pruned false positives effectively to allow the human operator to achieve high precision within the time limit. EaRLaP detected and located all the objects within the rovers' reach. The filtered, reconciled and ranked reports helped the human operator to validate and report most of the reports within the mission time (see Fig.~\ref{fig:art_result}). The one missing report noted as "UI/Human operation missed" in  Fig.~\ref{fig:art_result}), which is a backpack, was due to GUI/operator interface issues. The detected backpack was naturally hard to identify, especially when the operator was under pressure due to time constraints. The backpack was hanging on a doorknob and occluded by a yellow pole from the angle of the captured images in dim lighting under high contrast. The object was detected in only 2 frames (see Fig.~\ref{fig:missing_bp}). The 4 selected best images to represent the detection are duplicates of the same image. The enhancement tool did not enhance the image because the image’s intensity was not low enough due to the presence of large white walls in the image. The cause of the “out of time” missing report was that the hour-long duration of the final round of the competition had already expired when the report was submitted.

\subsection{Spatially Diffuse Localization}
\label{art:signal_loc}
Localizing spatially diffuse signals such as radio or sound is another component of NeBula's detection framework. We focus on radio and audio-based signals assisting human operators in high-level decisions.

\ph{Bluetooth and Wi-Fi Detection}
Bluetooth and WiFi signal strength (RSSI) over the threshold are independently recorded with the robot position, along with a model including a damping factor for wall and corner attenuation. These multi-modal signals are shown on a global 3D semantic map along with peak signal strength and are used for cross-validation. 

\ph{Audio Detection} 
Given short segments of audio from our robots, a way to augment tracking ability is to classify whether aural cues are detected. The robots use a spectrogram to represent the amplitude of each frequency at each time throughout a temporal window. Distinctive rhythmic patterns of the human voice are then recognized and a detection threshold is applied.

%% file: figures/artifacts/earlap_func.tex
\begin{tikzpicture}[
robotnode/.style={rounded corners=0.2cm, draw=black!60, fill=green!5, , minimum size=15mm, align=center},
basenode/.style={rounded corners=0.2cm, draw=black!60, fill=gray!5, , minimum size=15mm, align=center},
humannode/.style={rounded corners=0.2cm, draw=black!60, fill=red!5, , minimum size=15mm, align=center},
topnode/.style={rounded corners=0.2cm, draw=black!60, fill=red!0, , minimum size=15mm, align=center},
squarednode/.style={circle, draw=black!60, fill=red!0, , minimum size=10mm, align=center},
]
\node[squarednode,draw]      (SensorData)         {Sensor \\ Data};
\node[robotnode,draw]        (cnn)       [right=4mm of SensorData, label=$ $] {Object \\  Detector};
\node[robotnode,draw]      (ColorFilter)       [right=3mm of cnn, label=$ $] {Color \\ Filter};
\node[robotnode,draw]        (robotrecon)       [right=12mm of ColorFilter, label=$ $] {Robot \\ Reconciler};
\node[robotnode,draw]        (comms)       [right=3mmof robotrecon, label=$ $] {Communication \\ Filter};
\node[basenode,draw]        (baseRecon)       [right=15mm of comms, label=$ $] {Base \\ Reconciler};
\node[basenode,draw]        (confidence)       [right=3mm of baseRecon, label=$ $] {Scorability \\ Ranker};
\node[humannode,draw]        (humanFilter)       [right=15mm of confidence, label=$ $] {Human \\ Filter};
\node[squarednode,draw]        (solution)       [right=3mm of humanFilter] {Solution};

\begin{scope}[yshift=3cm]
\node[color=black] at (5.9,-1){\large Robot};
\node[color=black] at (14,-1){\large Base Station};
\node[color=black] at (18.21,-1){\large Human};
\end{scope}


\draw[-stealth] (SensorData) -- (cnn);
\draw[-stealth] (cnn) -- (ColorFilter);
\draw[-stealth] (ColorFilter) -- (robotrecon);

\draw[-stealth] (robotrecon) -- (comms);
\draw[dashed,->] (comms) -- (baseRecon);
\draw[-stealth] (baseRecon) -- (confidence);
\draw[dashed,->] (confidence) -- (humanFilter);
\draw[-stealth] (humanFilter) -- (solution);

\node[align=center] at (5.1,1) {\small Detection \\ Output}; 
\node[align=center] at (11.3,1) {\small Robot \\ Report}; 
\node[align=center] at (16.8,1) {\small Base \\ Report}; 






\draw [decorate,decoration={brace,amplitude=10pt,mirror,raise=4pt},yshift=0pt](1.35,1.5) -- (1.35,-1) node [black,midway,xshift=0.8cm] {};
\draw [decorate,decoration={brace,amplitude=10pt},xshift=-0.5cm,yshift=0pt](11.0,1.5) -- (11.0,-1) node [black,midway,xshift=-0.6cm] {};

\draw [decorate,decoration={brace,amplitude=10pt,mirror,raise=4pt},yshift=0pt](12.2,1.5) -- (12.2,-1) node [black,midway,xshift=0.8cm] {};
\draw [decorate,decoration={brace,amplitude=10pt},xshift=-0.5cm,yshift=0pt](16.5,1.5) -- (16.5,-1) node [black,midway,xshift=-0.6cm] {};

\draw [decorate,decoration={brace,amplitude=10pt,mirror,raise=4pt},yshift=0pt](17.7,1.5) -- (17.7,-1) node [black,midway,xshift=0.8cm] {};
\draw [decorate,decoration={brace,amplitude=10pt},xshift=-0.5cm,yshift=0pt](19.5,1.5) -- (19.5,-1) node [black,midway,xshift=-0.6cm] {};

\end{tikzpicture}

%% file: sections/5.motion_planning.tex
\section{Risk-Aware Traversability and Motion Planning} \label{sec:traversability}
\subsection{Motivation and High-Level Overview}

A fundamental component of NeBula is its risk-aware traversability and motion planning. This component, which we call STEP (Stochastic Traversability Evaluation and Planning), allows the robots to safely traverse extreme and challenging terrains by quantifying uncertainty and risk associated with various elements of the terrain. Please see \cite{agha2021nebula} for prior work on NeBula, which is extended and improved here.  Additionally, see \cite{thakkur2020} and \cite{fan2021step} for detailed discussion of traversability and motion planning components within our approach.  We first give a high level overview of STEP and it's role within NeBula, then discuss in more detail our recent extensions and improvements.

\begin{figure*}[ht]
    \centering
    \includegraphics[width=\linewidth]{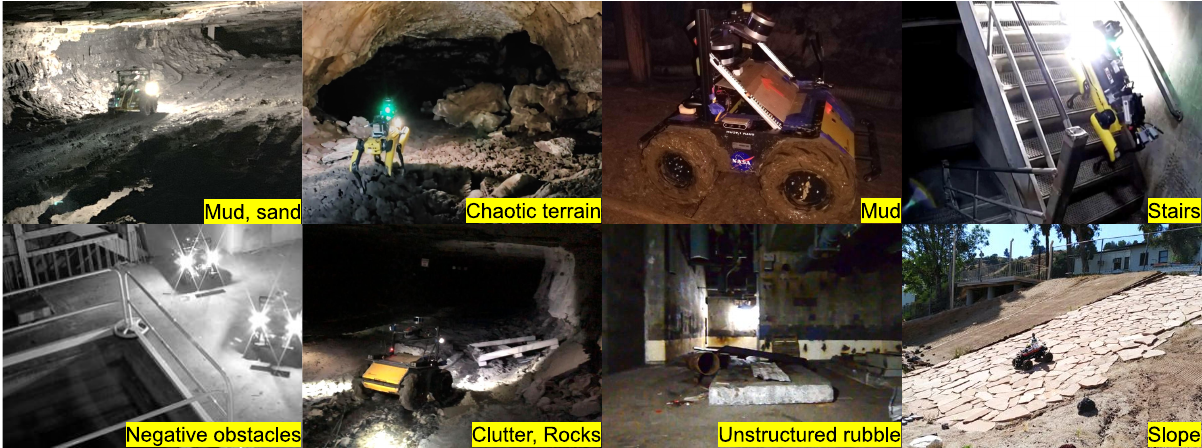}
    \caption{Mobility-stressing elements commonly found during testing in various tunnel, urban, and cave environments, including 800ft underground at Arch Mine in Beckley, WV, Valentine Cave at Lava Beds National Monument, CA, Satsop power plant in Elma, WA, and Mars Yard at JPL, Pasadena, CA.}
    \label{fig:TravTerrain}
\end{figure*}

\pr{Challenges in extreme terrain motion planning}
Unstructured obstacle-laden environments pose large challenges for ground roving vehicles with a variety of mobility-stressing elements. Common assumptions of a benign world with flat ground and clearly identifiable obstacles do not hold; Environments introduce high risks to robot operations, containing difficult geometries (e.g. rubble, slopes) and non-forgiving hazards (e.g. large drops, sharp rocks) \cite{kalita2018path,leveille2010lava}. Additionally, subterranean environments pose unique challenges, such as overhangs, extremely narrow passages, etc. See Fig.~\ref{fig:TravTerrain} for representative terrain features. Determining where the robot may safely travel has several key challenges: (\textit{i}) Localization error severely affects how sensor measurements are accumulated to generate dense maps of the environment. (\textit{ii})
 Sensor noise, sparsity, and occlusion induces large biases and uncertainty in mapping and analysis of traversability. (\textit{iii}) The combination of various risk regions in the environment create highly complex constraints on the motion of the robot, which are compounded by the kinodynamic constraints of the robot itself.

\begin{figure*}[h!]
    \vspace{-5mm}
    \centering
    \includegraphics[width=\linewidth]{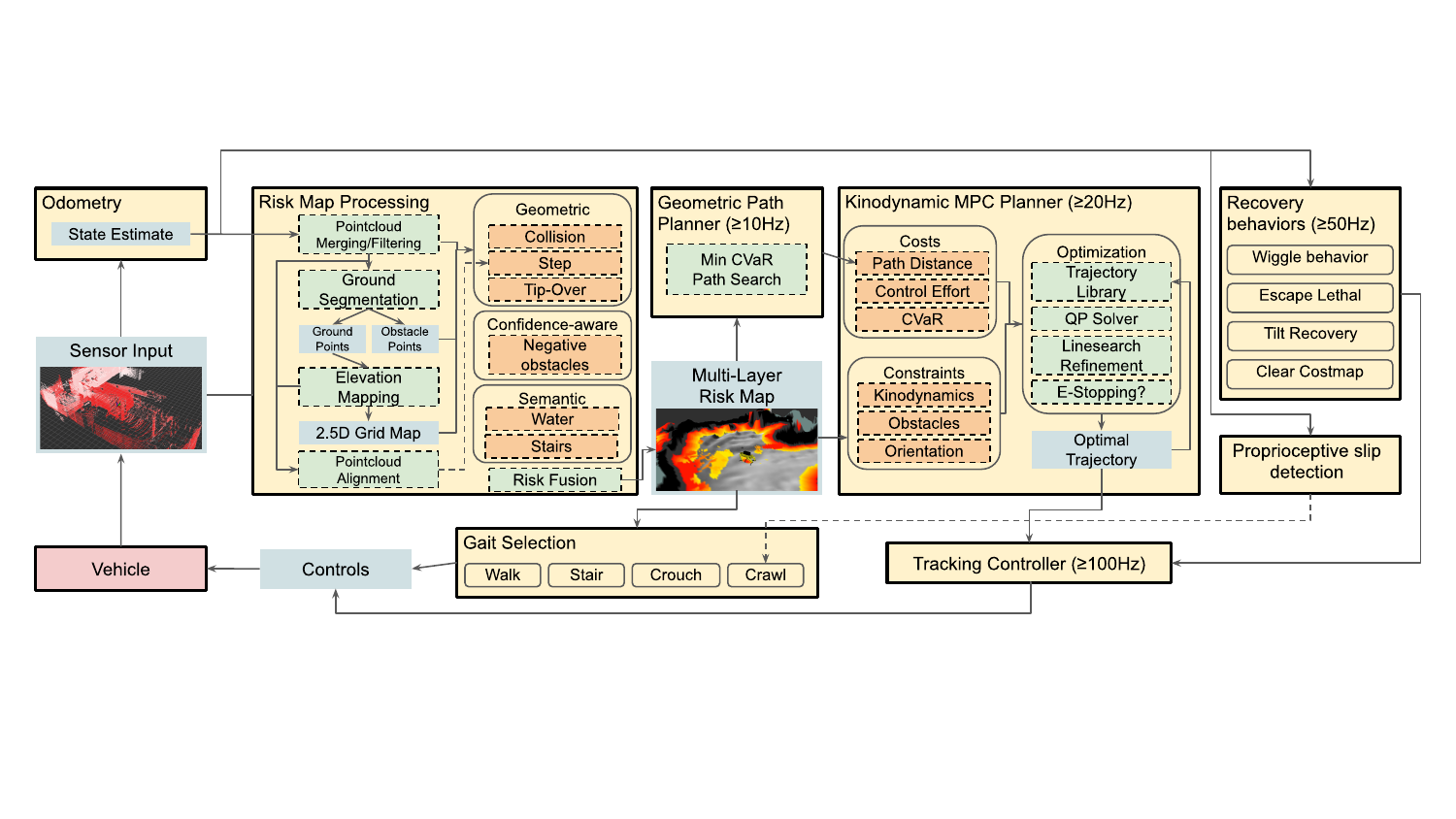}
     \vspace{-20mm}
    \caption{Traversability and Motion Planning Architecture Overview.
    }
    \label{fig:TravOverview}
\end{figure*}

\pr{System Architecture}
To address these issues, we develop a risk-aware traversability analysis and motion planning method, which 1) assesses the traversability of terrains at different fidelity levels based on the quality of perception, 2) encodes the confidence of traversability assessment in its map representation, and 3) plans kinodynamically feasible paths while considering mobility risks. Fig.~\ref{fig:TravOverview} shows an overview of the local motion planning approach. The sensor input (pointcloud) and odometry is sent to the risk analysis module, evaluating the traversability risk with its estimation confidence. The generated risk map is used by hierarchical planners consisting of a geometric path planner and a kinodynamic MPC (Model Predictive Control) planner. The planners replan at a higher rate to react to the sudden changes in the risk map. The planned trajectory is executed with a tracking controller, which sends a velocity command to the platform.

\subsection{Key Features of STEP within NeBula}

Key features of STEP and our approach to traversability and motion planning within NeBula include:

\pr{Robot agnosticism}
Our approach is highly extensible and general to our different ground robot types.  We are able to specify a wide array of constraints and costs, such as limiting pitch or roll of the vehicle on slopes, preferring one direction of motion, keeping some distance from obstacles, fitting through narrow passages, or slowing down / stopping around risky areas.  This flexibility has proven important in achieving robust navigation across the extreme traversability challenges encountered in highly unstructured environments.

\pr{Uncertainty-aware Traversability}
A key idea of STEP is to incorporate uncertainty-awareness into our mapping for traversability and motion planning. Sensor data is aggregated and used to assess the risk of traverse from various perspectives including collision, tip-over, traction loss, and negative obstacles.  Individual risk analysis is fused into a single risk value estimate (using the Conditional Value-at-Risk, i.e. CVaR, and sent to the planning module). This allows us to dynamically change the total conservativeness of the planner by changing a single parameter - the desired risk-level.

\pr{Semantic traversability factors}
In addition to geometric traversability analyses, we can also identify certain terrain features semantically and incorporate them into the risk map.  This information is useful for identifying non-geometric mobility risks, as well as notifying our planners to approach certain hazards differently (e.g. walking down stairs, presence of mud or water). 

\pr{Efficient Risk-aware Kinodynamic Planning}
Using the computed CVaR metric values on the map, we must search for a path which minimizes these values. This is done in a two-stage hierarchical fashion. The first stage operates on longer distances and takes into account positional risk (geometric planning).  The second stage operates on shorter distances and searches for a kinodynamically feasible trajectory that minimizes CVaR, while maintaining satisfaction of various constraints including obstacles, dynamics, orientation, and control effort (Model Predictive control strategy). Once a trajectory is optimized, it is sent to an underlying tracking controller for execution on the platform. For the quadrupedal robot (Boston Dynamics Spot), this also includes desired gait, such as, walk, crouch, crawl, and stair gaits.

\pr{Recovery Behaviors}
In the real world, failures are unavoidable.  As a last line-of-defense, we design behaviors to recover the system from non-fatal failures.  These behaviors including clearing/resetting the local traversability map, increasing the allowable threshold of risk (to try to escape an untraversable area), and moving the robot in an open-loop fashion towards the direction of maximum known free space.

\pr{Learning and Adaptation}
Over the course of a mission we often see changes in vehicle dynamics or environmental factors.  To adapt to these changes we employ learning-based methods \rev{using Gaussian processes} which adapt critical vehicle parameters and dynamics models based on the past history of performance \cite{fan2020bayesian} \cite{fan2020deep}. These methods ensure safety and robustness even in light of changing dynamics models.

\subsection{Augmenting STEP for Large-Scale Environments}

For robots operating over large-scale environments and large distances, reliability is of utmost importance.  In an effort to improve the health and lifespan of robots operating with the NeBula framework, we focus on reducing the number of catastrophic or fatal events which arise from interaction with challenging terrain.  Here we present a method for robustifying motion planning by incorporating proprioceptive slip detection, which allows the robot to adapt to challenging terrain by adjusting its gait or velocity when encountering anomalous terrain.

\subsubsection{Proprioceptive Slip Detection}
We leverage the proprioceptive states of the robot to design a slip-predictive model for enabling real-time detection of potential imbalance and fall events during traversability. The model predicts the probability of a slip event in advance ($p_{slip}(y_{t+n})$) based on the proprioceptive history ($\mathbf{x}_{t-k:t}$) of the robot. If the probability of a slip event is greater than a threshold, $\tau$, a conservative locomotion mode such as "crawl" or "" mode is activated for the next 60 seconds to allow better stability and fall prevention. In this work, a $\tau = 0.5$ is used. The parameter $\tau$ can be adjusted to vary the allowable detection confidence. See \cite{dey2022prepare} for more details.

\begin{figure}[!ht]
    \centering
    \includegraphics[width=0.5\linewidth]{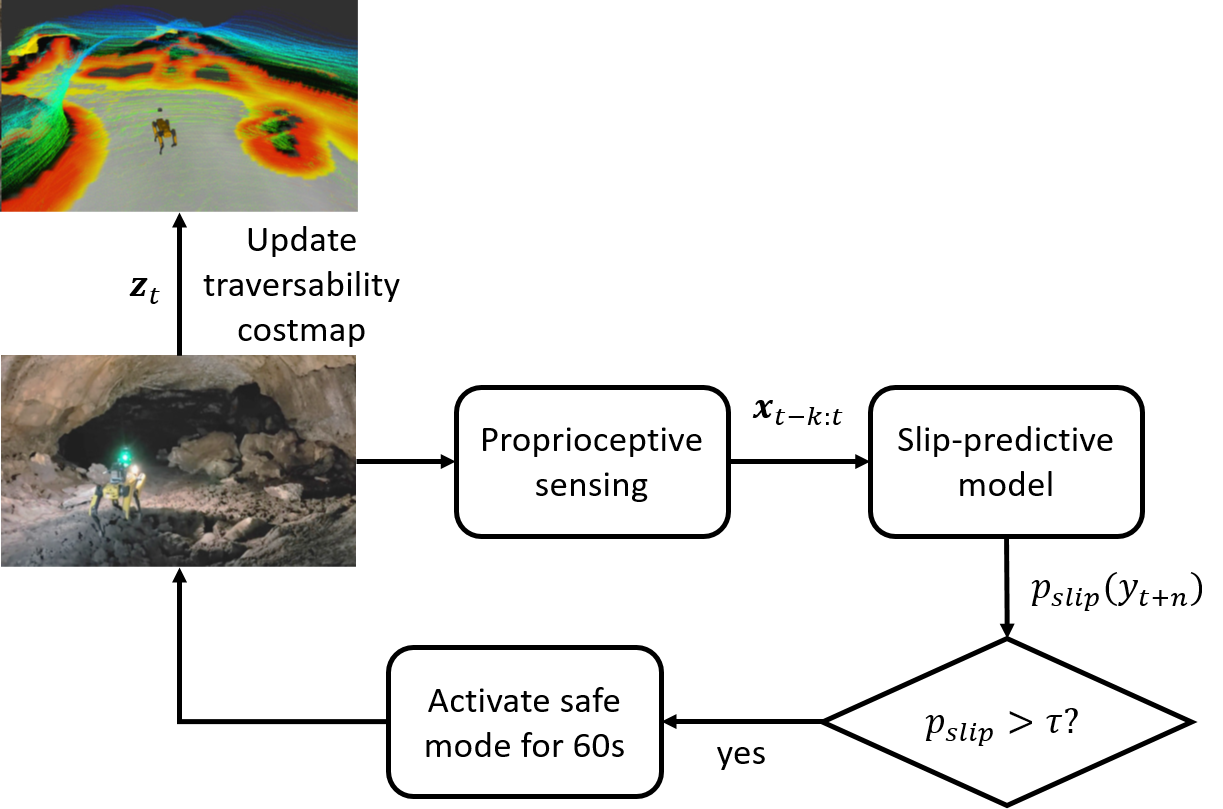}
    \caption{Overview of the slip-aware traversability architecture for legged robots. }
    \label{fig:traversability_architecture}
\end{figure}

\ph{Architecture}
Fig.~\ref{fig:traversability_architecture} shows the overview of a slip-aware traversability architecture for legged robots walking in extreme terrains. As the robot traverses through a challenging terrain, the proprioceptive state of the robot is measured using the internal position encoders and the force-torque sensors of the robot.  The robot can additionally use the slip event probability information and the information about its current position to update the traversability cost map to increase the risk of traversing through a slippery region \cite{fan2021step}. 

\ph{A Learning-based approach for slip detection}

\begin{figure*}[!h]
    \centering
    \includegraphics[width = 0.9\linewidth]{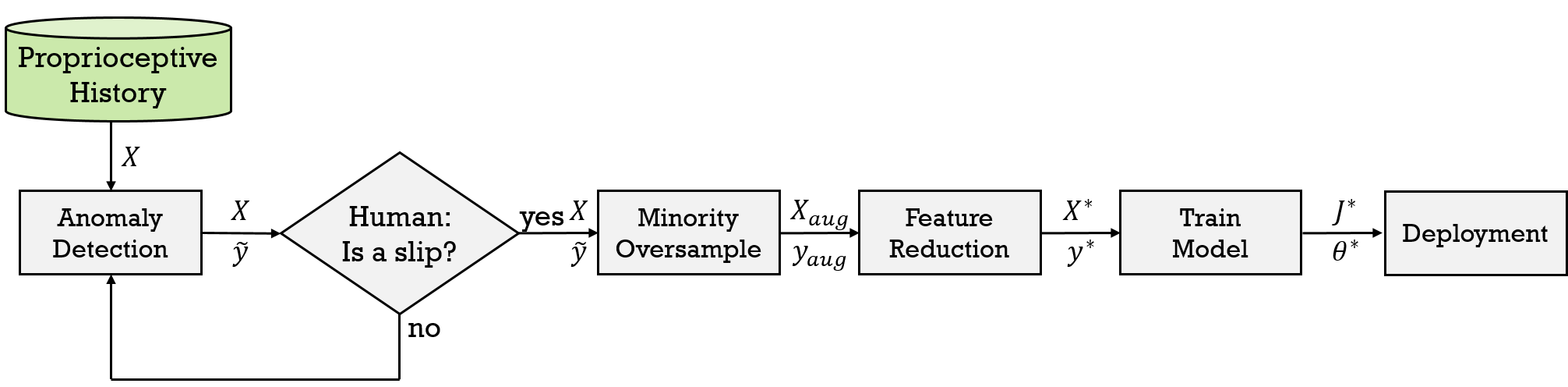}
    \caption{
    Pipeline for training the learning-based slip prediction model. First an unsupervised anomaly detection marks anomalous sequences in the robot state history. A human annotator will confirm if the event was a slip. The minority class then gets up-sampled and the features reduced. With this data, an ensemble-based decision learner model gets trained. For evaluation, the probability of slip is predicted based on a sequence of robot states. If the probability of slip is above a certain threshold a `Safe Mode' gets returned.}
    \label{fig:flow_algorithm_1}
\end{figure*}

We propose a semi-supervised learning-based approach  for developing the slip predictive pipeline. This procedure is adopted considering two cases: 1) to minimize expert annotation efforts, 2) to deal with limitations of learning-based models due to slip event rarity. Below, we concisely describe the components of the pipeline (see Fig.~\ref{fig:flow_algorithm_1}).

\ph{Anomaly detection} We use an anomaly detection algorithm, called an  \textit{isolation forest} \cite{liu2008isolation}, to detect anomalous occurrences in the data. This step reduces the human annotation space to the detected anomaly events.  
Since it is important to maximize the true positives, we apply a window of one second centered at each detected anomalous sample, so that slip events around these samples could be captured. All the frames outside these windows are considered to be normal (non-outlier). Finally we consider all the outlier frames for further human labelling.

\ph{Human feedback} Once the search space for annotating the slip events are restricted, by inspection of the robot locomotion during the outlier frames in a replay simulation, we label the frames where the robot slipped. An initial inspection of the joint state data revealed that the proprioceptive signals depicted different patterns during normal locomotion and slip and thus can be efficiently used for identifying slips.

\ph{Minority Oversampling} To reduce training bias due to data imbalance between slip and non-slip events, we adopt a minority oversampling approach to augment the the slip events to match the cardinality of the non-slip events. We use a support vector-based borderline oversampling method \cite{nguyen2011borderline} for the minority oversampling. In this oversampling method, the support vectors  learned by a support vector machine (SVM) classifier from the training data are used as the borderline between the minority and majority classes. New samples belonging to the minority class are generated along the borderline (line joining support vectors of minority class) by interpolation or extrapolation based on the class identity of the nearest neighbors. 

\ph{Feature Ablation} To reduce the feature space dimensionality, we perform a feature importance analysis using the mean decrease in impurity of each feature as the metric of importance. We find that for most of the sensor signals, information in the near history are more important than the ones further into the past. The input features that have very low importance were not considered further for fitting the models (For each proprioceptive signal, the time points which are found to be less important within a history of 60 samples is removed).

\ph{Model training}
We use an ensemble of decision learners \cite{breiman2001random} to map the proprioceptive history into a probability of slip occurrence. The slip occurrences, $y_{t+n}$ are represented at the terminal nodes or the leafs of the decision learner, 
and a set of proprioceptive histories, $\{\mathbf{x}_{t-k:t} \in \mathbb{R}^{k\times d}\}$, leading to the slip occurrences or non-occurrences represented at the intermediate nodes, starting from the root node. The decision learner splits recursively on an optimal value $\zeta$ of one of the features of the robot proprioceptive history, $u \in \mathbf{x}_{t-k:t}$ decided by the \textit{gini impurity} until the terminal nodes are reached. This builds a network of decision paths leading to a tree-structured learned model.

\ph{Field deployment} Finally, we deploy our proprioceptive slip predictive model on the Spot robot to predict slip events in real-time and autonomously switch its gait mode. If the slip probability $p_{slip}$ is above 0.5, the robot switches to a safer mode like crawl or tip-toe for the next 60 seconds. If no further slip events were detected during this window, the robot switches back to the normal mode.  We find that the slip prediction model successfully predicts slip events in real-time when deployed on the Spot robot. Correspondingly, the gait controller switches the robot to a safer locomotion mode.

%% file: sections/6.global_planning.tex
\section{Uncertainty-aware Global Planning} \label{sec:global_planning}


Autonomous global planning for environment exploration and coverage is a core part of the NeBula architecture. 
NeBula formulates the autonomous exploration in unknown environments under motion and sensing uncertainty by a Partially Observable Markov Decision Process (POMDP), one of the most general models for sequential decision making.
This formulation allows NeBula to jointly consider sequential outcomes of perception and control at the planning phase in order to achieve higher levels of resiliency during the mission operation.
In this section, we discuss our POMDP-based global planning. 
For more details, please see \cite{kim_plgrim_icaps_2021,AutoSpot,youtubespotpaper}. 

\subsection{Problem Formulation}
\ph{POMDP Formulation}
A POMDP is described as a tuple $\langle \mathbb{S}, \mathbb{A}, \mathbb{Z}, T, O, R \rangle$, where $\mathbb{S}$ is the set of states of the robot and world, $\mathbb{A}$ and $\mathbb{Z}$ are the set of robot actions and observations, respectively \cite{KLC98,Pineau03}.
At every time step, the agent performs an action $a \in \mathbb{A}$ and receives an observation $z \in \mathbb{Z}$ resulting from the robot's perceptual interaction with the environment.
The motion model $T(s, a, s') = P(s'\,|\,s, a)$ defines the probability of being at state $s'$ after taking an action $a$ at state $s$.
The observation model $O(s, a, z) = P(z\,|\,s, a)$ is the probability of receiving observation $z$ after taking action $a$ at state $s$.
The reward function $R(s, a)$ returns the expected utility for executing action $a$ at state $s$.
In addition, a belief state $b_t \in \mathbb{B}$ at time $t$ is introduced to denote a posterior distribution over states conditioned on the initial belief $b_0$ and past action-observation sequence, i.e., $b_{t} = P(s \,|\, b_0, a_{0:t-1}, z_{1:t})$.
The optimal policy $\pi^* \! : \mathbb{B} \to \mathbb{A}$ of a POMDP for a finite receding horizon is defined as follows:
\begin{align}
  \pi_{t:t+T}^*(b) &= \argmax_{\pi \in \Pi_{t:t+T}} \, \mathbb{E} \sum_{t'=t}^{t+T} \gamma^{t'-t} r(b_{t'}, \pi(b_{t'})),
  \label{eq:receding_objective_function}
\end{align}
where $\gamma \in (0, 1]$ is a discount factor for the future rewards, and $r(b,a)=\int_s R(s,a)b(s)\mathrm{d}s$ denotes a belief reward which is the expected reward of taking action $a$ at belief $b$.
$T$ is a finite planning horizon for a planning episode at time $t$.
Given the policy for the last planning episode, only a part of the optimal policy, $\pi^*_{t:t+\Delta t}$ for $\Delta t \in (0, T]$, will be executed at run time. A new planning episode will start at time $t+\Delta t$ given the updated belief $b_{t+\Delta t}$. %
The computational complexity of a POMDP grows exponentially with the planning horizon \cite{Pineau03}, and we tackle this challenge with hierarchical belief space representation and planning as to be detailed in Fig.~\ref{sec:plgrim}.

\ph{Application to Simultaneous Mapping and Planning (SMAP)}
To formalize our SMAP problem as a POMDP, we define the state $s = (q, W)$ as a pair of robot $q$ and world state $W$.
We further decompose the world state as $W = (W_{occ}, W_{cov})$ where $W_{occ}$ and $W_{cov}$ describe the occupancy and the coverage states of the world, respectively associated with their uncertainties (e.g., \cite{CRM}).
A reward function for coverage can be defined as a function of information gain $I$ and action cost $C$ as follows:
\begin{align}
  R(s, a) = \mathrm{fn}(I(W_{cov}, z),\; C(W_{occ}, q, a)),
  \label{eq:coverage_reward}
\end{align}
where $I(W_{cov}, z) = H(W_{cov}) - H(W_{cov} \,|\, z)$ is quantified as reduction of the entropy $H$ in $W_{cov}$ after observation $z$,  %
and $C(W_{occ}, q, a)$ is evaluated from actuation efforts and risks to take action $a$ at robot state $q$ on $W_{occ}$. \rev{Minimizing this cost function in Eq.~(\ref{eq:receding_objective_function}) simultaneously solves for the mapping and planning (SMAP), maximizing the coverage for \textit{artifact detection} and minimizing the action risk (e.g., collision chance).}
This reward function can be generalized to Simultaneous Localization and Planning (SLAP) problems (e.g., \cite{agha2018slap} or \cite{BVL}) by incorporating information gain based on localization entropy reduction events, such as a loop closure or landmark detection.

\begin{figure}[t!]
  \centering
  \includegraphics[width=0.7\columnwidth]{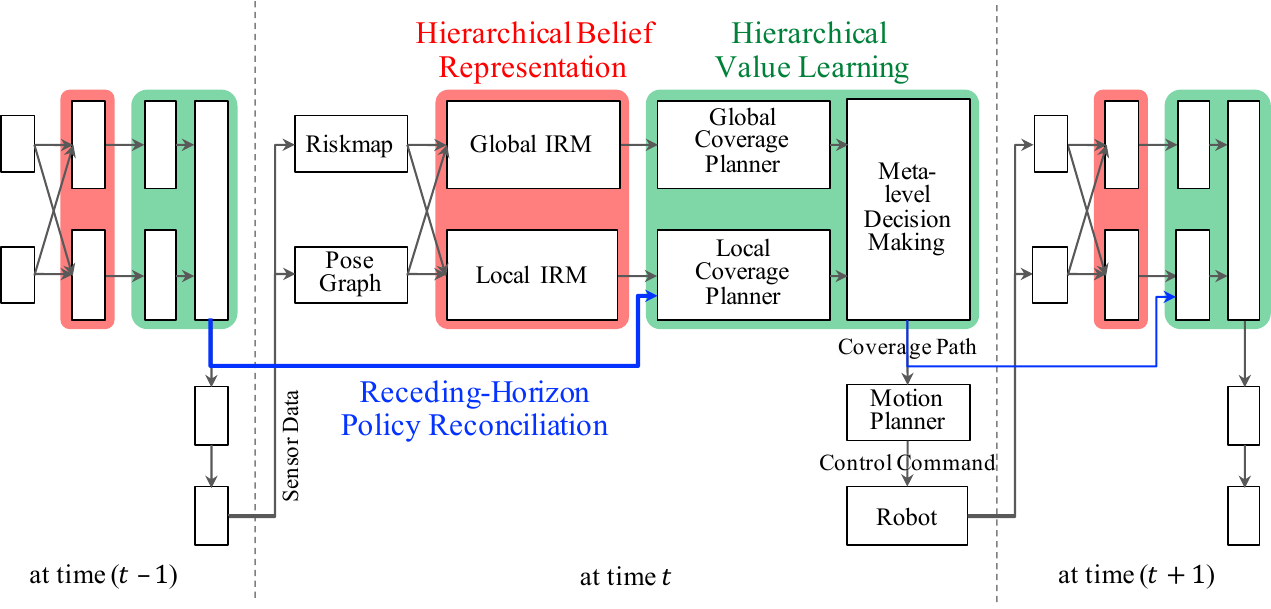}
  \caption{\rev{Illustration of PLGRIM framework for large-scale exploration in unknown environments.
 Over the receding-horizon planning episodes, PLGRIM (\textit{i}) maintains hierarchical beliefs about the traversal risks and coverage states, and
  (\textit{ii}) performs hierarchical value learning to construct an exploration policy.}} 
  \label{fig:gp_framework} 
\end{figure}

\subsection{Hierarchical Coverage Planning on Information Roadmaps} \label{sec:plgrim}
In this subsection, we introduce NeBula's solution for uncertainty-aware global coverage planning, PLGRIM (Probabilistic Local and Global Reasoning on Information roadMaps) \cite{kim_plgrim_icaps_2021}.
PLGRIM proposes a hierarchical belief representation and belief space planning structure to scale up to spatially large problems while pursuing locally near-optimal performance (see Fig.~\ref{fig:gp_framework}).
At each hierarchical level, it maintains a belief about the world and robot states in a compact form, called Information RoadMap (IRM), and solves for a POMDP policy to generate a coverage plan over a non-myopic temporal horizon, in a receding horizon fashion.

\begin{figure}[t!]
\centering
    \begin{tikzpicture}
    \node[anchor=south west,inner sep=0] (image) at (0,0) {\includegraphics[width=0.5\columnwidth]{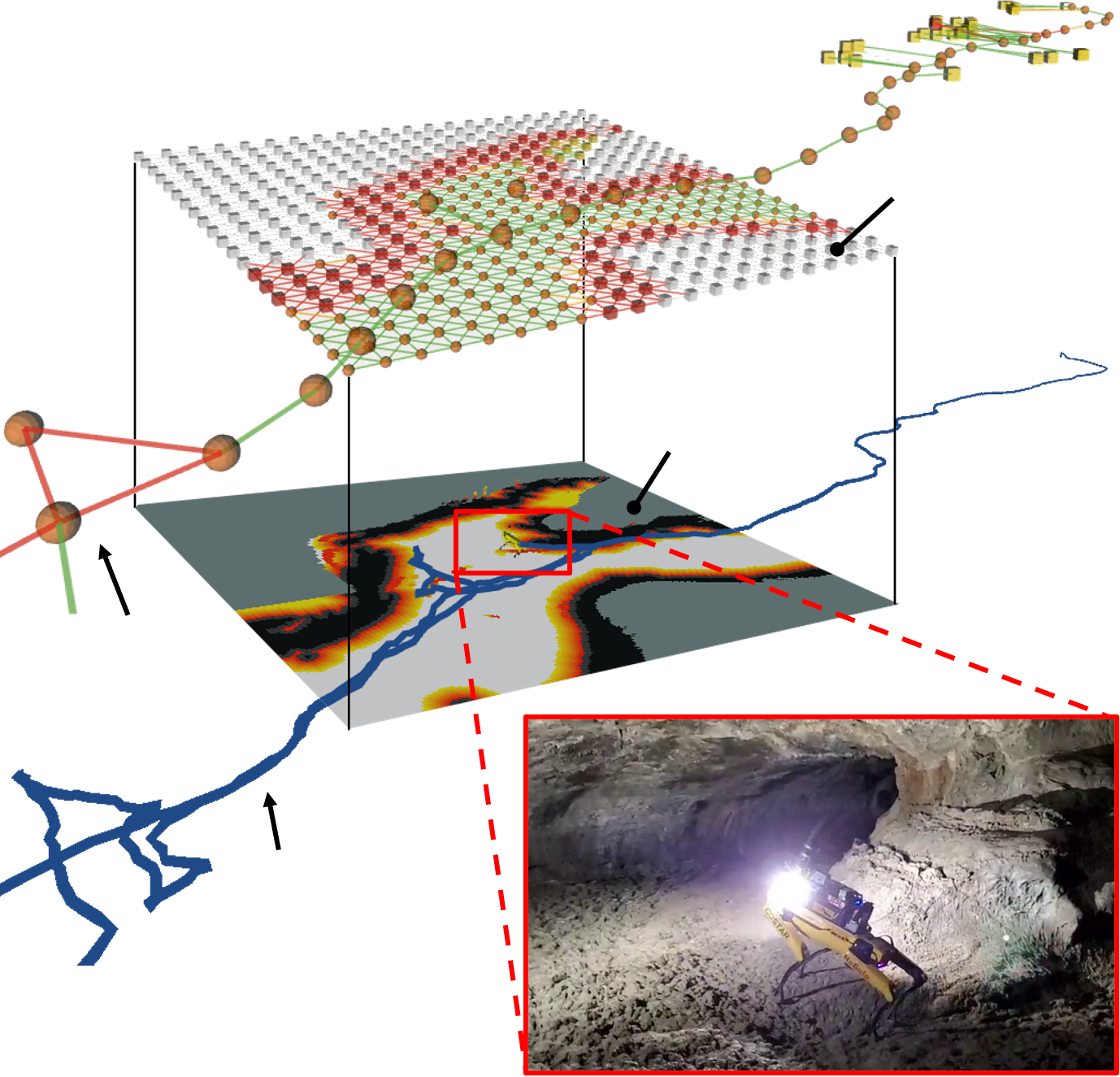}};
	    \begin{scope}[x={(image.south east)},y={(image.north west)}]

	    	\node [font=\scriptsize,above left,align=right,black] at (0.93,0.82) {Local IRM}; %
	    	\node [font=\scriptsize,above left,align=right,black] at (0.7,0.58) {Riskmap}; %
	    	\node [font=\scriptsize,above left,align=right,black] at (0.38,0.16) {Pose Graph};
	    	\node [font=\scriptsize,above left,align=right,black] at (0.23,0.38) {Global IRM};

	    \end{scope}
	\end{tikzpicture}	
  \caption{Hierarchical IRM generated during autonomous exploration of Valentine's cave at Lava Beds National Monument, Tulelake, CA, using a quadruped robot} 
  \label{fig:IRMs} 
\end{figure}
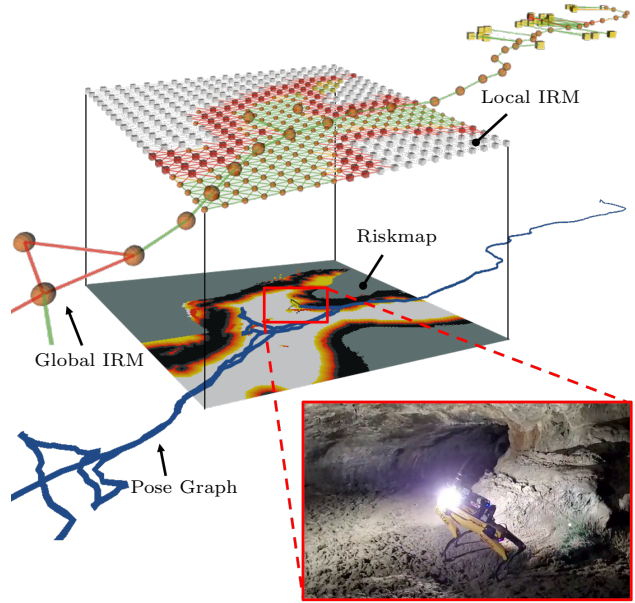

\ph{Hierarchical POMDP Formulation}
First, we formulate the receding-horizon SMAP in Eq.~(\ref{eq:receding_objective_function}) into a hierarchical POMDP problem \cite{kaelbling2011planning,kim2019pomhdp}.
Let us decompose a belief state $b$ into local and global belief states, $b^\ell = P(q, W^\ell)$ and $b^g = P(q, W^g)$, respectively.
$W^\ell$ is a local, rolling-window world representation with high-fidelity information, while $W^g$ is a global, unbounded world representation with approximate information (see Fig.~\ref{fig:IRMs}).
With $\pi^\ell$ and $\pi^g$ denoting the local and global policies, respectively, we approximate Eq.~(\ref{eq:receding_objective_function}) as cascaded hierarchical optimization problems as follows:
\begin{align}
  &\pi_{t:t+T}(b)
   \approx \argmax_{\pi^\ell \in \Pi^\ell_{t:t+T}} \, \mathbb{E} \sum_{t'=t}^{t+T} \gamma^{t'-t} r^\ell(b^\ell_{t'}, \pi^\ell(b^\ell_{t'}; \pi_{t:t+T}^g(b^g_t))),
  \label{eq:llp_optimization}
  \\
  &\,\,\,\text{where }
  \pi_{t:t+T}^g(b^g) = \argmax_{\pi^g \in \Pi^g_{t:t+T}} \, \mathbb{E} \sum_{t'=t}^{t+T} \gamma^{t'-t} r^g(b^g_{t'}, \pi^g(b^g_{t'})).
  \label{eq:glp_optimization}
\end{align}
$r^\ell(b^\ell, \pi^\ell(b^\ell))$ and $r^g(b^g, \pi^g(b^g))$ are approximate belief reward functions for the local and global belief spaces, respectively.
Note that the co-domain of the global policy $\pi^g(b^g)$ is a parameter space $\Theta^\ell$ of the local policy $\pi^\ell(b^\ell; \theta^\ell)$, $\theta^\ell \!\! \in \! \Theta^\ell\!$.\,

\ph{Hierarchical Belief Representation}
For a compact and versatile representation of the world, we rely on a graph structure, $G = (N, E)$ with nodes $N$ and edges $E$, as the data structure to represent the belief about the world state.
We refer to this representation as an IRM \cite{Ali14-IJRR}.
We construct and maintain IRMs at two hierarchical levels, namely, Local IRM and Global IRM (see Fig.~\ref{fig:IRMs}).
The Local IRM is a dense high-resolution graph that contains high-fidelity information about the occupancy, coverage, and traversal risks, but locally around the robot.
In contrast, the Global IRM sparsely captures the free-space connectivity. It encodes uncovered area by so-called frontier nodes, which allow for effective representation of large environments, spanning up to several kilometers.
In addition to the map uncertainty, IRM can be generalized to incorporate the robot localization uncertainty (e.g., \cite{BVL} or \cite{agha2018slap}) in the planning framework when traversing narrow passages and challenging environments where robot location uncertainty can hinder robot's ability to navigate the environment.

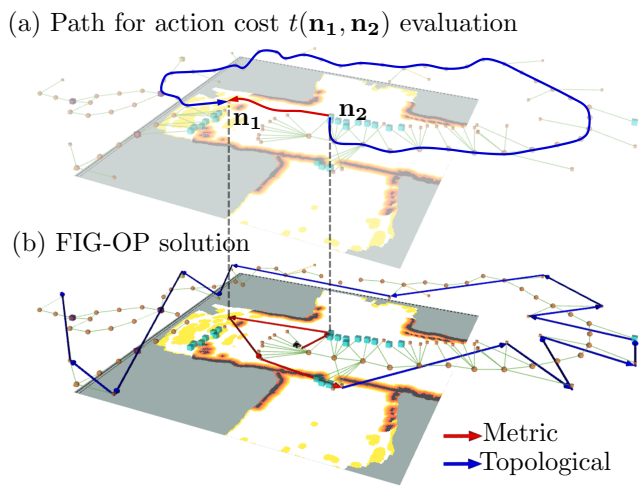
\begin{figure}[!t]
\centering
    \begin{tikzpicture}
	    \node[anchor=south west,inner sep=0] (image) at (0,0) {\includegraphics[width=0.5\columnwidth]{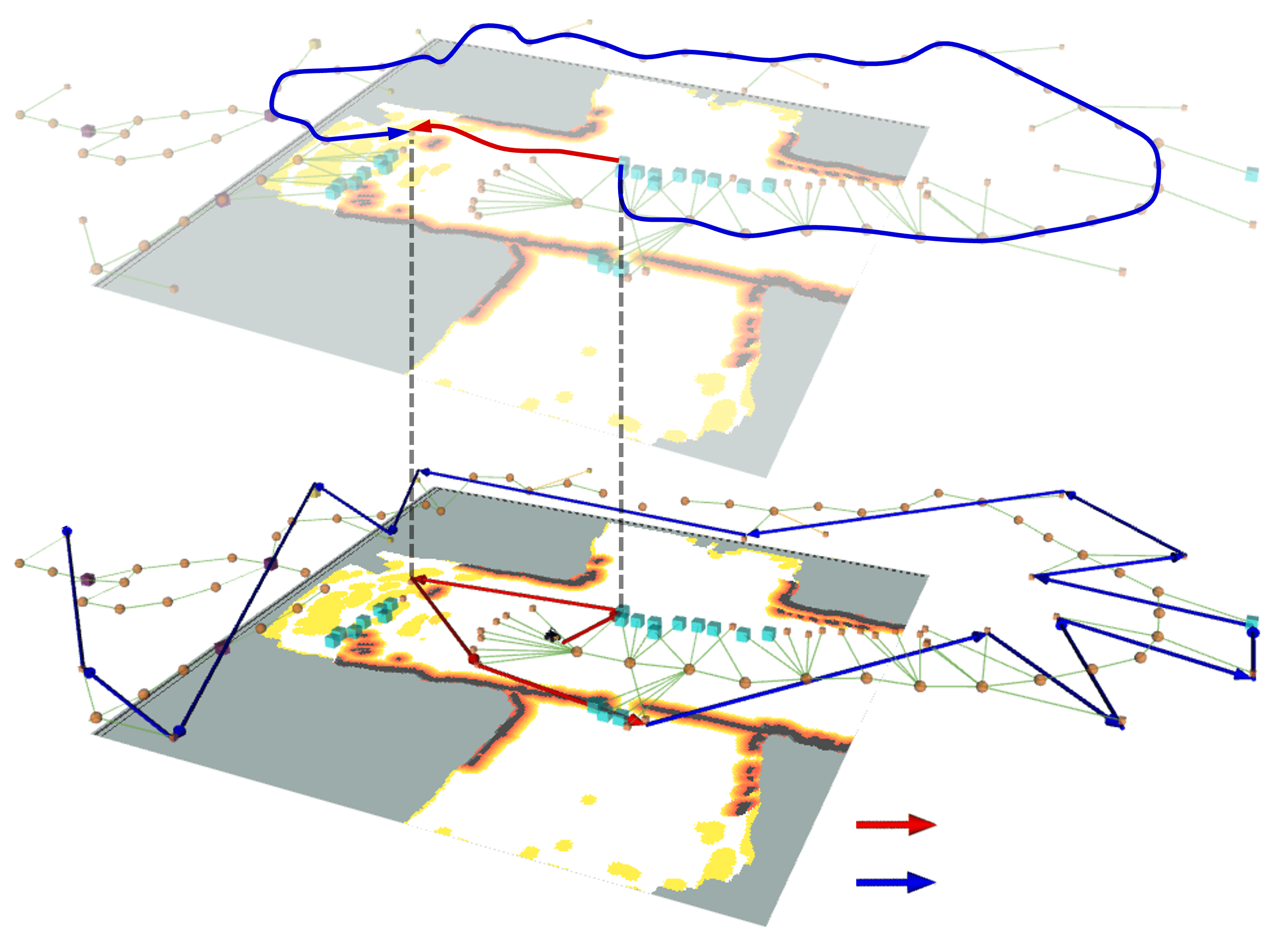}};
	    \begin{scope}[x={(image.south east)},y={(image.north west)}]
	    
        
    	\node [above left,align=left,black] at (0.80,0.98) {(a) Path for action cost $t(\mathbf{n_1},\mathbf{n_2})$ evaluation};
        \node [above left,align=left,black] at (0.37,0.5) {(b) FIG-OP solution};
        
        \node [above left,align=left,black] at (0.39,0.78) {$\mathbf{n_1}$};
        \node [above left,align=left,black] at (0.56,0.80) {$\mathbf{n_2}$};
        
    	\node [above left,align=right,black] at (0.865,0.10) {Metric};
    	\node [above left,align=right,black] at (0.95,0.025) {Topological};
        	
	    \end{scope}
	\end{tikzpicture}	
\caption{The FIG-OP graph combines low-fidelity action costs computed on the topological map (blue) and high-fidelity action costs computed on the metric map (red). Here, the topological-based and metric-based paths between two frontiers is displayed on the top layer. The topological-based path can only provide a crude estimate of path cost due to the sparsity of the graph. By using the more accurate metric-based path to computing path costs, FIG-OP finds a path that ``closes a loop" on the topological map -- a critical feature for accurately estimating action costs globally. The lower layer shows the solution returned by FIG-OP.} 
\label{fig:multifidelity}
\end{figure}

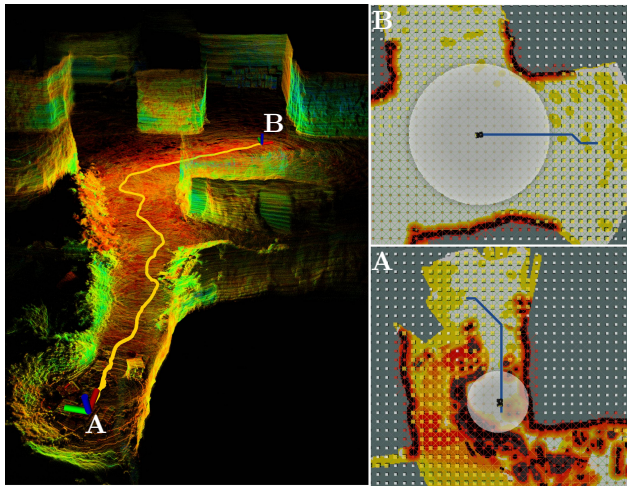
\begin{figure}[t]
\centering
    \begin{tikzpicture}
    \node[anchor=south west,inner sep=0] (image) at (0,0) {\includegraphics[width=0.5\columnwidth]{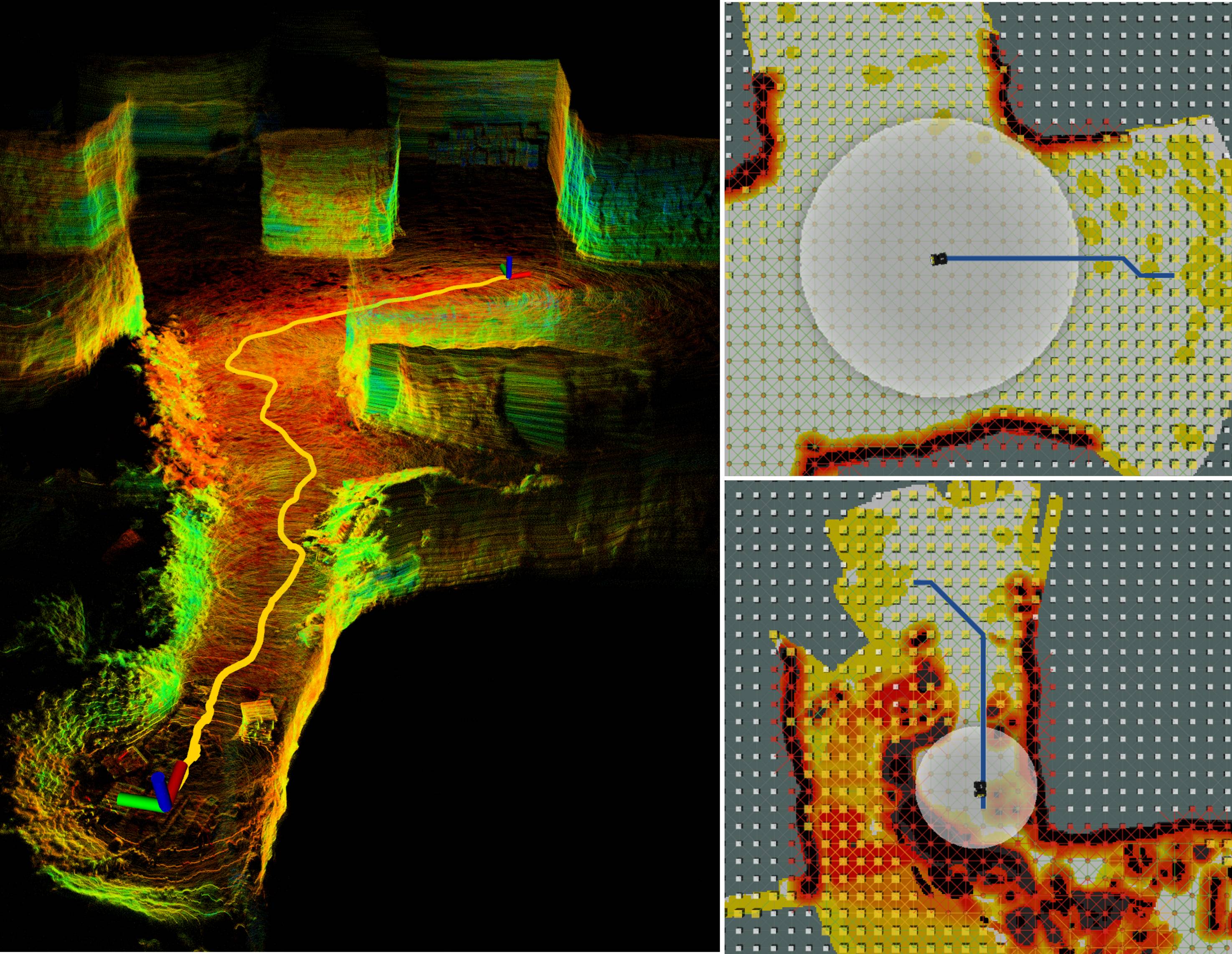}};
	    \begin{scope}[x={(image.south east)},y={(image.north west)}]
	    	
	    	\node [above left,align=right,white] at (0.18,0.09) {\textbf{A}};
	    	\node [above left,align=right,white] at (0.64,0.43) {\textbf{A}};
	    	\node [above left,align=right,white] at (0.465,0.715) {\textbf{B}};
	    	\node [above left,align=right,white] at (0.64,0.93) {\textbf{B}};

	    \end{scope}
	\end{tikzpicture}	
  \caption{Adaptive coverage range (translucent circle) and resulting exploratory path (blue) in a locally confined (A) and spacious area (B) during Husky's autonomous exploration of a limestone mine in Nicholasville, KY.} \label{fig:adaptive_IRMs}
  \vspace{-4mm}
\end{figure}


\ph{Hierarchical Coverage Planning--GCP}
Given Local and Global IRMs as the hierarchical belief representation of $W^\ell$ and $W^g$, respectively, we solve the cascaded hierarchical POMDP problems. %
At first, Global Coverage Planner (GCP) solves for the global policy in Eq.~(\ref{eq:glp_optimization}), providing \textit{global guidance} to Local Coverage Planner (LCP) of Eq.~(\ref{eq:llp_optimization}).
The global guidance enhances the coverage performance and global completeness. 
We propose a variant of the orienteering problem, called the \textit{Frontloaded Information Gain Orienteering Problem} (FIG-OP), for planning paths. The FIG-OP objective is a function of both information gain and travel distance, resulting in solutions that shift, or \emph{frontload}, information gain earlier in time. An illustrative example of the FIG-OP graph is shown in Fig.~\ref{fig:multifidelity}. We introduce an algorithm for solving FIG-OP in real time based on Guided Local Search \cite{gls}.



\ph{Hierarchical Coverage Planning--LCP}
In the hierarchical optimization framework, LCP solves Eq.~(\ref{eq:llp_optimization}), given a parameter input from GCP.
LCP constructs a high-fidelity policy by considering the information gathering (for volumetric coverage and artifact signals), traversal risks (based on terrain roughness, slopes, and mapping confidence), and robot's mobility constraints (e.g., acceleration limits and non-holonomic constraints of wheeled robots).
We solve such a simultaneous multi-objective optimization problem using \textit{POMHDP}, a multi-heuristic rollout-based search algorithm \cite{kim2019pomhdp}. This allows us to evaluate how the robot's observation of the world affects the utility of future observations, while simultaneously accounting for traversal risks and mobility constraints. In order to efficiently investigate the search space, iterative rollouts  are initially guided by an ensemble of heuristics and gradually exploit the learned values with theoretical suboptimality bounds. We also propose runtime adaptation of the coverage range to the local environment as an approximation to the effective coverage sensor model. 
As a result, LCP is enabled to be versatile in different mission contexts and resilient to real-world uncertainty. An example is shown in Fig.~\ref{fig:adaptive_IRMs}.



\ph{Meta-level Decision Making} 
Meta-level Decision Making for hierarchical coverage planning is focused on switching between the LCP and GCP using risk-aware information. The challenge lies in balancing between exploring a local area, which will provide some immediate coverage reward, or relocating to a different global frontier, which could provide more coverage reward in the future while also requiring more travel time. Instead of relying on heuristic rule-based switching, we use risk-aware information to reason about the probability of successfully executing a policy to decide between the local and global policies. To address this problem, we use the probability of successful policy execution, which is based on risk assessment of the environment and kinodynamic constraints. By reasoning about the probability of successful policy execution, we are able to balance safety and performance in a principled manner. Incorporating this probability is one of the key aspects of our risk-aware solution. Our decision problem can therefore be stated as: 

\begin{equation}
\max_{\pi^{a_t} \in \left\{\pi_{t:t+T^g}^g, \, \pi_{t:t+T^\ell}^\ell \right\}} \, \hat{P}(\pi^{a_{t}}) U(b_t; \pi^{a_t}).
\label{eq:Q_hat}
\end{equation} In other words, we are selecting the policy at time $t$ that has the greatest coverage reward if the plan is successful multiplied by the probability that the plan is able to be successfully executed.

\subsection{Experimental Evaluation}\label{sec:exp_results}
In order to evaluate our proposed framework, we perform high-fidelity simulation studies with a four-wheeled vehicle (Husky robot) and real-world experiments with a quadruped (Boston Dynamics Spot robot). Both robots are equipped with custom sensing and computing systems, enabling high levels of autonomy and communication capabilities~\cite{Otsu2020}. The entire autonomy stack runs in real-time on an Intel Core i7 processor with 32 GB of RAM. The stack relies on a multi-sensor fusion framework. The core of this framework is 3D point cloud data provided by LiDAR range sensors mounted on the robots~\cite{LAMP}. We refer to our autonomy stack-integrated Spot as Au-Spot~\cite{AutoSpot}.

\ph{Baseline Algorithms}
We compare our PLGRIM framework against a local coverage planner \rev{baseline} (next-best-view \rev{method}) and a global coverage planner \rev{baseline} (frontier-based \rev{method}). 
\vspace{-16pt}
\begin{enumerate}[label={\arabic*)}]
  \itemsep2em 
  \setlength{\itemsep}{2pt}
  \setlength{\parskip}{0pt}
  \item \textbf{Next-Best-View (NBV)}: 
  NBV~\cite{dang2019explore} is a widely-adopted local coverage planner that returns a path to the best next view point to move to.
  It uses an information gain-based reward function as ours but limits the policy search space to a set of shortest paths to sampled view points around the robot.
\rev{While NBV is able to leverage local high-fidelity information, it suffers from spatially limited world belief and sparse policy space.}
  \item \textbf{Hierarchical Frontier-based Exploration (HFE)}:
  Frontier-based exploration is a prevalent global coverage planning approach that interleaves moving to a frontier node and creating new frontiers until there are no more frontiers left (e.g., \cite{umari2017autonomous}).
  While it optimizes for the global completeness of environment exploration but often suffers from the local suboptimality due to its large scale of the policy space and myopic one-step look-ahead decision making.
	\rev{The performance of frontier-based methods can be enhanced by modulating the spatial scope of frontier selection, but it still suffers from downsampling artifacts and a sparse policy space composed of large action steps.}
\end{enumerate}
\vspace{-16pt}

\rev{
\subsubsection{Simulation Evaluation}
We demonstrate PLGRIM's performance, as well as that of the baseline algorithms, in a simulated subway, maze, and cave environment. Fig.~\ref{fig:maps_of_cave} visualizes these environments. In our comparisons, in order to achieve reasonable performance with the baseline methods in complex simulated environments, we allow baseline methods to leverage our Local and Global IRM structures as the underlying search space. We provide simulation results in Fig.~\ref{fig:figop_subway}.

\ph{Simulated Subway Station}
The subway station consists of large interconnected, polygonal rooms with smooth floors, devoid of obstacles. There are three varying sized subway environments, whose scales are denoted by 1x, 2x, and 3x. 

\begin{figure}[t!]
\centering
	\subfloat[][\rev{Simulated Subway 1x}]{
        \begin{tikzpicture}
    	    \node[anchor=south west,inner sep=0] (image) at (0,0) {\includegraphics[width=0.28\columnwidth]{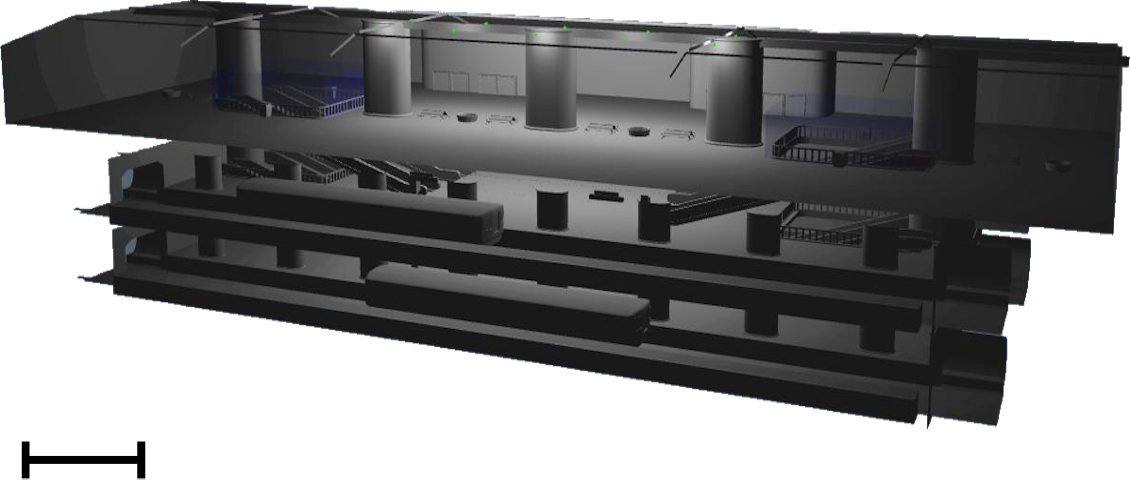}};
    	    \begin{scope}[x={(image.south east)},y={(image.north west)}]
    	    	\node [font=\scriptsize,above left,align=right,black] at (0.17,0.1) {15 m};
    	    \end{scope}
    	\end{tikzpicture}	
	}
	\subfloat[][Simulated Maze]{
        \begin{tikzpicture}
    	    \node[anchor=south west,inner sep=0] (image) at (0,0) {\includegraphics[width=0.25\columnwidth]{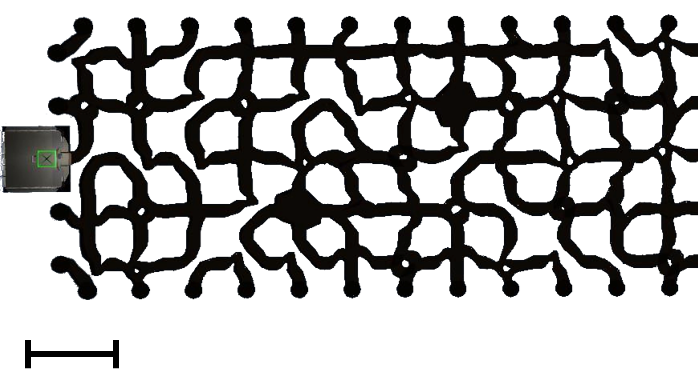}};
    	    \begin{scope}[x={(image.south east)},y={(image.north west)}]
    	    	\node [font=\scriptsize,above left,align=right,black] at (0.2,0.09) {10 m};
    	    \end{scope}
    	\end{tikzpicture}	
	}
	\subfloat[][Simulated Cave]{
        \begin{tikzpicture}
    	    \node[anchor=south west,inner sep=0] (image) at (0,0) {\includegraphics[width=0.38\columnwidth]{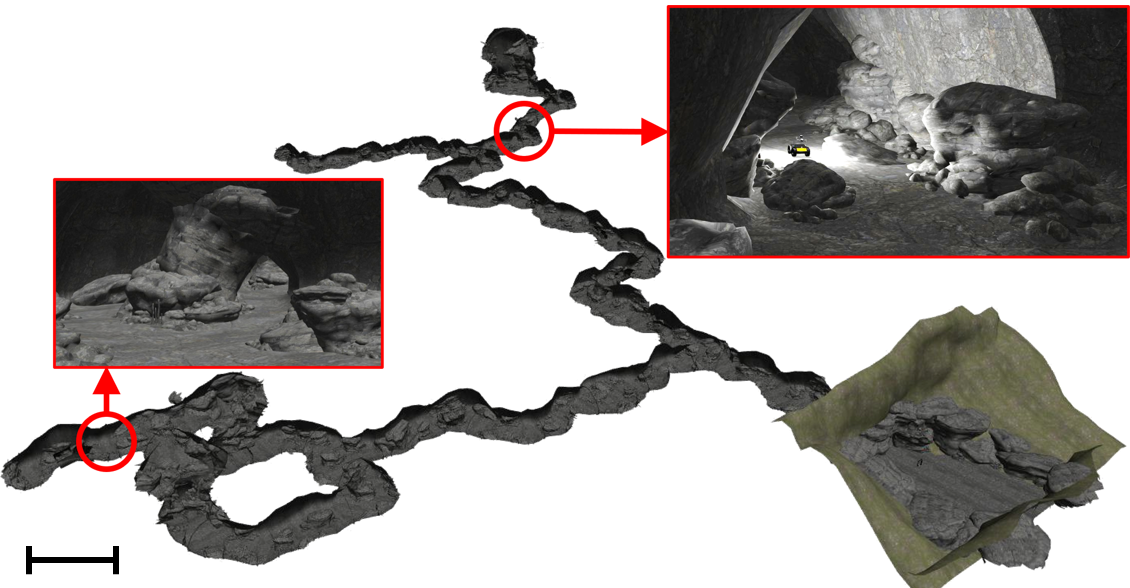}};
    	    \begin{scope}[x={(image.south east)},y={(image.north west)}]
    	    	\node [font=\scriptsize,above left,align=right,black] at (0.13,0.055) {10 m};
    	    \end{scope}
    	\end{tikzpicture}	
	}
\caption{\rev{Simulated environments for performance validation: (a) subway station,} (b) maze (top-down view), and (c) cave.}
\label{fig:maps_of_cave}
\end{figure}
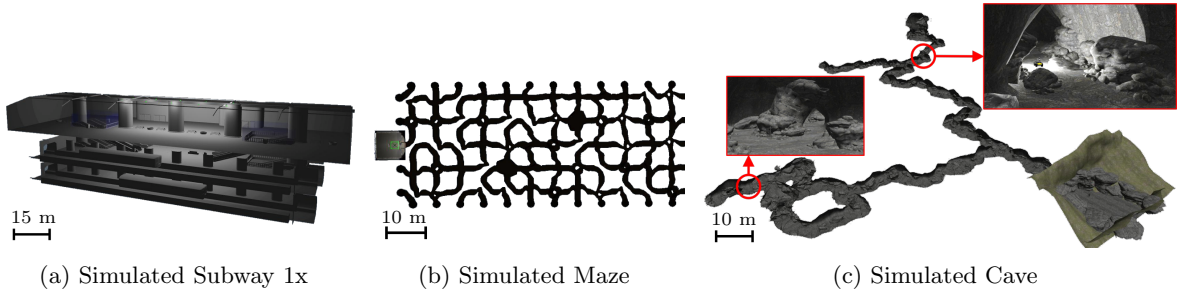

\begin{figure}[t!]

    \centering
    
	\subfloat[][]{\includegraphics[clip,width=0.24\columnwidth]{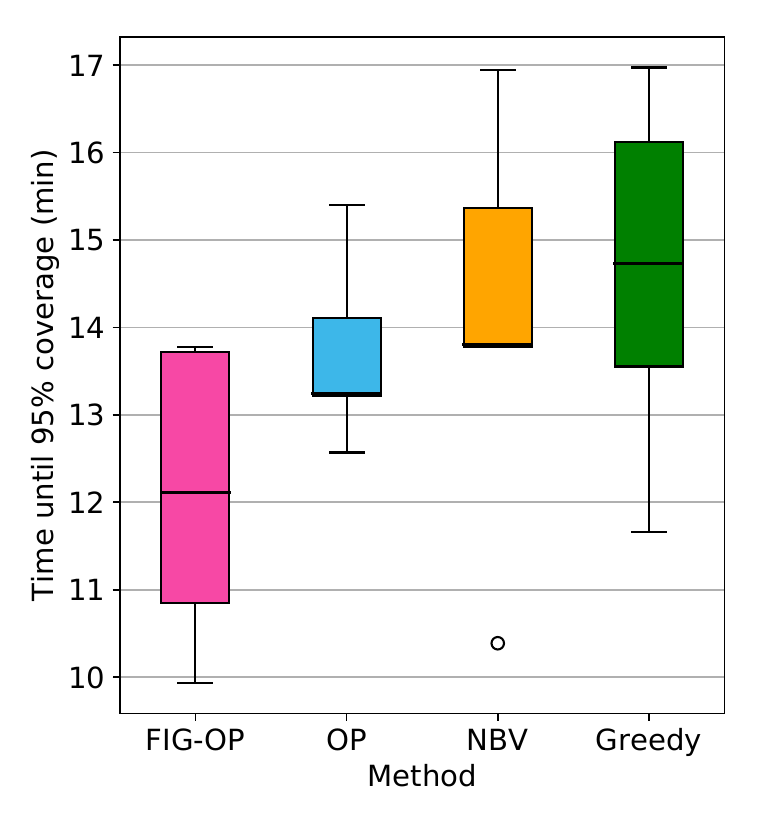}}
	\subfloat[][]{\includegraphics[clip,width=0.24\columnwidth]{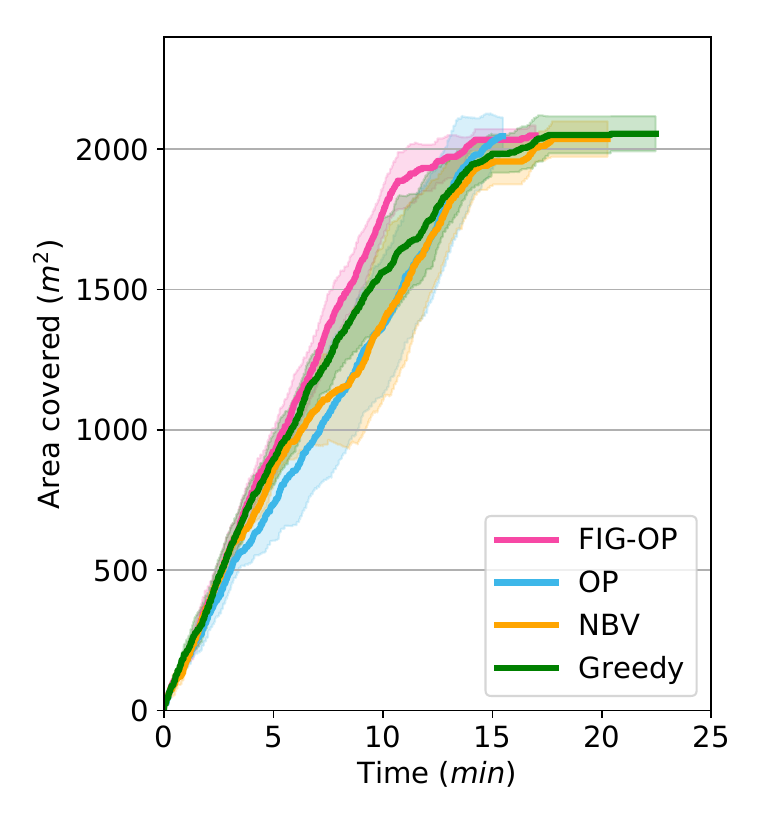}}	
	\subfloat[][]{\includegraphics[clip,width=0.52\columnwidth]{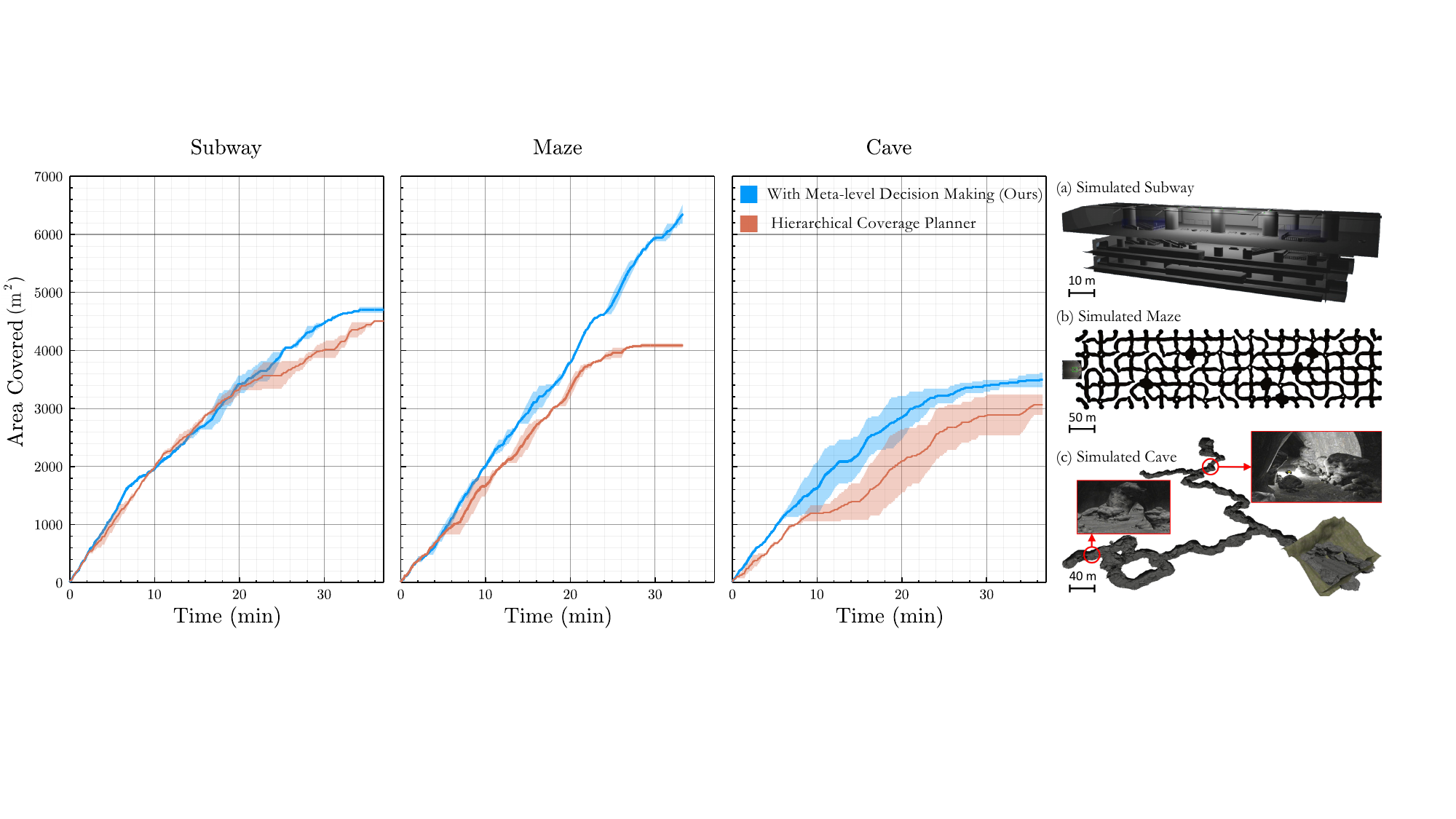}}
    
	\caption{
	(a) and (b): Exploration by FIG-OP and baseline global planning methods (no local planner) in a simulated subway. The exploration metrics for each method consist of five runs, with curve (b) displaying the average. Results showing the exploration performance with and without the meta-level decision making algorithm in the simulated subway, maze, and cave environments (c). The covered area is the average of two runs and the bounds denote maximum and minimum values between the runs. Note that (a,b) are isolated comparisons for GCP whereas (c) is focused on the Meta-level decision making problem discussed above.}
    \label{fig:figop_subway}
\end{figure}

\ph{Simulated Maze and Cave}
The maze and cave are both unstructured environments with complex terrain (rocks, steep slopes, etc.) and topology (narrow passages, sharp bends, dead-ends, open-spaces, etc.).
%
}

\begin{figure}[t!]

    \centering
    
	\subfloat[][]{\includegraphics[clip,width=0.33\columnwidth]{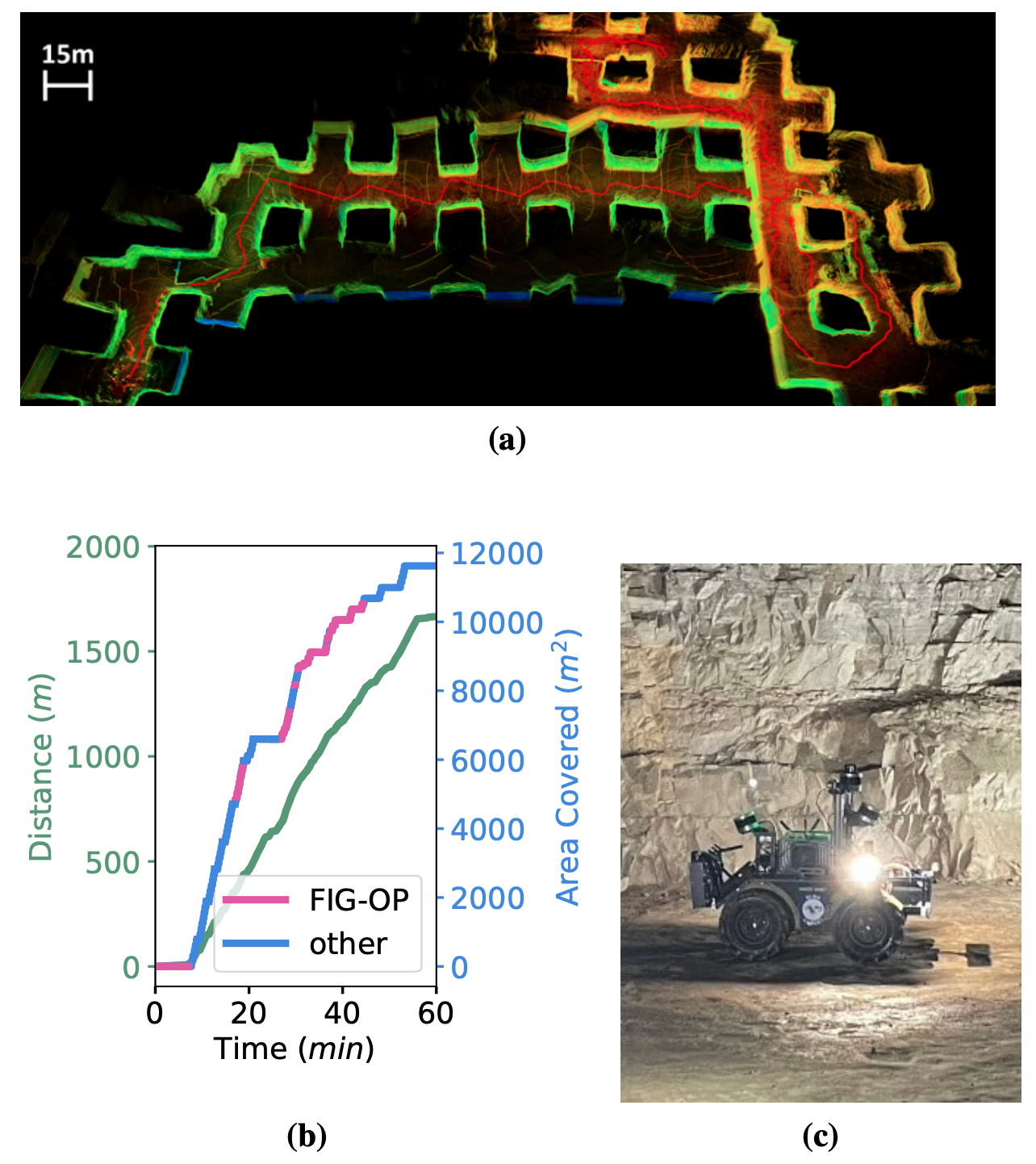}}
	\subfloat[][]{\includegraphics[clip,width=0.67\columnwidth]{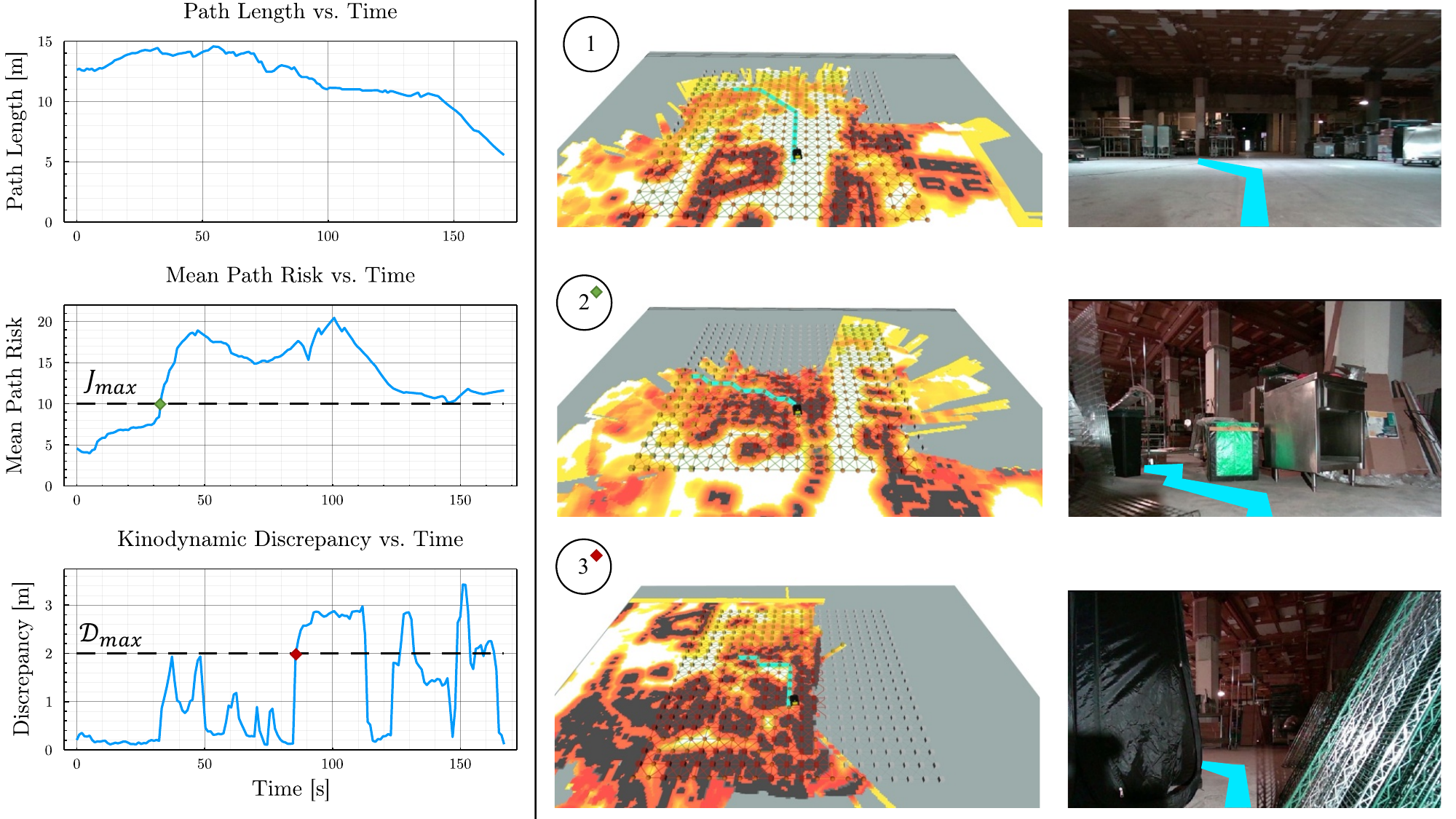}}

	\caption{Autonomous exploration of a limestone mine by the Husky robot (c). The robot's trajectory is overlaid on the aggregated LiDAR point cloud (a). In the exploration metric (b), pink denotes time intervals where FIG-OP was directly guiding the robot. Blue designates other planning modes, such as Local Planning \cite{kim2021plgrim}, user-specified goals, or return to communications. When the robot has been outside of the communications range of the base station for more than 10 minutes, the robot retraces its steps until communication is reestablished. The figure on the right (b) is an example highlighting the motivation for local to global switching from a real world test conducted in the LA Subway. This test was run without meta-level decision making onboard to demonstrate the necessity of using the algorithm. Initially, the robot is in an open space as shown in the center and right images of row 1. The robot then enters a cluttered environment that has high reward (since it has not been covered yet) and very high traversability risk (row 2). Obstacles are shown in black, areas of high traversability risk are shown in red, and white indicates open space. By replaying this data with meta-level decision making running, we can see that at this point the algorithm would recommend switching to a global goal rather than continuing with the risky local plan; however, since the algorithm was not running, the robot continues to explore the constrained environment. It slowly makes its way to a point shown in the third row where the local path is able to make it through a narrow passageway that is not kinodynamically feasible for the robot to execute. At this point the meta-level decision making algorithm would again recommend switching to a new global goal since it recognizes that executing the current local plan will result in the robot getting stuck. The green and red diamonds indicate the points in time where these events would have occurred if the algorithm had been running in real time on the robot. The robot's local plan is indicated by the cyan line.}
    \label{fig:gp_hardware}
\end{figure}

\subsubsection{\rev{Real-World Evaluation}}
We extensively validated our solution on physical robots in real-world environments. In particular, our solution was run on both a quadruped and wheeled robot in the Los Angeles Subway Station and the Limestone Mines of the Kentucky Underground, Nicholasville, KY. The LA Subway environment is a complex multi-level space with narrow corridors, many different rooms, obstacles, and a long tunnel passageway. The mine consists of a grid-like network of large passageways. The floor is characterized by very muddy terrain and other rubble areas with large boulders, from ceiling breakdowns, scattered throughout the environment. Inside the mine, there is also a file storage facility with smooth floors and many narrow passageways in between large shelves. Fig.~\ref{fig:gp_hardware} discuss how our planning architecture is able to overcome the challenges posed by large-scale environments with complex terrain and efficiently guide the robot's exploration. 

%% file: sections/7.comms.tex
\section{Multi-Robot Networking} \label{sec:multirobot_networking}
Efficient multi-robot exploration and operation in unknown environments requires communication between the robots for coordination and with the human operator at the base station for situational awareness (and scoring in the case of the DARPA Subterranean Challenge). Subterranean environments have a high level of uncertainty in the reliability, capacity, and availability of the links between robotic explorers themselves and the base station due to: i) limited line-of-sight opportunities and ii) the signals’ interactions with complex structures of the environment (e.g. reflecting, scattering). Despite these challenges, we want a communications architecture that can meet the needs of multi-robot exploration that has: high throughput, low latency, large effective communication ranges, reliable data transfer, and allows cooperation between robots via mesh networking.
Our previous communication solution, CHORD~\cite{CHORD,agha2021nebula} used off the shelf radios (Silvus) in combination with ROS for intra-robot communication and ROS2 for inter-robot communication. As we scaled this solution to larger numbers of robots and larger environments, we found an increased need for understanding how much data each robot has to transmit to the other agents and how much communication bandwidth would be available at arbitrary locations in the environment.  Additionally, controlling the order and priority of information transmission became more important as the amount of data needed to be transmitted to the base station increased. Here, we will briefly describe the advances added to ACHORD~\cite{saboia2022achord}, our new communication solution, and PropEM-L~\cite{clark2022prop}, our communication modelling solution.

\begin{figure}
    \centering
    \includegraphics[width=0.8\textwidth]{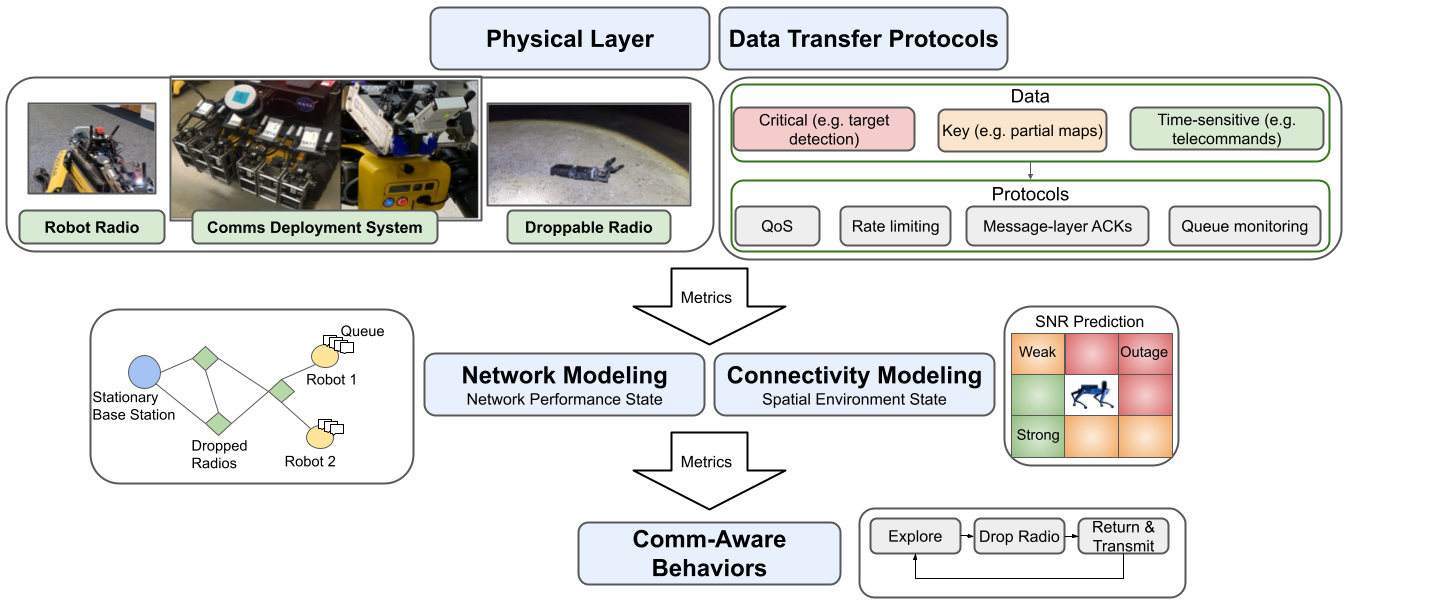}
    \caption{Multi-layer architecture of ACHORD (Autonomous and Collaborative High-bandwidth Operations with Radio Droppables). Lower-layer metrics exposed by the physical layer and data transfer protocols inform rich models of the network connectivity which inform high-level autonomy to allow comms-aware coordination. 
    }
    \label{fig:archord_arch}
\end{figure}

\subsection{ACHORD System Design}
Our communication architecture (seen in Fig.~\ref{fig:archord_arch}) seeks to extend the effective communication range from the base station and enable high-bandwidth data transfer over large distances. Our approach can be thought of in three parts -- i) The network architecture itself, which leverages droppable radios to build up a mesh network. ii) Distinct protocols for reliable and low-latency data transfer which meet the needs of mission-critical data and time-sensitive data, respectively. iii) Communication modelling and communication-aware autonomy, which allows our robots to make intelligent decisions, for example where to drop radios. 

\subsubsection{Mesh Network and Deployable Nodes}
Each of our robots is equipped with a Silvus MIMO radio, connected to the main computer via a dedicated Ethernet port. The radios form a mesh network and have layer 2 and 3 protocol implementations for multi-hop routing. Using a dedicated Ethernet port on the main computer (rather than sharing a single network switch for inter and intra-robot communications) limits accidental radio transmissions. Each Spot and Husky robots can also carry and drop an additional two and six deployable radios respectively, via custom hardware as described in Section~\ref{sec:ncds} and shown in Fig.~\ref{fig:comm_dropper}. 

\subsubsection{Communication Protocols}
Despite its popularity, previous work has shown that ROS is not well-suited for networks of robots subject to intermittent connectivity \cite{sikand2021robofleet}.
With this in mind, most field-hardened networking approaches rely on custom solutions built directly on UDP/TCP which act as a bridge between ROS-enabled robots \cite{hudson2021heterogeneous, ohradzansky2021multi, otsu2020supervised,tranzatto2022cerberus}.
With the addition of Data Distribution Service (DDS), ROS2 provides significant improvements including configurable Quality of Service (QoS) parameters. However, it has two shortcomings: (1) it lacks the option to resend only certain messages after reconnecting with the network, limiting the developer to select a static number of messages to resend (if too large, this will flood the network), and (2) it does not offer the application-layer a way to monitor the amount of data waiting to transmit, which is desirable in our case for high-level decision-making. For disruption-tolerant networks, a resend policy at the application layer which also offers buffer size monitoring is key.

\ph{ACHORD protocols} As described in \cite{saboia2022achord} for intra-robot comms, ACHORD leverages ROS for ease of development as inter-process comms on the same machine is not subject to strict bandwidth-limitations.
For inter-robot comms, we consider three types of data to transfer: (i) key data, which needs to be shared reliably and in-order but without timing restrictions; (ii) mission-critical data, which need to be shared reliably and with low-latency if possible; and (iii) time-sensitive data, which needs to be shared with low-latency but doesn't need repeating if dropped.
All data which is transmitted over the wireless network is compressed
into a generic, compressed data blob using bzip2\footnote{https://www.sourceware.org/bzip2/}. 
For neighbor discovery and transmitting time-sensitive data, we use ROS2 DDS (eProsima FastRTPS). As described in \cite{CHORD}, each robot has a ROS1/2 bridge, based on the ros1\_bridge package\footnote{https://github.com/ros2/ros1\_bridge}.
To address the need for application-layer monitoring and a resend policy, we introduce the \textit{Data Reporter} for buffering reliable data in queues, sending acknowledgement messages (ACKs), and monitoring buffer sizes \cite{saboia2022achord}. Because the data reporter is implemented in ROS, metrics on per-topic queue sizes can easily propagate up to higher levels of ACHORD. 
Key and mission-critical topics require a solution which is highly configurable and provides guarantees of message delivery. JPL multi-master (JPL~MM) \cite{otsu2020supervised} is a module which provides inter-robot comms that is compatible with ROS and built on top of the User Datagram Protocol (UDP). JPL~MM allows specifying configurations for each ROS topic and network port, and provides many customizable features, including token bucket rate limiting. A fixed number of tokens is allocated representing the maximum bandwidth available, and these tokens are allocated to each key/mission-critical ROS topic to prevent certain types of messages from overwhelming the network.

\subsubsection{Modelling and Autonomy}
Designing comm-aware exploration strategies requires modelling connectivity, and existing works model link qualities as a function of distance \cite{pei2013connectivity} or other factors like shadowing and multi-path components \cite{mostofi2010estimation, clark2022prop}. 
However, decision-making based solely on connectivity fails to consider whether robots have new information to transfer, and realistic memory constraints \cite{guo2018multirobot}. Thus, in \cite{saboia2022achord} we introduce two representations of state used for comm-aware decision-making: the spatial environment state and network performance state. 

\ph{State representations} The environment state is captured by \textit{comms checkpoints}, or IRM nodes (see Sec. \ref{sec:global_planning}) with an associated Signal-to-Noise Ratio (SNR)~\cite{schwengler2016wireless}. This value represents the maximum bottleneck SNR over all possible multi-hop routes from the transmitter at the base to the receiver at this checkpoint, where the bottleneck SNR indicates the weakest link along the route. Because the theoretical maxi data capacity is given by $C = B \log_2{(1 + \textrm{SNR})}$, where $C$ is the capacity and $B$ is the bandwidth available~\cite{schwengler2016wireless}, increasing SNR increases the amount of data a link can support. This information is shared by all robots via the IRM.
When new relay nodes are deployed, our predictive model updates the comms checkpoints that have become outdated. This predictive model can be based on free space path loss model which estimates signal attenuation in the absence of obstacles \cite{saboia2022achord}, or can incorporate 3D point cloud measurements to learn models which capture shadowing and scattering of the signal due to the geometry of the environment \cite{clark2022prop}. As demonstrated in \cite{clark2022prop}, we can predict SNR within a few dB of accuracy.
The network performance state is captured by statistics maintained on each robot: (1) buffer size: the amount of data (in bytes) that needs to be transferred; (2) measured data rate: the amount of data transferred per unit time; and (3) estimated transfer time: the amount of time required to empty the buffer. High buffer sizes indicate a robot has explored autonomously for a long period of time, and low data rates can be a sign of network congestion.

\ph{Comm-aware autonomy} Equipped with these metrics, robots can autonomously improve the communication network.
For example, when the network performance state indicates that the buffers have grown too large, the robot will autonomously sacrifice nominal exploration and move towards a comms checkpoint with high SNR to return to comms and ensure this data is transferred.
Additionally, when the environment state indicates the signal strength has become too weak, the robot will autonomously deploy a relay radio node via the NeBula Communication Deployment System (see Sec. \ref{sec:ncds}).
Other comm-aware autonomous behaviour is discussed in \cite{saboia2022achord} and the following section.


%% file: sections/8.mission_planning.tex
\section{Mission Autonomy, Planning, and Interfacing} \label{sec:mission_planning}



The main objectives of mission autonomy is to derive a course of actions to enable safe deployment, extended supervised monitoring, and continuous self-monitoring of a team of heterogeneous robots while maximizing the amount of distinct findings communicated to the base station in the allotted time~\cite{otsu2020supervised}. This capability is a crucial component of NeBula's autonomy framework, enabling coordinated exploration of large, complex, and unknown environments. Especially in the context of the SubT challenge, where only a single human supervisor interacts with the robot team, reduced human workload and increased levels of autonomy are crucial to achieve mission objectives, under time and resource constraints.

In this section, we present NeBula’s mission planning and autonomy components:

\subsection{Supervised Autonomy and Interface}
\ph{Complex Systems} Operating complex multi-robot systems with autonomous capabilities can be quite challenging. Large teams of robots, especially a priori unknown environments, and a large number of agents in one of the teams can lead to unpredictable failures. It is important to keep the single human supervisor's situational awareness at a high level to be able to swiftly react in case of failures, especially during high stress situations like the DARPA SubT Challenge.

\ph{Assistive Autonomy} CoSTAR deploys its autonomy assistant Copilot and a co-designed video game-inspired user interface \cite{kaufmann2021copilotIEEE,kaufmann2022Copiloting} to address robotic and human limitations while taking into account the uncertainty that is introduced by the complex systems. Copilot consists of monitoring and assistive capabilities that take over task planning, scheduling, and in non safety critical cases also the execution of tasks with varying levels of autonomy. The execution of tasks is verified and Copilot tries to correct for predictable errors by re-scheduling and re-trying tasks. The system notifies the human in the loop, if assistance is required or critical issues have been detected.

\begin{figure}
\centering
\includegraphics[width=0.9\textwidth]{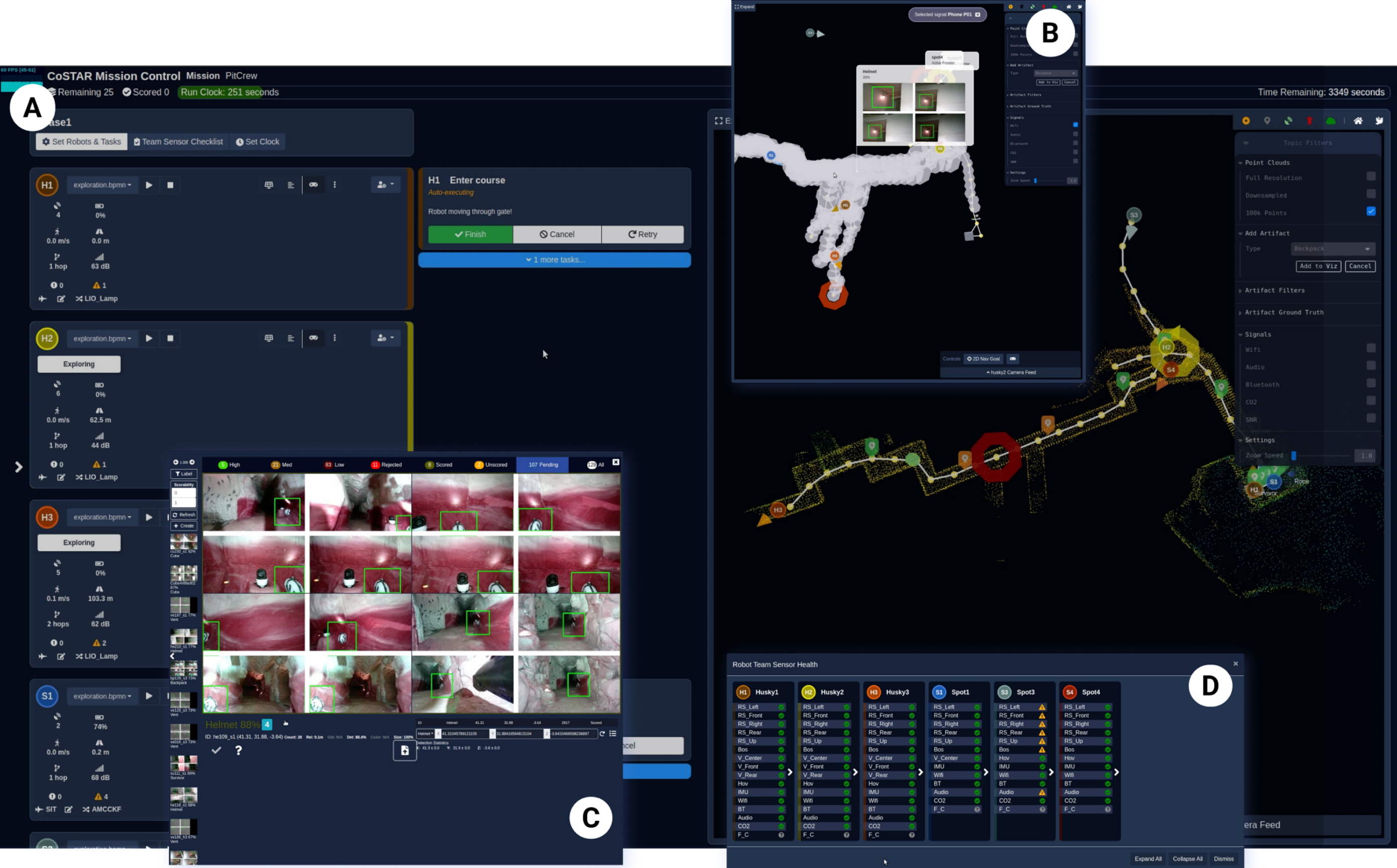}
\caption{An overview of the major UI components. (A) Three column layout with Robot and associated Copilot task cards, plus a split-screen 3D visualization of the perceived environment. (B) The split-screen 3D visualization with view controls, WiFi signal strength overlay, and an artifact card showing on the map. (C) The artifact drawer. (D) The robot health systems component. Modals (B),(C), and (D) are shown as an overlay for illustrative purposes.}
\vspace{-0.5cm}
\label{fig:ui_overview}
\end{figure}

\ph{Game-inspired Interface} Real-time strategy games such as Age of Empires,
StarCraft, and {Command \& Conquer} inspired CoSTAR's Mission Control Interface. In these games and for our real-world deployment of robots micro and macro-management of agents and the interaction with the uncovered environment are crucial to achieve mission success. Fig.~\ref{fig:ui_overview} provides an overview of the custom interface and its various components. The main view has been split into a three column design that is shown in part (A) of the figure. It consists of Robot Cards highlighting information that has been identified as most critical for each agent in the team (left) and their associated Copilot tasks (center). The 3D visualization of the environment as perceived by the robot agents is shown in the right part of the interface. (B) depicts a different layer of the 3D visualization where WiFi signal strength is shown. The artifact drawer (C) can be expanded when the human wants to focus on the search task and gives quick access to all artifact detections that then can be submitted for scoring. The robot health systems component (D) is used to rapidly identify sensor issues and provides further diagnostic information in case failures are detected without the need to switch applications.

\ph{Results} Copilot and the video-game inspired interface lead to rapid deployment of the robots, reduced the supervisor's need for application switching during operations, and increased the time available for the primary exploration and search tasks. For nominal deployment, we achieved sending six robots into the course in under one minute per robot. We would like to refer the reader to \cite{kaufmann2022Copiloting} for a more in depth description of CoSTAR's autonomy Copilot and supervisory control interface. 

\subsection{Mission Autonomy}

\ph{Settings and objectives} The main objective of the mission autonomy module is to manage the robots' activities with a focus on 1) maximizing exploration coverage and 2) timely reporting of findings. Using the TRACE tool~\cite{de2017mission, de2020event}, which tailors the Business Process Model and Notation (BPMN) to the robotics domain, we design and represent the mission that the robots must accomplish. 
The main components of our mission architecture have been described in~\cite{agha2021nebula} (Section 11). We have extended the mission architecture with additional behaviors, and have modified existing ones to make them more robust and responsive to changes in the environment. The nominal behavior of our mission is exploration, which is described in great detail in Section~\ref{sec:global_planning}. However, triggered in response to some key changes in the environment, in the robotic system, or instructed by the human supervisor, this nominal behavior might be interrupted and replaced by other support tasks that are critical to extending the uptime of our robotic systems as well as exploration and communication coverage. 

\begin{figure}
\centering
\includegraphics[trim={0cm 0.cm 0.2cm 0.80cm},clip,width=0.5\textwidth]{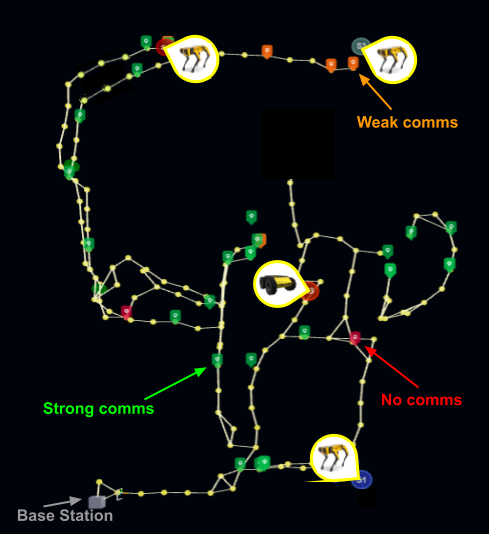}
\caption{Augmented Information RoadMap (IRM) with comms environment state:  green, orange, and red comms checkpoints represent strong, weak, and no comms, respectively.}
\vspace{-0.5cm}
\label{fig:environmen_representation}
\end{figure}

\ph{Mission health and representation} In addition to robot health, the mission autonomy uses information about the environment, such as traversability risk (see Section~\ref{sec:traversability}), communication state (see Section~\ref{sec:multirobot_networking}) and semantic information (see Section~\ref{sec:artifacts}) for decision making. Mission autonomy also aggregates mission-critical data and transmits them to the base station for supervised autonomy and situational awareness. Spatial information, such as stairs, doors, water puddles, radio locations, dead ends, signal strength represented as comms checkpoints (see Section~\ref{sec:multirobot_networking}), etc. that are measured or detected by a robot and useful to the mission autonomy of other robots, are incorporated into the shared IRM (see Section~\ref{sec:global_planning}). Fig.~\ref{fig:environmen_representation} shows an IRM highlighting the robots and the base station locations and comms checkpoints. The \textit{Mission Health}, such as the execution state, mission progress, type and amount of data in a robot's output queues, time since last connection, etc., which refers to the current state of autonomy of each individual robot, are communicated to the base station continuously.  

\ph{Mission components} Mission components are categorized into two types: monitors and behaviors. Monitors trigger the execution of behaviors based on some stimulus or when instructed by the human supervisor. In the following paragraphs, we describe the various components of our mission:
\begin{itemize}
    \item \ph{Mesh Network Deployment} To expand the effective range of communication, robots exploring the environment deploy a communication network by placing communication nodes at strategic locations in the environment. By monitoring the network conditions~\cite{saboia2022achord} and the environment geometry~\cite{vaquero2020traversability} the robots initiate and perform the deployment procedure autonomously. Coordination between robots is required during radio deployment to prevent redundant deployments in the same area. Before deploying an additional radio, other robots will consider whether the expected radio coverage of the two dropped radios would have significant overlap, and skip redundant deployments.
    The robots are equipped with the NeBula Communication Deployment System (NCDS), shown in Fig.~\ref{fig:comm_dropper}, which gives real-time feedback on whether the radio is deployed or is jammed. A failed radio drop then triggers a call to initiate a new drop to replace the node that has failed to deploy, improving the resiliency of the system to operational challenges. 

    \item \ph{Re-establishing Connectivity}
    After periods of disconnected exploration, the robot will sacrifice nominal exploration and instead move towards an area with strong communication. 
    Our solution selects the closest comms checkpoint from the IRM with SNR greater than the threshold $T_C=20$dB, 
    or a frontier neighboring this strong comms checkpoint, and moves towards it. 
    If the buffer size drops below a desired threshold ($T^l_{B} = 200$KB) before reaching the target checkpoint, nominal exploration continues immediately. Otherwise, the robot will wait at the target checkpoint until the buffer size drops below $T^l_{B}$. 
    If the buffer size does not decrease within a timeout (60 secs), which could indicate that the network is congested, the robot will choose a strong comms checkpoint closer to the base station as the new target. It will continue in this manner until the buffer size drops.
    
    \item \ph{Failure Detection, Recovery and Mitigation} Extended autonomous operation relies on the flawless interaction between complex software, hardware, and extreme environment. But even if we subject the system to endless testing and verification, latent software bugs, environmental factors and sensor degradation can cause major system failures~\cite{schumann2012software}. Although each component of our architecture is able to handle the known faults, unexpected interactions between hardware and software, and unforeseen environmental conditions can still drive the system to failures. The types of faults in a robotic system can be classified as endogenous and exogenous. Endogenous faults are the type of faults caused by the robot's internal components. On the other hand, exogenous faults can be caused by external factors in the environment. The mission autonomy module accounts for both kinds of faults. In particular, endogenous faults are detected through a self-monitoring mechanism that identifies the malfunction of a software component or driver software of a hardware component. When the robot health monitor determines an endogenous fault, recovery actions are triggered, starting from restarting the particular software component. When all the routines are exhausted, the robot notifies the operator of the fault to be rectified manually. Similarly, exogenous faults are identified by dedicated monitoring mechanisms that determine a particular type of fault and trigger an associated recovery mechanism. Two sets of modules identify and recover from exogenous faults: 1: stuck detection and recovery, and 2. fall detection and recovery.

    \item \ph{Stuck Detection and Recovery} Stuck detection primarily identifies whether the robot is physically stuck in the environment or virtually stuck inside a planner. The approach used to identify a stuck robot is by computing a moving average difference between the commanded velocity and the measured velocity of the robot. The robot is deemed stuck if the discrepancy between the commanded and measured velocity exceeds a certain threshold. When the stuck monitor determines the robot to be stuck, it triggers a sequence of recovery actions. The recovery action starts by resetting the planner and attempting to resume exploration, and this recovery action is beneficial if the robot is virtually stuck inside the planner. Other recovery actions target recovery from a physically stuck situation, where the robot appears to be stuck within a particular environmental feature. The robot performs a pre-programmed motion sequence like a wiggling motion to get unstuck from the environment. When all types of recovery action are exhausted, the robot notifies the human supervisor of the recovery failure and waits for instructions from the human supervisor. 

    \item \ph{Fall Detection and Recovery} Fall detection mainly targets the quadruped Spot robot. The robot could get unbalanced and fall when the quadruped attempts to traverse challenging terrains like a rocky slope or a slippery surface. The fall detection mechanism identifies when the robot incurs a behavioral fault like a fall and triggers a recovery behavior. The fall recovery mechanism attempts to perform a pre-programmed motion from the robot driver to recover the robot to a sitting position. The recovery behavior is repeated up to \textit{n} times to try and bring the robot to a sitting posture. Multiple recovery attempts are helpful if the robot appears to be stuck within the environment. Once the robot is in the sitting posture, it attempts to stand up and resets its planners to continue nominal exploration. When the \textit{n} recovery attempts are exhausted, the robot notifies the human supervisor and waits for further instructions.

    
    

\end{itemize}




%% file: sections/9.hardware.tex
\section{Ground mobility Systems and Hardware Integration} \label{sec:hardware} 

For the final DARPA event at the Mega Cavern in Louisville, Kentucky, we deployed a team of heterogeneous robots whose design and capabilities were aimed at exploration in a variety of environments.
These environments included tunnels, caves and urban structures with consideration given to our expectation that they would be combined into one `super environment.’
Learning from the previous DARPA SubT event at Urban, our ground robot team was focused into two categories for the Final event: (i) Wheeled rovers designed to carry large payloads and cover large areas quickly, and (ii) Legged robots to nimbly cover rugged and uneven terrains including steep gradients, staircases, voids and large ledges. For an in depth look at our previous mobility and hardware architecture for the Urban event, please refer to our Phase I and Phase II NeBula solution \cite{agha2021nebula}. 

\begin{table}[h!]
    \centering
    \caption{Comparison of the Wheeled and Legged Vehicles used by Team CoSTAR at the Final Event.}
    \label{tab:vehicle_comparison}
    \begin{tabular}{|c|c|c|}
    \hline
    \textbf{Topology}         &\textbf{ Wheeled Vehicles}    & \textbf{Legged Platform} \\
                              & ClearPath Husky               & Boston Dynamics Spot   \\ 
                              &       \includegraphics[height=3cm]{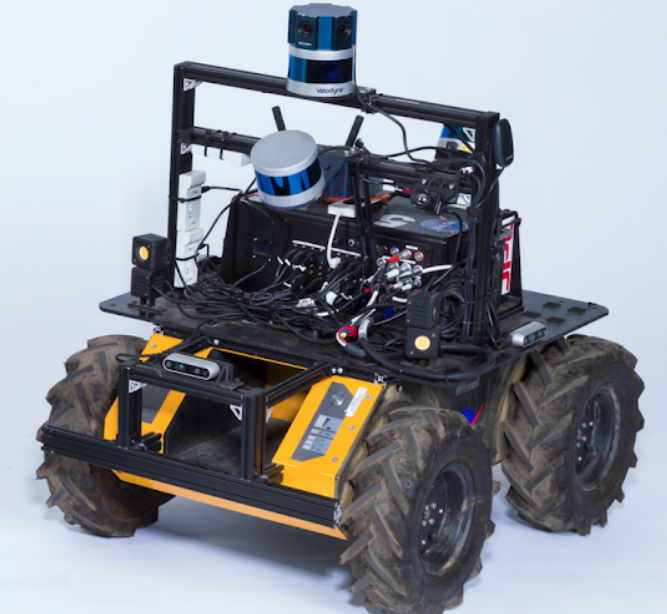}         &    \includegraphics[height=3cm]{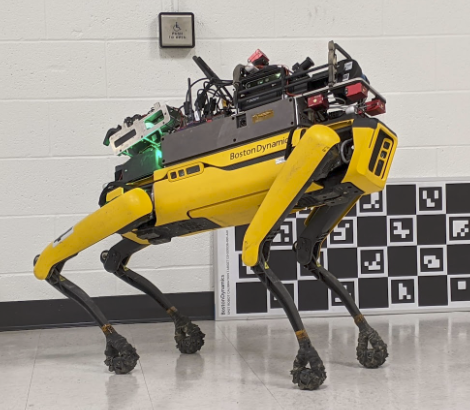}        \\ \hline
    \textbf{High-level}       & \textbf{SLAM:} Multiple, High-power CPU & \textbf{SLAM:} Mid-power CPU \\
    \textbf{Computers}        & \textbf{Object Recognition:} NVIDIA GPU & \textbf{Object Recognition:} NVIDIA GPU \\ \hline
    \textbf{Low-level  }      & \textbf{Safety Features:} Atmel MCU     & \textbf{Safety Features:} Atmel MCU  \\
    \textbf{Microcontrollers} & \textbf{Comm Dropper:} STM MCU          & \textbf{Comm Dropper:} STM MCU  \\ \hline
    \textbf{Sensors}          & LiDAR, RGBD and LW-IR cameras,       & LiDAR, RGBD and LW-IR cameras,  \\
                              & gas and signal sensors, external IMU    & gas and signal sensors, external IMU \\ \hline
    \textbf{Comm Dropper}     & 6 radios     & 2 radios   \\ \hline
    \textbf{Mass}             & 50 kg    & 35 kg   \\ \hline
    \textbf{Battery}          & Lithium Ion   & Lithium Ion  \\ \hline
    \textbf{Operational Time} & 3+ hours  & 1.5+ hours  \\ \hline
    \end{tabular}
\end{table}

Our ground robots (the ClearPath Robotics Husky and Boston Dynamics Spot) feature the largest payload capacity and greatest operational endurance of any of the robots in our fleet (Table \ref{tab:vehicle_comparison}).
This allows them to be equipped with the most sensors, largest radios and most powerful computers of any of Team COSTAR’s robots, enabling them to effectively act as the vanguard for our exploration.  

\subsection{Robot Modifications}
The Husky robots were modified significantly from the `stock’ factory configuration.
Larger motors were installed to counter the large mass of the compute and sensor payload.
This extra torque also helped the Huskies traverse higher gradients and taller obstacles, enabling them to tackle a wider range of terrain.
Higher-capacity Li-Ion packs were installed to significantly increase the run-time of the Huskies, meaning they could operate for longer during the mission without risking running out of battery.
Structurally, the custom main mounting plate and superstructure was kept from previous competitions.

The Spot’s feet were modified with higher traction rubber covers.
They were formed using mountain bike tires and fastened to the robot’s feet like shoes.
These rubber shoes enabled the Spot to traverse wet/muddy terrain with a reduced risk of a slipping.
In addition, the shoes enabled the robot to traverse steeper gradients and looser terrain as the deeper treads provided improved traction in less stable ground. 

\subsection{System Architecture}
Between Urban to Final circuit, the architecture of our systems \cite{agha2021nebula} remained stable, allowing development to focus on refining our system.  
The core modules of the system, which form an 'autonomy payload backpack', the NeBula Sensor Package (NSP), NeBula Computing Core (NCC), NeBula Power and Diagnostics Board (NPDB), and NeBula Communications Deployment System (NCDS) changed noticeably but slightly, however, significant upgrades in the firmware allowed us to improve the system reliability, added new features, and logging capabilities.
These changes are described in more detail in the following sections.

\subsection{Design Principles}
Final Circuit of the DARPA SubT Challenge combined all the challenging elements of the previous three competitions into one single course.  
Whereas previously our robots could be optimized for a single environment, we had to modify our set of design principles to provide a balanced performance over each of the environments. 
We considered our target environment, Louisville Mega Cavern, Kentucky, and determined that the following would be necessary to maximize our navigability of the environment, and hence, our artifact scoring potential: 

\begin{itemize}
    \item \textbf{Durability} We anticipated several falls and collisions with environmental features including rock walls and floors, concrete slabs and metal urban elements. By improving the resiliency of the system, we increased the probability that the robot could survive an impact, recover and continue the mission. 

    \item \textbf{Size Reduction} Footage released by DARPA prior to the event showed tight junctions, many of which could prove difficult to pass through with our Urban Circuit payload stacks. 
    This necessitated miniaturization of the payload, an especially challenging task as additional sensors were added to improve perception of the environment.
    
    \item \textbf{Weight reduction} Concerns for the robot’s balance during faster gaits prompted a reduction in the weight of the payload. 
    This was accomplished through use of lighter composite materials and optimizing the component layout to favor a lower center of mass.
    This improved the robots’ balance during traversal of rougher terrains and reduced the energy required to move the robot through the terrain.
    
    \item \textbf{Energy Increase} Reducing the sensor structure weight of the payload enabled us to increase the battery size carried by the robots.  
    This allowed us to push the operational time of the system to beyond the setup+operational time for the competition, negating the need to swap batteries mid-run.  
    It also additionally helped during the development phase of the project as tests could be run for longer, enabling bugs to be caught and fixed without constantly needing to change batteries.
    
    \item \textbf{Computational Power Increase} In order to accommodate the heavy computational loads of machine vision, SLAM, comms and motion planning, a more powerful computer was installed.
    This improved the speed of the overall autonomy stack. 

\end{itemize}

\subsection{Lessons Learned}
The prevalence of sensitive sensors on our robots and our extensive field testing augmented our design principles, briefly mentioned in Section 9.3. Our earlier designs had a focus on research and testing enabling functionality. This meant that many emerging and novel submodules with new functionalities were being developed with fast turnaround and sometimes with kinks that are to be expected in prototypes. Our later designs had a large focus on repeatable operation and physical resilience to extreme environments. Over the course of the entire SubT Challenge, and having documented our observations during mission testing, we iterated several versions of our `autonomy payload backpack'. Our final system, having been improved beyond the initial versions, was one that could be simply powered on with minimal steps. A developer or field engineer could basically load a battery or connect shore-power, press the on button and the overall system, including several computers, cameras, sub-modules and power core, would run without the need to debug, check connections or reboot. Our initial versions also saw a variety of different breakages, including potentially system compromising damage of the payload enclosures, cameras and auxiliary systems such as the communication deployment system. Further, having tested our systems extensively in a plethora of extreme environments including caves, mines, lave tubes, caverns, and cluttered urban structures, we were able to augment our system to a level where high-impact falls and collisions did not compromise the system's ability to perceive or navigate its environment. Some of the scenarios which were tested and observed include impacts with concrete and piles of quarry stones, tumbles down staircases, exposure to extremely high temperatures and collisions with protruding metal objects such as cast iron pipes. 

During the Final Run of the DARPA Subterranean Challenge in the Kentucky Megacavern, no hardware or system failures were observed in our autonomy payload backpack, including its subsystems, the NSP, NCC, NPDB and NCDS. They all ran without failure from beginning to end.

\subsection{NeBula Sensor Package}
The NeBula Sensor Package (NSP) gathers sensory information from the robot’s surrounding environment and relays this in real-time to the Power and Computing Core.
For the Final Circuit, we significantly enhanced the sensor packages for both the Husky and Spot robots, most notably to improve the fields of view for each type of sensor (LiDAR, electro-optical, etc.) by adding more sensors to cover blind-spots.
The Husky sensor package consists of 3 static Velodyne VLP16 LiDARs, a rotating Hovermap LiDAR from Emesent, a custom CO\textsubscript{2} sensor, 5 RGB-D D455 Intel Realsense cameras, a Bluetooth sensor, WiFi sensors, 2 FLIR long-wave thermal cameras, 6 high luminosity custom LEDs, and a VectorNav VN100 IMU.
The Spot sensor package feature the exact same sensors with the exception of possessing only a single static or spinning LiDAR.
The superstructure designed for this version iterates upon the prior design \cite{agha2021nebula} by using less material, reducing the size and mass, while maintaining high impact resilience, field of view and rigidity for sensor fidelity. Materials used include hard resins, carbon-fiber infused nylon and aircraft-grade aluminum.   

\begin{figure}[h!]
    \centering
    \includegraphics[height=5cm]{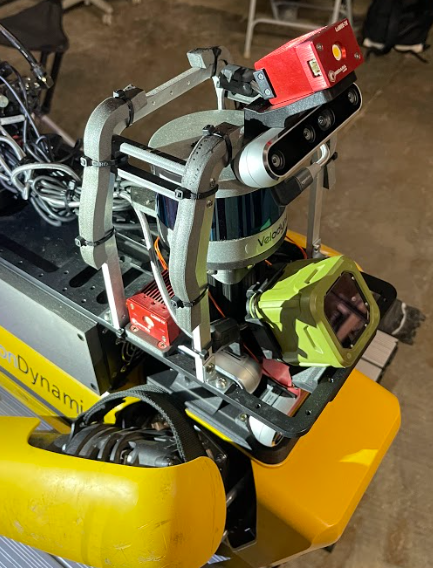}~\includegraphics[angle=90,height=5cm]{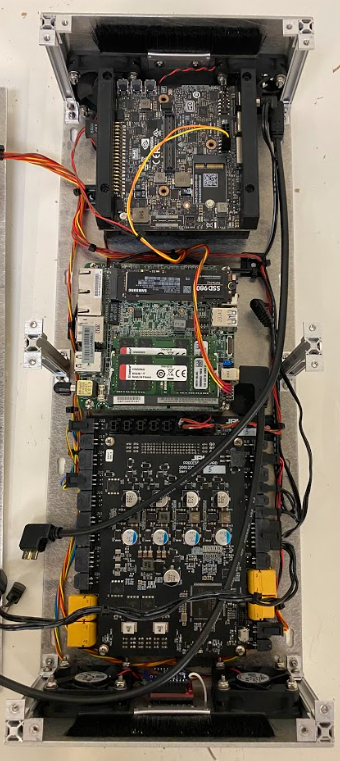}
    \caption{Left to Right: NeBula Sensor Package, and NeBula Computing Core and Diagnostics Board}
    \label{fig:computing_core_and_diagnostics_board}
\end{figure}

\subsection{NeBula Power and Diagnostics Board}
The NeBula Power and Diagnostics Board (NPDB) handles all of the power regulation and safety functions for the robots.
It acts as the primary source of power for the NCDS, NSP, and NCC.
A common NPDB design was used across all the ground robots, with hardware changes made to accommodate the various voltage ranges present within the vehicles.
The NPDB tailors its hardware safety circuitry to provide protection against excessive voltage and turn the system off if there is a fault, protecting the sensors from damage.

The NPDB also features a shore power circuit which allows for an external power supply to be used alongside the robot’s battery.
The NeBula payload automatically prioritizes the shore power input over the batteries, meaning regardless of the set up time taken, the robots would always begin the mission with a full battery.
This feature also enables batteries to be switched without needing to power off the robots, a significant advantage during long development sessions where battery changes would typically require the system to be shut down.

Lower-level software such as low-level system diagnostics, safety e-stops, power monitoring and thermal cooling flow rates are controlled by a microprocessor embedded on the NPDB flashed with custom firmware.
This firmware is also responsible for sequencing the power up of the robot and sensors, ensuring the peak loads on the system experienced during the initial boot phase are spread out, minimizing the total power spike.
Later revision of the NPDB featured additional circuitry to protect the system from transient voltage spikes. 

There were significant challenges with the development of the NPDB due to the international chip shortage.
The embedded processor was upgraded and the entire firmware was rewritten. The chip shortage forced the team to purchase components for the design prior to finalizing the design.
This would significantly constrain the development since the components would have to be purchased prior to detailed design of the system.

\subsubsection{General Power Distribution Architecture}

An overview of the NPDB is shown in the block diagram, Figure X. All power to the payload is passed through an ideal diode, implemented using the LTC4359 ideal diode controller. This provides reverse polarity protection to the whole system and allows for seamless power trade-offs between the battery and shore power supply. There is a fast blow fuse (SF-2923HC40C-2) before the ideal diode to protect the system in case there is something that causes a hard short on the ideal diode controller or anywhere downstream. The microcontroller (STM32F429ZGT6TR) monitors the input voltage of the system to make informed decisions on whether the system is presently on battery or shore power. There is a secondary protection for the power system which is implemented with an LTC4368-1 surge stopper. The surge stopper provides overvoltage and overcurrent protection downstream with the use of low resistance mosfets (NVMTS0D7N06CL) to minimize losses. The regulated voltages on the board are 5V, 12V and 17V. These voltages are provided using the LTC3891 synchronous step down controller with external top and bottom mosfets (NTTFS5C673NL and SiSS22LDN respectively). 

\subsubsection{Typical Power Sequencing Event}

The microcontroller is responsible for all power sequencing on the power board. Initially, the voltage at the input of the surge stopper is monitored to make sure it is within the allowable range. After the surge stopper is enabled, a voltage and current sense circuit using an LTC6102 and a voltage divider provide current and voltage measurements to the microcontroller. If there is a sudden rush of current, the microcontroller detects this as a short and will disable the surge stopper to prevent excessive currents from proceeding downstream. If no fault is detected on the bus, then each of the step down converters are enabled in series while the microcontroller checks that the output voltage is within the acceptable range. After the regulator voltages have stabilized, the microcontroller then enables the high side switch controllers one at a time to power the load. These high side switches are enabled with delays in between to prevent large inrush currents. The system arm state ultimately drives a high power RGB LED which itself is driven by three individual LED drivers (BCR421UFD). Each LED driver controls the brightness of its driven LED using a PWM signal provided by the microcontroller. The robot’s environment illumination LED’s are LuMEE-1K by Redara Labs. These LEDs are powered using a load switch connected to unregulated battery power. The brightness is controlled via PWM generated by the microcontroller. Battery state of charge is monitored using SMBUS and the system can alert the operator when the battery state of charge is approaching a critical level. A 12V buzzer is used to alert the user when the system is armed and when a warning is present on the vehicle (such as a low battery alarm). For safety, the system uses an RF-e-stop (XBee) in combination with a software and physical e-stop. There are a total of two signals emanating from the physical e-stop and one emanating from the microcontroller. These communicate with a 3 channel AND gate (SN74LVC1G11). This gate controls a safety relay which is responsible for disabling power to the vehicle. To prevent bouncing, a debouncing IC (MAX6817) is used to produce clean transitions for the e-stop input signals. The XBee e-stop is monitored using a UART interface and also has an additional digital output indicating when an e-stop request has been received. The microcontroller uses the e-stop output pin to indicate when either an RF or software e-stop has been received and has the ability to disable power to the vehicle via the safety relay.

\begin{figure}[h!]
    \centering
    \includegraphics[width=0.65\columnwidth]{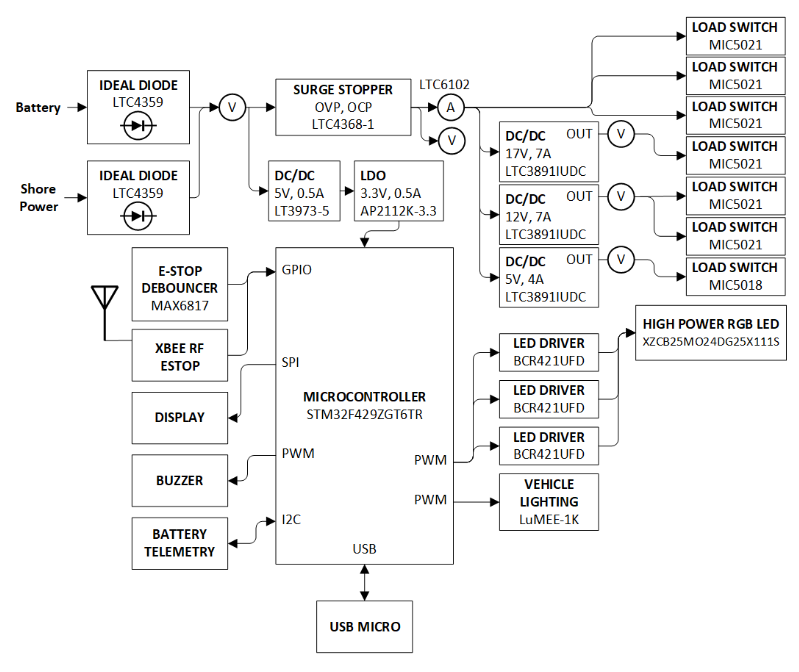}
    \caption{NPBD Block Diagram Overview}
    \label{fig:block__elec_architecture}
\end{figure}

\subsection{NeBula Communications Deployment System}\label{sec:ncds}
The control board for the NeBula Communications Deployment System (NCDS) was completely redesigned after the Urban Circuit.
A new control board was designed to house a more powerful microprocessor, necessitating the need to rewrite the firmware.
Extensive testing was done to identify cases where the communication nodes were failing to deploy, and vehicle angle limits were imposed to maximize the chance of a successful node drop.

\begin{figure}[h!]
    \centering
    \includegraphics[width=0.45\columnwidth]{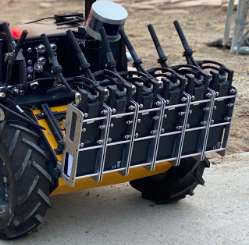}
    \caption{NeBula Communications Deployment System Mounted to a Husky}
    \label{fig:comm_dropper}
\end{figure}

An elastic band was added to provide additional force on the communications nodes, meaning we no longer solely rely on gravity to eject the radios. 
The servos used to hold the communication nodes were programmed to act as `pseudo sensors,’ checking for a successful deployment by attempting to re-close the servo, and monitoring for any obstruction (i.e. a jammed communication node).

On the user side, the communication node loading and unloading sequences were optimized to enable radios to be loaded quickly in the field, and the overall simplification of the process helped reduce loading errors in high-stress environments such as during practice games and the competition.
More robust monitoring of successful communications node drops helped inform the operator if the radio was still stuck within the Communications Deployment System.
An autonomy framework was built around this monitoring to drop an additional communication node if the first node failed to drop.

\subsection{NeBula Computing Core} The NeBula Computing Core (NCC) was developed to expand the autonomous capabilities of both the Spot and Husky robots.
There are two versions of the NCC, one for the Husky and one for the Spot.
As the Husky has a larger payload capacity, the NCC of the Husky robots had more computational power than the Spots.
The NCC consists of an Nvidia Jetson Xavier GPU and a CPU.
The Husky NCC came equipped with a mini-ITX motherboard and a 3\textsuperscript{rd} generation 16-core AMD Ryzen 3950X.
The Spot NCC came equipped with a Ruby R8 single board computer, equipped with an 8-core AMD Ryzen CPU.
The computers in the NCC are responsible for conducting artifact detection, motion planning, SLAM, communication of data, and other autonomy behaviors.
The NCC interfaces to the NSP to gather data and make informed decisions about the behavior of the vehicle.

%% file: sections/10.drone_hw_and_sw.tex
\section{Drone Hardware and Software} \label{sec:drone_hw_&_sw}

NeBula drone systems contribute to the multi-robot team by providing rapid and targeted access to
regions of subterranean environments that are denied or difficult to access via wheeled or legged robots.
Our team deployed two aerial systems \emph{Aquila} and \emph{Proxima} which are described in the following sections.
These aerial robots are subject to tighter constraints (SWaP or size, weight, and power) than either 
NeBula ground platform.
Both robots weigh less than 6kg and have endurance of approximately 20 minutes---significantly less than the Spot and Husky robots and three times shorter than the duration of a 60-minute final competition round.
The designs of these robots were also driven by the parameters and rules for the Subterranean Challenge 
competition as well as implicit and explicit expectations regarding the competition environments.
Both systems were expected to fly through the 2.5m wide starting gate and spaces as narrow as 1m.
Likewise, search in an unknown, harsh, and potentially dark environment dictated development of comparatively heavy autonomy payloads including onboard computers, cameras, lidar, lights, and radios.
The autonomy capabilities (NeBula) deployed on these robots were also heavily
specialized for aerial operation.
This specialization spanned localization, control, and autonomy design.
Regarding the aerial robots themselves, the Proxima and Aquila robots are similar in terms of size and their sensing and autonomy payloads.
However, they differ in terms of autonomy systems and concept-of-operation.
The following sections describe these similarities and differences.

\subsection{Concept of Operation}

The development of the concept of operation for NeBula drone systems was driven by four factors:
platform endurance, operator overhead, limited communication, and system maturity (and maturation).
Due to the limited endurance of aerial platforms, careful use of mission time was critical---coverage from a higher vantage point or access to even a single denied area could constitute a significant contribution to the team overall.
However, the team's desires for the robots to reach these views and regions were balanced against 
limited operator involvement
(due to interactions with other robots and scoring),
limited autonomy at the time of the competition,
and lack of communication with the robots far from the staging area.

These constraints in mind, deployments of the Proxima and Aquila systems each followed a similar pattern.
The operator would provide a mission specification while the drone was in the staging area or still within line of sight.
The drone would fly that mission, leaving line of sight and thereby losing communication with the operator.
After completing the mission or \emph{failing to do so}, the drone would return to the staging area \emph{before depleting the onboard battery and reaching the extent of mission endurance}.
Upon return to the staging area, mission logs were offloaded to an SSD for offline target detection and scoring.\footnote{%
    Neither drone carried a graphics unit capable of running detection and localization algorithms.
    As such, return to the staging area was critical to scoring whereas other robots could become stuck and still contribute significantly to scoring.
}

We designed missions for Proxima and Aquila according to slightly different philosophies.
Missions for Proxima consisted of sequences of waypoints forming rough paths for the robot to follow.
These missions would enable Proxima to quickly provide aerial views of the environment near the staging area in areas of the environment that had already been mapped or were otherwise expected to be navigable.
Aquila's missions consisted of a single goal point.
Navigating toward a distant goal point could enable Aquila to probe deep into unmapped parts of the environment, and targeted selection of goal points could enable rapid deployment to environment regions denied to ground robots.

\subsubsection{Capabilities Affecting Concept of Operation}

At the time of the competition, a number of capabilities that would each significantly impact the concept of operation were in various stages of development:


\paragraph{Shore power and battery swapping}
Both robots supported shore power units that enabled performance of startup operations without depleting
the batteries and subsequently reducing flight time.
The shore power units also provided the capability to swap batteries in the staging area which would 
significantly improve overall operation time.
However, we did not avail ourselves of this capability due to several factors:
because the shorter duration of the practice runs did not admit repeat deployments,
due to delay in detection and scoring from offline processing,
and due to limited time available for the pit crew to practice this operation and to validate the 
functionality of the autonomy system over repeat deployments.

\paragraph{Mesh radios and hand-deployed nodes}
The Silvus SC4200 radios carried and dropped by the ground robots were far too large and heavy for 
aerial deployment.
Instead we acquired a number of lighter ruggedized and OEM SL4200 radios for integration with our aerial platforms.
Although these were intended for use on both the Aquila and Proxima platforms, only Aquila included one of these radios in its payload at the time of the competition.
Due to reliance on operator input to specify missions for each drone, these radios could significantly 
improve abilities to extend and adapt missions online, after deploying the robots.

\begin{figure}
    \centering
    \includegraphics[height=0.33\linewidth]{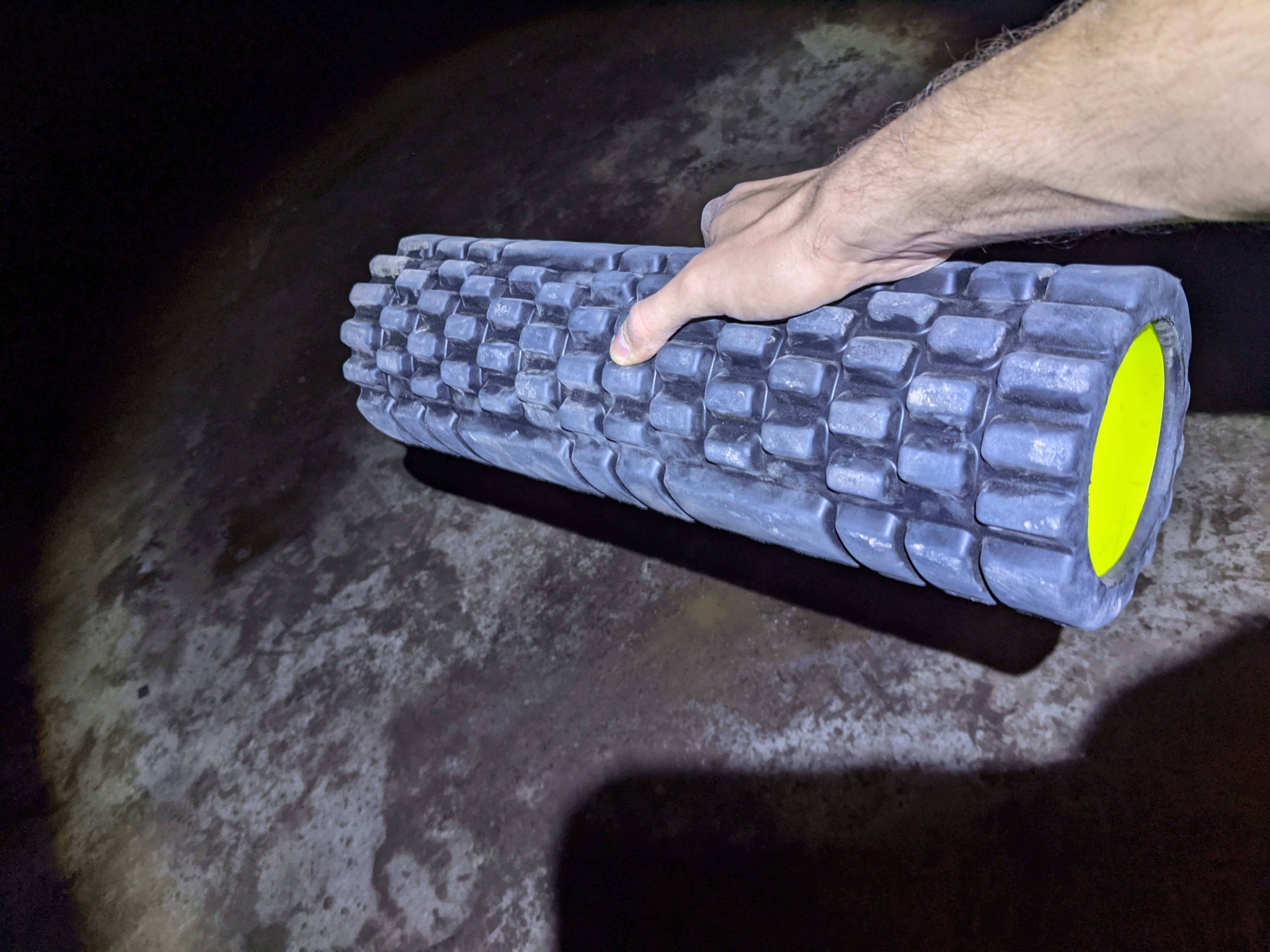}
    \includegraphics[width=0.33\linewidth,angle=90]{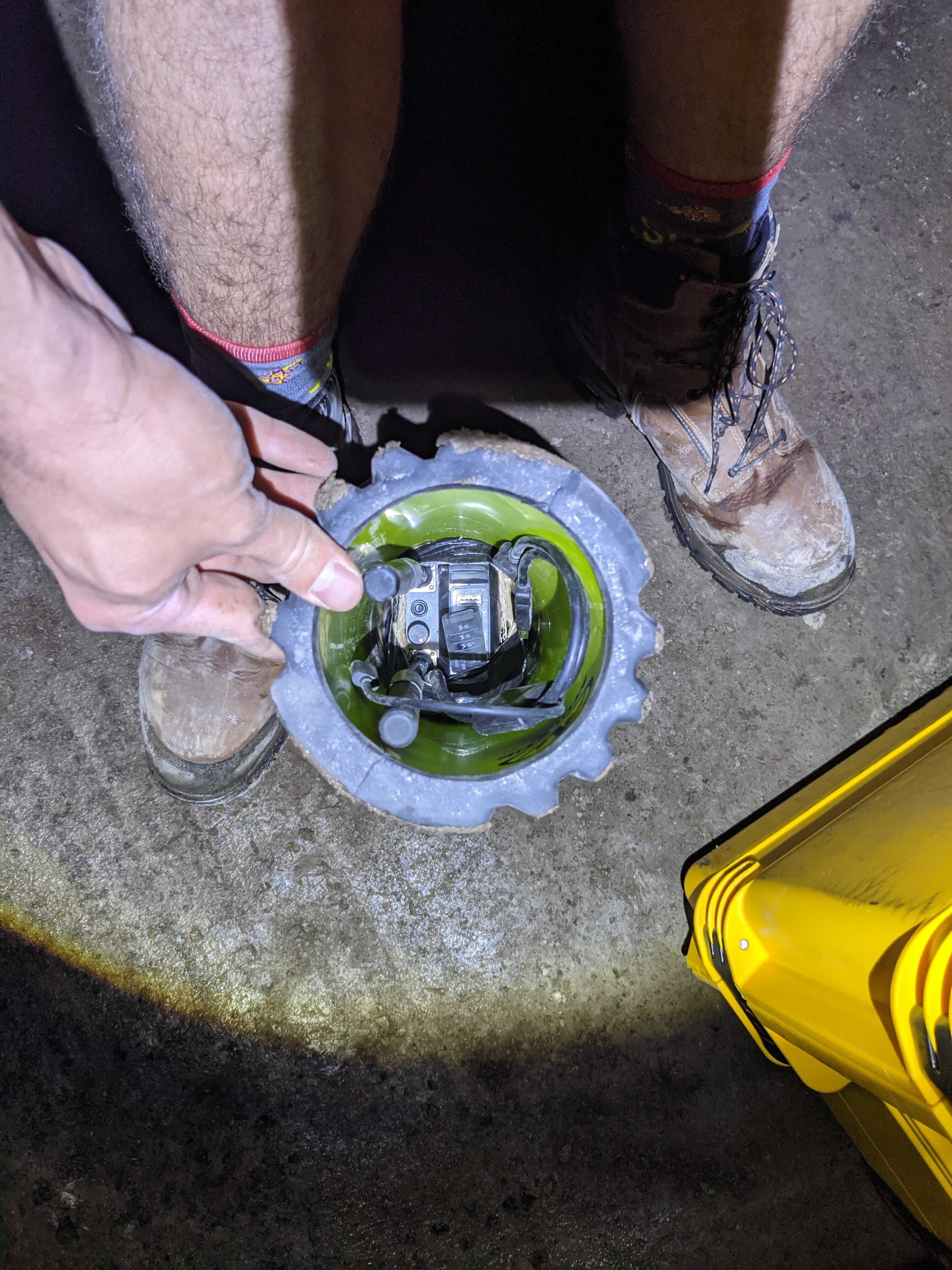}
    \caption{A hand-deployed mesh communication node, a \emph{communication roller}.
        The node shown above consists of a Silvus mesh radio and battery fit snugly inside an off-the-shelf athletic foam roller.
        Although simple, this design reliably protected protected the radio inside
        The communication roller could be deployed from the staging area by rolling the device with a motion similar to throwing a bowling ball. 
        Versions of this communication roller were developed for both aerial and ground robots and
        were deployed during competition rounds.
    }
    \label{fig:comms_roller}
\end{figure}

Although these radios were nominally capable of communicating on the same mesh network as the ground robots, they were not configured to do so.\footnote{%
    These radios were integrated at a late stage of preparation for the final competition.
    Late integration of mesh radios had two effects:
    drone systems and concepts of operation were developed with the assumption that communication would be denied at any significant distance from the staging area (e.g. past line-of-sight);
    and once communication capabilities had come available, the drone systems had not been proven to satisfy communication constraints \emph{alongside other robots}.
    The latter implied that deployment of aerial robots on the same network as the rest of the team
    could pose a significant risk as increased bandwidth usage---nominally or due to a bug---could 
    significantly impair communication for the rest of the robots.
}
However, communication on a separate network did not entirely deny the advantage of increased 
communication coverage due to deployment and operation of a mesh network. 
Toward this end, a \emph{hand-deployed communication roller} (Fig.~\ref{fig:comms_roller})
deployed from the staging area, generally targeting a nearby intersection, could significantly improve communication coverage.

\paragraph{Autonomous local coverage}

The team developed a local coverage behavior as an extension of the Aquila concept of operation.
After navigating to a goal point, Aquila could continue to explore locally, seeking to maximize coverage, before returning to the staging area.

\subsubsection{Aquila}

\begin{figure*}[t!]
    \centering
    \includegraphics[trim={0cm 0cm 0cm 1cm},clip, width=0.8\textwidth]{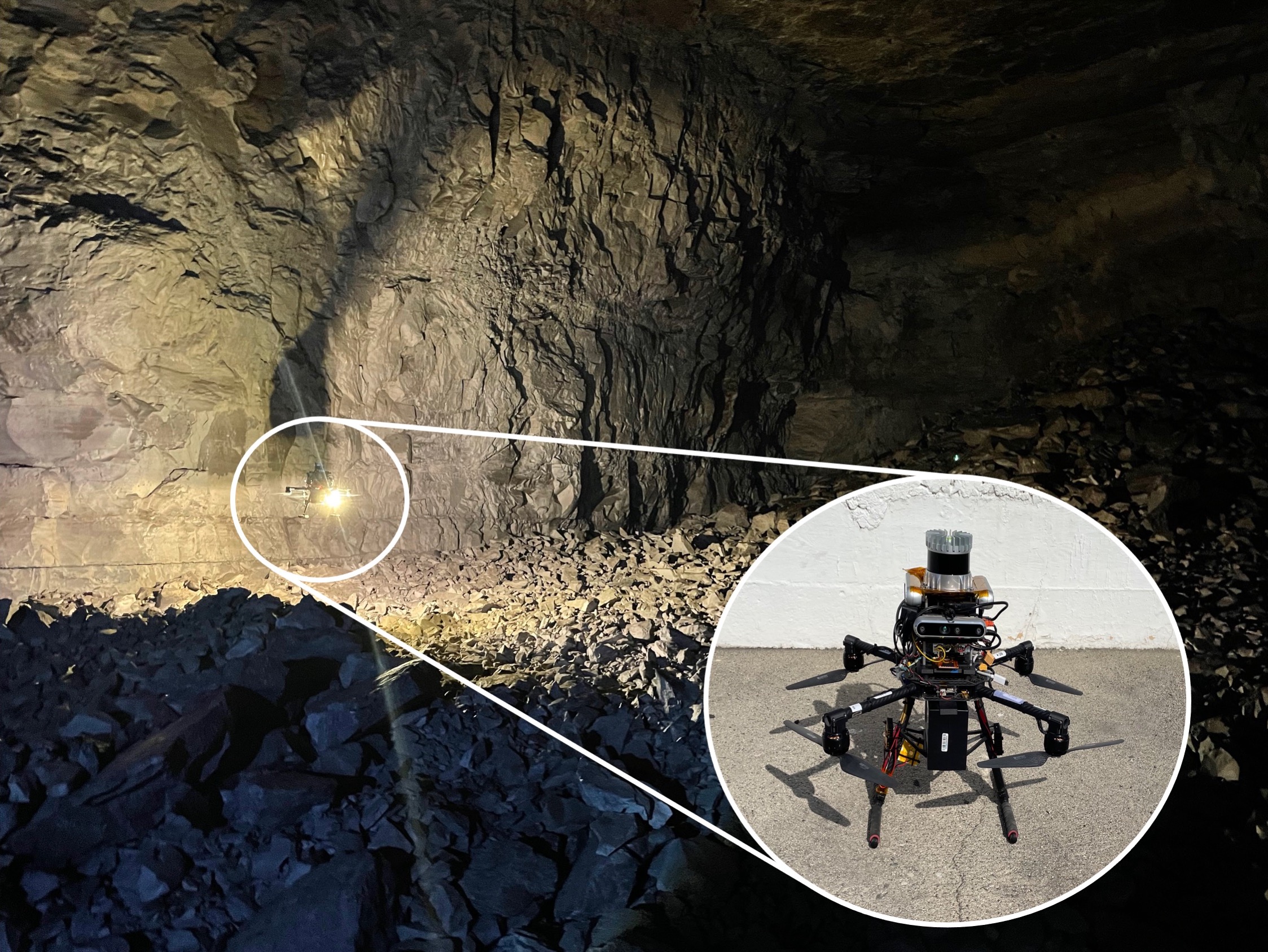}
    \caption{Aquila (customized platform) autonomously exploring a large-scale underground environment.}
    \label{fig:aquila_flying}
\end{figure*}

Aquila (Fig.~\ref{fig:aquila_flying}) is a custom quadrotor equipped with on-board sensors and a computer for fully autonomous navigation and artifact detection. Its main design objectives are compactness, ease of manufacturing, and long endurance. It is based on a modified COTS frame with a 45 cm motor-to-motor diagonal using 12 inch diameter carbon fiber propellers. Upside down motor mounts, with the propellers in a pusher configuration, maximize aerodynamic efficiency and reduce vibration. All the sensors and the computer are mounted on a detachable platform that is vibration isolated from the main quadrotor frame to reduce vibrations further. The detachable autonomy hardware stack simplifies swapping components between the UAVs. The quadrotor weighs 5.3 kg, including a 22 Ah 6-cell LiPo battery, and has a flight endurance of up to 20 minutes.

The sensor suite consists of a spinning Ouster OS-1 LiDAR for localization, mapping, and collision avoidance, a VectorNav VN-100 IMU, and three Intel Realsense D455 cameras facing forward, right, and left for artifact detection. The cameras are paired with powerful, dimmable 14 W LED lights to illuminate dark environments. An Intel NUC (7i7DNBE) is running the autonomy stack with a Pixracer autopilot running PX4 to provide attitude control.

\begin{figure*}[t!]
    \centering
    \includegraphics[trim={0cm 0cm 0cm 1cm},clip, width=0.8\textwidth]{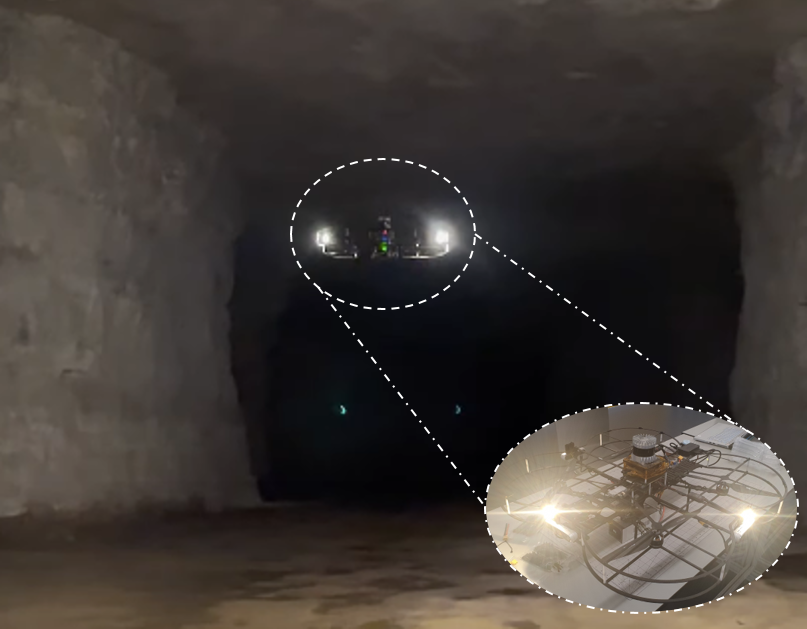}
    \caption{Proxima (customized drone platform) autonomously flying in large-scale cave environment.}
    \label{fig:proxima_flatform}
\end{figure*}

\subsubsection{Proxima}

\begin{figure*}[t!]
    \centering
    \includegraphics[width=0.8\textwidth]{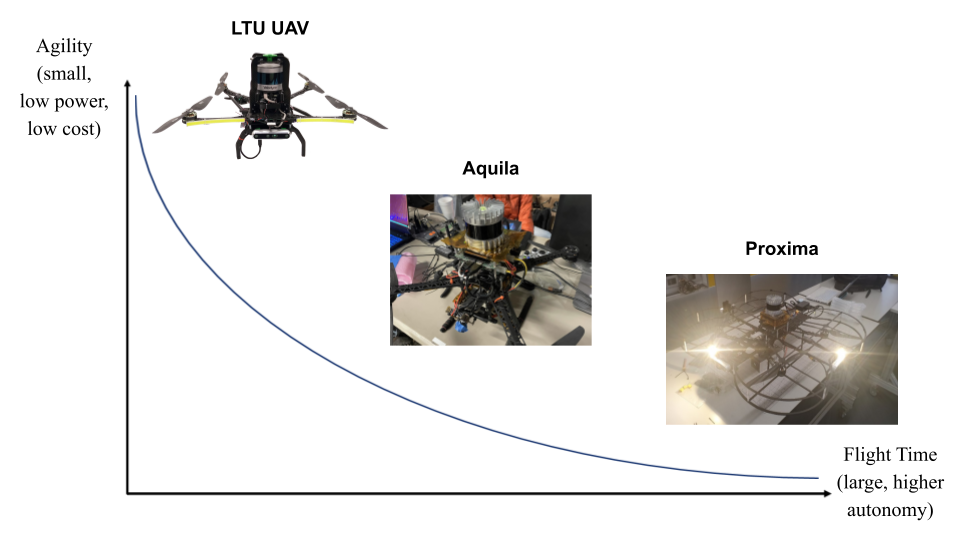}
    \caption{Agility vs Flight Time Curve of Platforms}
    \label{fig:agility_flight}
\end{figure*}

Proxima (Fig.~\ref{fig:proxima_flatform}) is designed to maximize flight time and chance of safe returns in its autonomous flight missions underground. Two carbon sheets were manufactured for the custom quadrotor frame to reduce its weight while using four 15 inch propellers. The propulsion system with 60 cm motor-to-motor diagonal length configuration is powerful and spacious enough that it allows all the necessary sensory and computational components to be mounted onboard. The drone weighing 5.6 kg with its 16 Ah 6-cell LiPo battery attached, shows flight endurance of up to 20 minutes. Fig.~\ref{fig:agility_flight} shows agility of different UAV platforms built during the competition w.r.t their flight endurance.

Similar to Aquila, Proxima uses OS-1 LiDAR and Pixracer's IMU for localization, mapping and path planning. And three Intel Realsense D455 cameras paired with the 14 W LED lights around the propeller guards were used for artifact detection. An Intel NUC is responsible for running the autonomy software stack while a Pixracer R15 stabilizes the drone's attitude.


\subsection{State Estimation}\label{sec:aq_se}

\subsubsection{Aquila}
Aquila's state estimation stack is primarily composed of the Direct LiDAR Odometry (DLO) algorithm \cite{chen2022dlo}, a custom lightweight frontend LiDAR odometry solution with consistent and accurate localization for computationally-limited robotic platforms. DLO uses dense point clouds to provide reliable trajectory estimation in perceptually-challenging environments and is categorized as an LO algorithm loosely-coupled with an IMU. To attain the robustness and performance required for lightweight platforms such as Aquila, DLO consists of several key algorithmic innovations which prioritize efficiency to enable to use of dense, minimally-preprocessed point clouds to provide accurate pose estimates in real-time. This is achieved through a novel keyframing system which efficiently manages historical map information, in addition to a custom iterative closest point solver for fast point cloud registration with data structure recycling. DLO is also built to adapt to differently-sized environments by adjusting the density or sparsity of keyframe placement via a novel spaciousness metric. Code and further details are in \cite{chen2022dlo}, and an improved version of the algorithm can be found in \cite{chen2022dlio}.

The output of DLO consists only of the robot's 6DOF pose, and so for velocity and sensor bias estimates, we coupled DLO with a novel backend filtering-based framework, namely a hierarchical geometric observer. Our observer fused high-rate IMU estimates for our downstream controller and provided several notable benefits: first, fusion with IMU measurements enabled smooth flight control, as an LO algorithm's trajectory output can be quite jittery. Second, our observer provided several theoretical convergence guarantees towards true state estimations over time to provide further robustness and failure protection. This backend coupled with our DLO algorithm provided fast state estimation for Aquila with very minimal system overhead.

\subsubsection{Proxima}
For accurate state estimation of Proxima in a large and complex environment, we used a tightly coupled LiDAR Inertial Odometry (LIO) algorithm. In particular, we focused on keyframe generation because it is considered as the smallest unit of information that can represent a entire map. Considering the importance of the keyframe, we developed Keyframe-centric LiDAR Inertial Odometry (KLIO) that can maintain low computational cost and accurate state estimation performance even for long-term operation. KLIO uses point-based scan matching rather than feature-based scan matching for registration. If all LiDAR measurements are used, the spatial information of the surrounding environment can be fully utilized and the state estimation accuracy can be improved through this.

\begin{figure*}[t!]
    \centering
    \includegraphics[ width=0.98\textwidth]{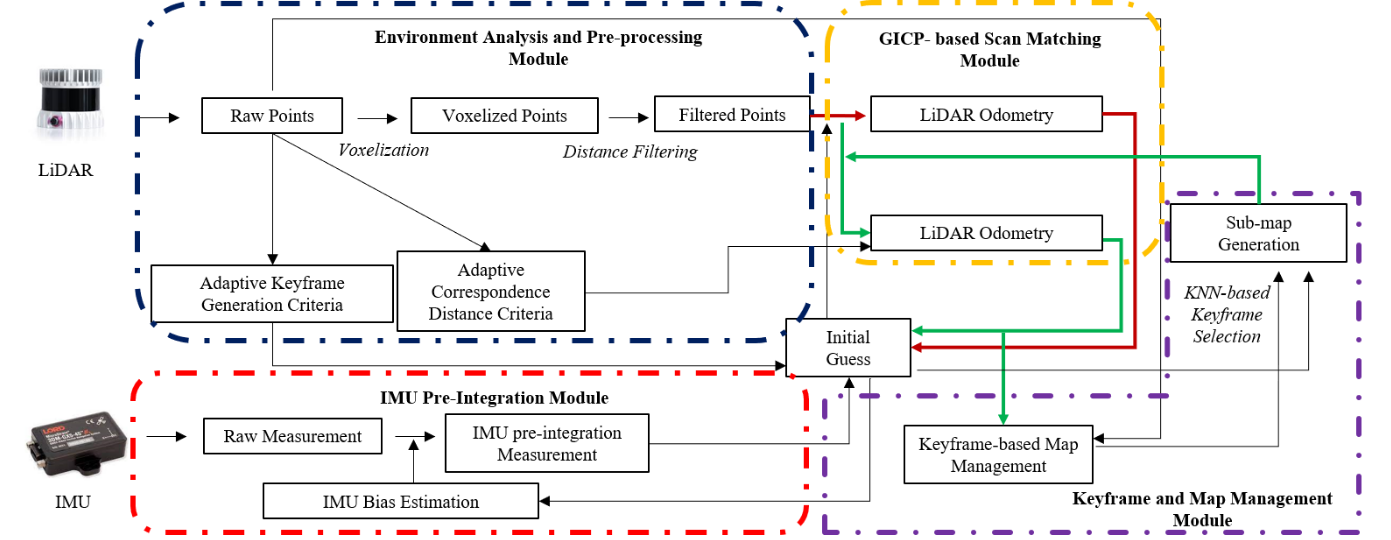}
    \caption{The system configuration of KLIO. The red and green lines in
the scan matching module represent \textit{scan to scan} matching and \textit{scan to sub-map} matching sequences, respectively.}
    \label{fig:proxima_KLIO}
\end{figure*}

KLIO consists of four modules as shown in the Fig. {\ref{fig:proxima_KLIO}}: IMU pre-integration, Environment analysis and pre-processing, GICP-based scan matching, and Keyframe and map management module.

\paragraph{IMU pre-integration module}: The IMU pre-integration module calculates and accumulates relative position, velocity, and rotation changes between the previous IMU frame and the current IMU frame using raw IMU measurement. IMU pre-integration measurements, the output of this module, are used as an initial guess when calculating the relative transformation of two consecutive LiDAR frames in the GICP-based scan matching module. After obtaining LiDAR odometry from the GICP-based scan matching module, the IMU bias is estimated through non-linear optimization.

\paragraph{Environment analysis and pre-processing module}: After obtaining the LiDAR measurements, we first define the wideness and narrowness of the surrounding environment in order to minimize the computation cost of the GICP-based scan matching and increase the state estimation accuracy. For example, in a very wide environment, sufficient geometric feature points are maintained even if LiDAR point downsampling is performed using large voxelization parameters. This means that accurate state estimation through scan matching is possible with sparse keyframe generation. On the other hand, in a very narrow environment, large voxelization parameters may distort the geometric feature points, which may lower scan matching performance. Also, since the environment can change rapidly due to corners or blind spots, the keyframe generation must be very dense. We also changed the correspondence distance parameter according to the wideness of the surrounding environment during scan matching. In a very wide area, the maximum correspondence distance parameter is set large because the keyframe generation interval is very wide and the geometrical common features are maintained for a long time. On the other hand, in a narrow area (stairs or a very narrow corner), the maximum correspondence distance parameter was set small as there were cases in which a common geometrical characteristic was lacking due to a sudden change in the environment. We define five states of the surrounding environment using voxelization results of current LiDAR measurements with voxel size $V_{\text{env}}$: Super wide, Wide, Normal, Narrow and Super narrow.

\paragraph{GICP-based scan matching module}: IMU pre-integration measurements, the output of the IMU pre-integration module, are sent to the scan matching module to calculate the relative transformation of two consecutive LiDAR frames. In order to use sufficient spatial information, we used point-based scan matching rather than feature-based scan matching. The GICP registration method was used for this purpose. The scan matching module consists of \textit{scan to scan} matching and \textit{scan to sub-map} matching that can improve the global scale state estimation accuracy. The \textit{scan to scan} matching calculates the relative transformation between two consecutive LiDAR frames when LiDAR measurements are received, and \textit{scan to sub-map} matching is performed when a keyframe is generated. The robot state estimated through \textit{scan to sub-map} matching is updated with the state reference of the global map scale, and the state estimation between two consecutive keyframes is estimated by accumulating the results of \textit{scan to scan} matching.
\begin{figure*}[t!]
    \centering
    \includegraphics[ width=0.8\textwidth]{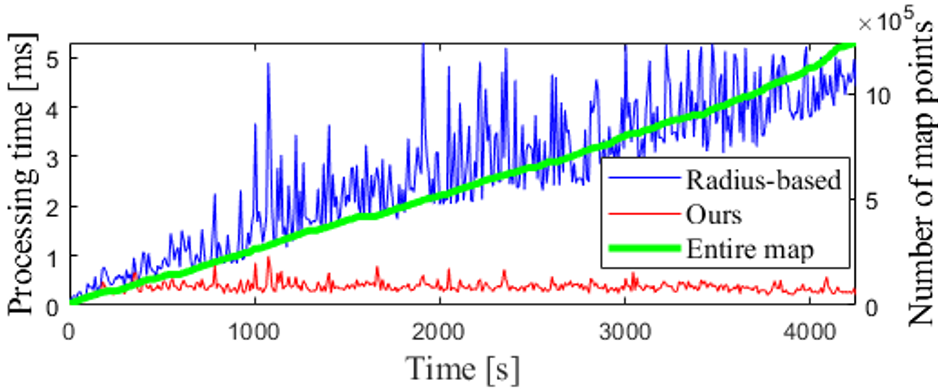}
    \caption{Comparison of processing time for sub-map generation}
    \label{fig:proxima_Submap}
\end{figure*}
\paragraph{Keyframe and map management module}: The keyframe and map management module enables efficient and fast sub-map generation used in the GICP-based scan matching module by using our novel data structure. The \textit{scan to sub-map} matching in the GICP-based scan matching module is an important factor in improving the state estimation performance of KLIO. A conventional sub-map generation method is radius-based filtering, which extracts points within a specific radius based on the current robot position. However, this method has a problem in that if the radius is increased for more spatial information, the computation cost is high. On the other way, if the radius is decreased for a small computation cost, sufficient spatial information cannot be used. To solve this problem, we used a keyframe-centric sub-map generation method. The keyframe and map management module manages keyframes and their corresponding LiDAR points at once. When a new keyframe is generated, sub-keyframes and the corresponding LiDAR points are extracted very quickly using $k$NN algorithm. Compared to the conventional radius-based sub-map generation method, the $k$NN-based sub-map generation method greatly improves the computation cost as shown in the Fig. {\ref{fig:proxima_Submap}}, enabling KLIO to have stable performance even if the map grows over time.

We implemented KLIO to Intel NUC computer with 6 core i7-10710U CPU, and conducted more than 70 flight experiments in the LA subway and Kentucky cave to prepare for the DARPA Subterranean Challenge Final Event. Finally, KLIO was used for autonomous flying of Proxima in the prize round of the DARPA Subterranean Challenge Final Event.

\subsection{Autonomy}

\subsubsection{Aquila}

\paragraph{Local collision avoidance and navigation}

The local mobility stack includes obstacle mapping, local path planning, collision-free trajectory generation, and trajectory tracking. The input to the local mobility stack is a waypoint to navigate to. Waypoint acceptance, timeouts, and higher level mission constraints such as battery life and mission progress are handled in the higher levels of the stack. The local mobility has to consider the perceptual limitations of the robot to avoid moving in blind directions. Notably, the UAV cannot climb or descend at steep angles since the LiDAR field of view is limited. This constraint is considered in the various layers of the local mobility stack.

Local obstacles are mapped in a 3D voxel map that follows the robot in a sliding window fashion. The map size and resolution are tuned to the scale of the environment in which the robot operates. For example, for large underground mines like in the Fig.~\ref{fig:aquila_flying}, a map size of 40 x 40 m wide and 6 m tall at a uniform 0.5 m resolution provided a good tradeoff between computational cost, scale of local obstacle memory, and size of the smallest opening the UAV can pass through. The local obstacle map is built by ray-tracing the returns from the scanning LiDAR. Obstacles and free space are treated probabilistically, aggregating LiDAR hits and passes over time. An explicit distinction between unknown and free space is made, which allows the robot to avoid flying into space it has not observed.

On this 3D voxel grid, a cost-to-go is computed for any point in the grid to the waypoint using Dijkstra’s algorithm. Path costs are higher the closer they are to the nearest obstacle. This lets the robot choose paths along the center of a corridor, thus increasing safety and improving artifact detection probability. If the waypoint is not inside the map, it is projected to the boundary of the map in the horizontal directions. Unknown space is treated as traversable but with high path cost.

The trajectory generation is based on dynamically feasible motion primitives. The motion primitive trajectories are parametrized as a polynomial starting at the current state and ending at a fixed distance of 2 m away from the robot and at a fixed time horizon of 4 seconds. The azimuth and elevation of the terminal point are varied for the different motion primitives. The motion primitives start at the nominal state from the previous trajectory instead of the current state estimate. This makes the planning robust to odometry and tracking errors. Obstacles are inflated to account for estimation and tracking uncertainty. The motion primitives are collision checked against the instantaneous LiDAR scan for fast reaction to dynamic obstacles as well as obstacles from the local voxel map. In this case, unknown space is treated as an obstacle to avoid planning trajectories that fly blindly into unobserved areas of the map. This and the fact that we pick terminal points for the motion primitives at shallow climb/descend angles keep the robot safe even with perceptual blind spots in the vertical direction. All motion primitives that are not colliding are then ranked by their terminal cost-to-go and the lowest cost trajectory is executed. Finally, the trajectories are tracked using a position and velocity tracking controller, which outputs the desired attitude and thrust to the PX4 autopilot in a hierarchical control architecture.




\subsubsection{Proxima}

\begin{figure}[!tbp]
    \centering
    \includegraphics[width=.9\columnwidth]{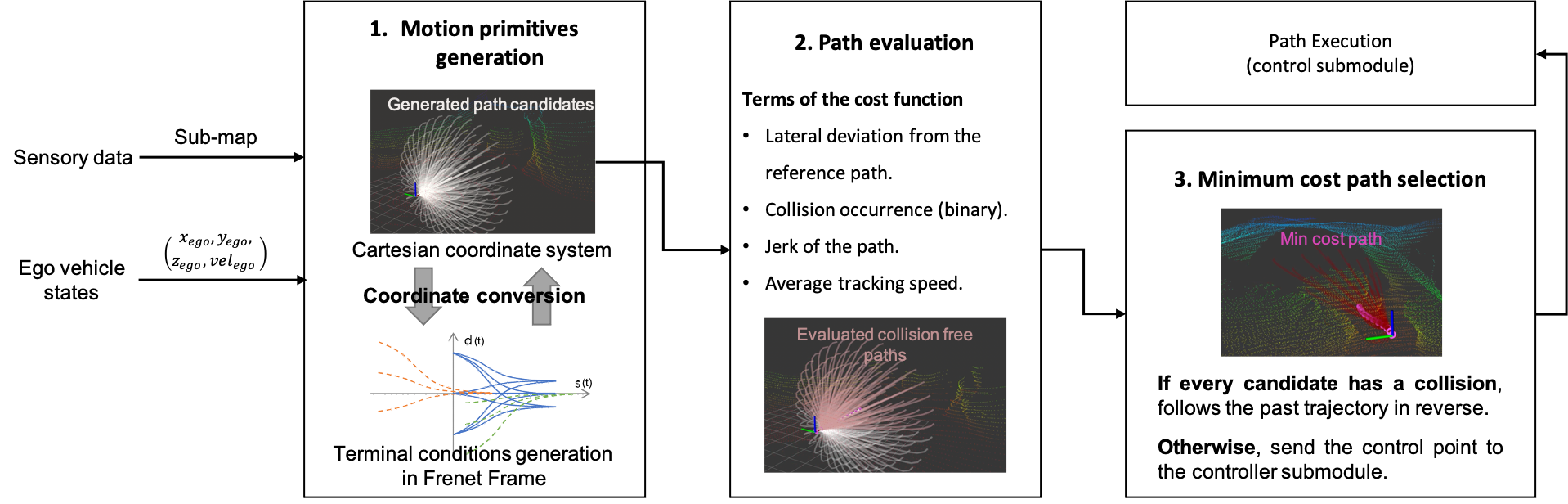}
    \caption{Overall pipeline of Proxima's local trajectory planner.}
    \label{fig:proxima_lpp}
\end{figure}

\paragraph{Local collision avoidance and navigation}
Proxima's local trajectory planner is responsible for generating control signals using motion primitives generated around the reference path provided by the higher task level planner. Instead of a fixed state space, our motion primitives are generated during flight through a set of predefined polynomial functions (here, we set third order) in Frenet-Serret coordinates. The planning horizon and sampling time interval were set to 2 and 0.2 seconds, respectively. The local trajectory planner didn't allow generation of trajectory over certain threshold of the jerkiness(i.e. the derivative of the acceleration) to prevent sudden thrust of the vehicle which can significantly degrade the mapping performance. Fig.~\ref{fig:proxima_lpp} shows the overall pipeline of Proxima's local trajectory planning module. 

Generated motion primitive are passed to the path evaluation module and assigned the cost by using linearly-weighted function. The terms of our cost function are lateral displacement from the reference path, jerkiness, average tracking speed, and collisions. Here, we set the weight of the collision to infinity. We calculated the euclidean distance between the lidar based sub-map data and points of the target primitives for collision checking. In consideration of the dimensions of the vehicle and the performance of the controller, we set 0.6 meter as a collision threshold value. When the distance between the two points from sub-map and target primitive is smaller than the threshold, we decide there is a collision and the primitive is removed from the candidates. In order to minimize the computation cost of the collision detection problem, which has to be repeatedly calculated as much as the number of points constituting the sub-map and motion primitives, we conducted collision checking in the order closest to the reference path. Therefore, as soon as it is determined to be collision-free and lower than the certain threshold of the cost, the trajectory is transmitted to the controller for vehicle control. On the other hand, when all motion primitives over the cost threshold, the vehicle follows the past trajectory(sequence of the past position) in reverse, and switch to standby mode for the next high-level task.

\paragraph{Returning to home and failsafe behaviors}:
Proxima's high-level mission planner stack for safe and reliable navigation includes several fail safe behaviors such as waypoint integrity checking, Returning To Home (RTH), time out, and emergency landing. Operator uses Wi-Fi communication to transmit global waypoints from the base station to Proxima's onboard computer. When Proxima receives operator's start command, Arming, Take-off, and Offboard mode in the PX4 autopilot are automatically executed in sequence, and autonomous flight is performed with the received global waypoints. When Proxima arrives at a specific radius of each waypoint, the goal position is changed to the next waypoint. When it arrives at the last waypoint, the goal position is changed in the reverse order of the waypoints, and finally, Landing and Dis-arming are performed.

The goal of waypoint integrity check, one of the high-level mission planner stacks, is to verify whether the current goal position is reachable in the local map. This is because the reachability of waypoints is uncertain due to the navigation characteristics of Proxima, which must fly in an unknown subterranean environment. Using the map generated through KLIO, we generate a local map using a radius-based point extraction algorithm according to the current position. In particular, we focused on the z value (altitude) of the goal position. If the goal altitude is higher than the maximum z value ($z^{\text{local}}_{\text{max}}$) or lower than the minimum z value ($z^{\text{local}}_{\text{min}}$) of points in the current local map, the goal altitude is set to $2*(z^{\text{local}}_{\text{max}} + z^{\text{local}}_{\text{min}})/3$. If the horizontal value (x and y values) of the goal position is larger than the radius used when generating the local map, it is projected to the corresponding radius, and if it is occupied, it is set to the nearest empty space.

The Returning to Home (RTH) mode is an essential failsafe factor for the operation of unmanned vehicles for safe and reliable exploration. We performed RTH mode in three situations. The first is when Proxima arrives at all received waypoints. This is to ensure that Proxima can safely return to its home position even in the case of environmental uncertainty because it is impossible to simply track the trajectory that has flown in a non-static environment. The second is when all waypoints are canceled due to time out. Time out, one of Proxima's fail safe modes, was used to prevent errors in path planning results. Using the distance from the current position of Proxima to the goal position and the desired velocity of Proxima, we calculate the reachable time and set the time out parameter by multiplying it by a specific safety constant. This allows Proxima to cancel all waypoints due to a time out and safely return to the home position in an environment that is physically impossible to fly. The third case is performed due to the battery state. Proxima monitors the battery status in real time during autonomous flight and calculates whether sufficient battery capacity is guaranteed when returning the same distance. The battery capacity parameter for RTH mode we used for the Proxima is 40\%, and it was set conservatively considering that the trajectory may be changed by an uncertain environment when returning to the home position.

Emergency landing mode is performed for the physical safety of Proxima in case of unrecoverable situations such as algorithm failure in the navigation software stack. The built-in functions of the PX4 autopilot are used. First, if the KLIO node dies or state estimation is impossible in real time, it will automatically land. Second, when the autonomous landing function mapped to the wireless controller is activated. This allows the operator to make an emergency landing of Proxima using a wireless controller. Finally, using the Xbee, a basic requirement of the DARPA SubT challenge, Proxima can make an emergency landing. 

By using these various failsafe functions, we can cope with  internal and external emergency situations of Proxima.

\subsection{Artifact Detection, Scoring, and Game Play}

The drone platforms enable detection of artifacts that are inaccessible to ground robots due to traversability, obstacles, or elevation.
Considering the challenges of long duration flight times, a game play strategy of allowing ground robots to scout out regions only accessible by aerial robot was developed.
After the fleet of ground robots has been deployed, the operator can direct either or both of the drones towards any under-explored or otherwise interesting regions.
The operator can choose which drone to deploy depending on the expected environment, with decision based on the significant scale difference between the Aquila and Proxima drones.

The artifact detection and positioning for both the Aquila and Proxima drones rely on the methods outlined in Section \ref{sec:artifacts}, including neural network and 3D depth-based artifact positioning.
With the majority of computational resources on the drones dedicated to operations and SLAM, additional computational hardware would be required to detect and position artifacts in real time at any reasonable frame rate.
Furthermore, more additional hardware would be required to stream detections over radio to the operator's base station.
In the interest of keeping the drone platform hardware minimal, a deploy and return model was developed for the drones during gameplay.
In this model, the drone logs all RGB camera streams, depth data from stereo cameras and lidar, and positioning and mapping results from the SLAM methods on a removable SSD.
When the drone battery drops below a threshold, the drone autonomously returns to the starting location and traverses the gate in reverse so the field ops team can access the SSD.
The data logs are then processed at the base station and positioned artifacts are sent to the operator to approve or reject in the same manner as for the ground robots.

\subsubsection{Gate traversal and alignment}
Gate alignment is necessary to provide the fleet of robots (both ground and aerial) with a consistent global coordinate system. The predefined gate consisted of several reflective fiducials (roughly 0.5m x 0.5m) on each side which could be used for alignment and calibration. During calibration, LiDAR points of high intensity are assumed to be from gate fiducials, and all other points are filtered out. After further filtering through a statistical outlier filter to remove noisy points, high intensity points are split into left and right groups by looking at the sign of each point's $y$ component. This is only possible if the aerial vehicle roughly points in the center of gate's opening. Planes are subsequently fit onto each left and right set of points to determine centroid points, and simple geometry is used to calculate position and yaw relative to the center of the gate.

To traverse the gate, we found that a pre-programmed, open-loop behavior (i.e., no collision avoidance) was easiest since the gate was a static and of known dimension, and placement of our aerial vehicle was done by the operator and introduced very little variance to the procedure. Once the drone hovered 1.5m off the ground and moved roughly 5m forward, behavior management, collision avoidance, exploration, and other autonomy modules started and the mission could be carried out. 

\subsection{On the Unification of Spot and UAVs} \label{subsec:Unification}
\begin{figure}[!htbp]
    \centering
\includegraphics[width=0.4\linewidth]{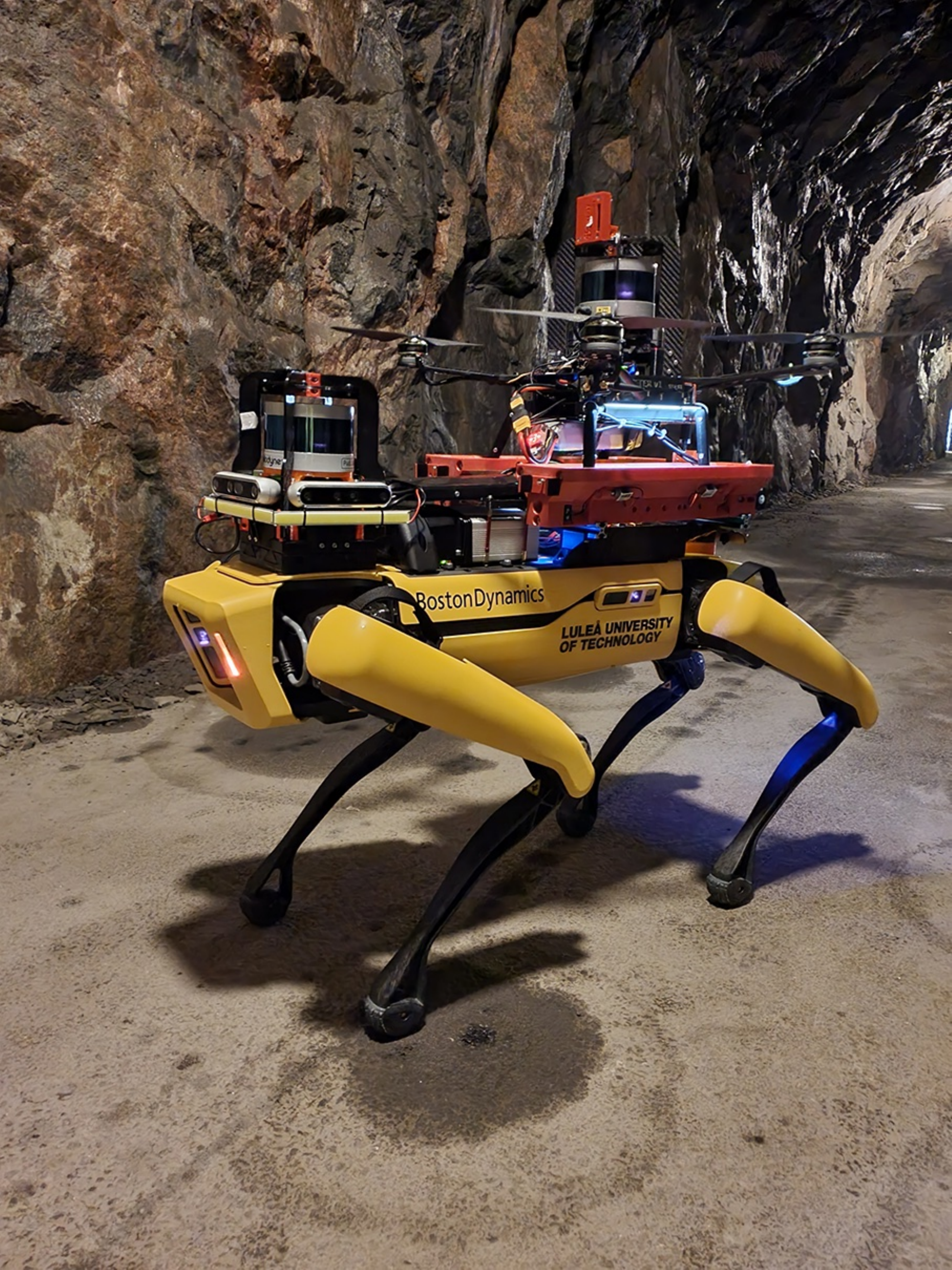}
    \caption{The legged-aerial explorer with its full sensor suite and the UAV carrier platform.}
    \label{fig:robots}
\end{figure}

This subsection presents an autonomous unified legged-aerial robotic system for subterranean exploration, exploiting the multimodality properties of the two robotic systems for rapid deployment in mapping and data collection application scenarios. The Boston Dynamics (BD) Spot legged platform is tasked to be the carrier of an aerial platform, since its locomotion merits on rough terrains and long-lasting endurance complement the short flight time and large reachability properties of the vertically navigating aerial system in a three dimensional space. As the need for autonomy requirements, mission complexity, and abilities in handling unforeseen terrain challenges increases, new multimodality approaches to enable hybrid locomotion will be on the forefront of critical mission applications in terrestrial subterranean environments, but also in the search for life in subterranean caverns on other planets~\cite{nikolakopoulos2021pushing}. Solving these challenges is the primary justification behind the proposed research efforts into legged-aerial robot unification with the long-term goal of extending the mission capabilities of existing robotic solutions when entering completely unknown and unknowable environments and their related terrain traversability challenges. A field-hardened version of the unified robotic solution is depicted in Fig.~\ref{fig:robots}.

\begin{figure}[!htbp]
    \centering
\includegraphics[width=1\columnwidth]{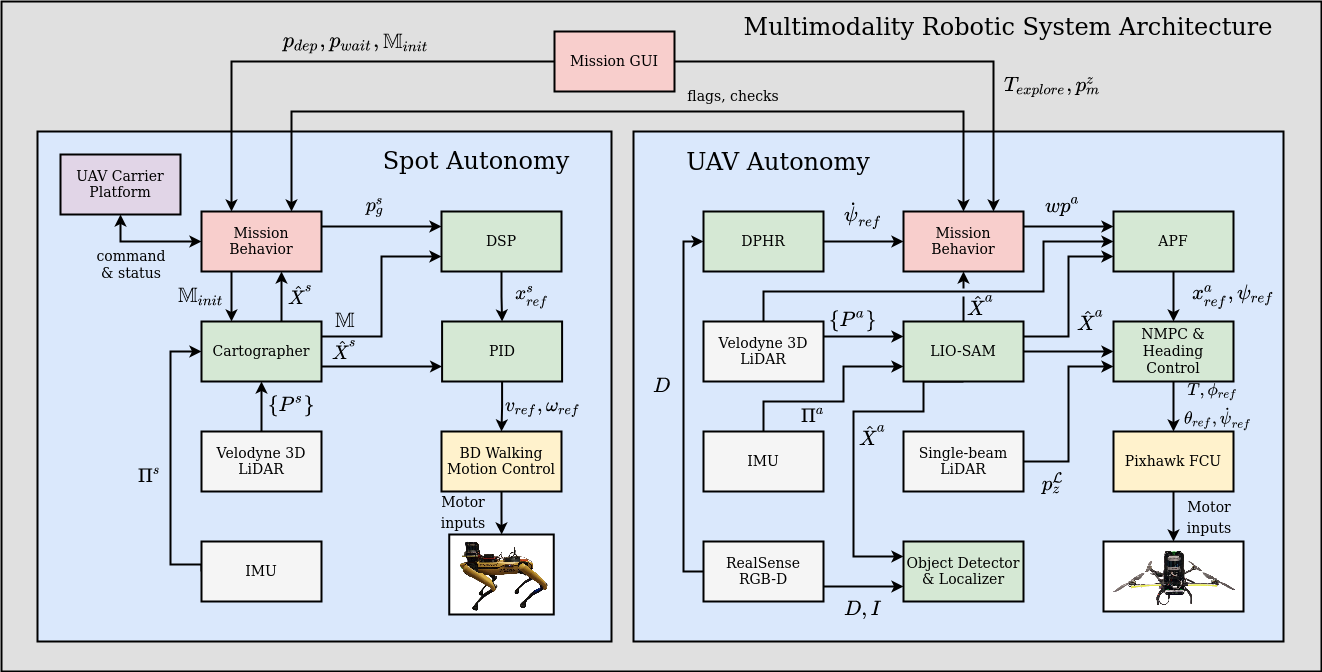}
    \caption{Complete autonomy architecture used for mission execution.}
    \label{fig:spot_uav_arch}
\end{figure}

A high-level architecture of the robotic autonomy system can be found in Fig.~\ref{fig:spot_uav_arch}. We denote aerial agent parameters with superscript \textit{a} and Spot parameters with superscript \textit{s}. 

\subsubsection{Shafter} \label{subsec:shafter}
\begin{figure}[!htbp]
    \centering
\includegraphics[width=0.45\columnwidth]{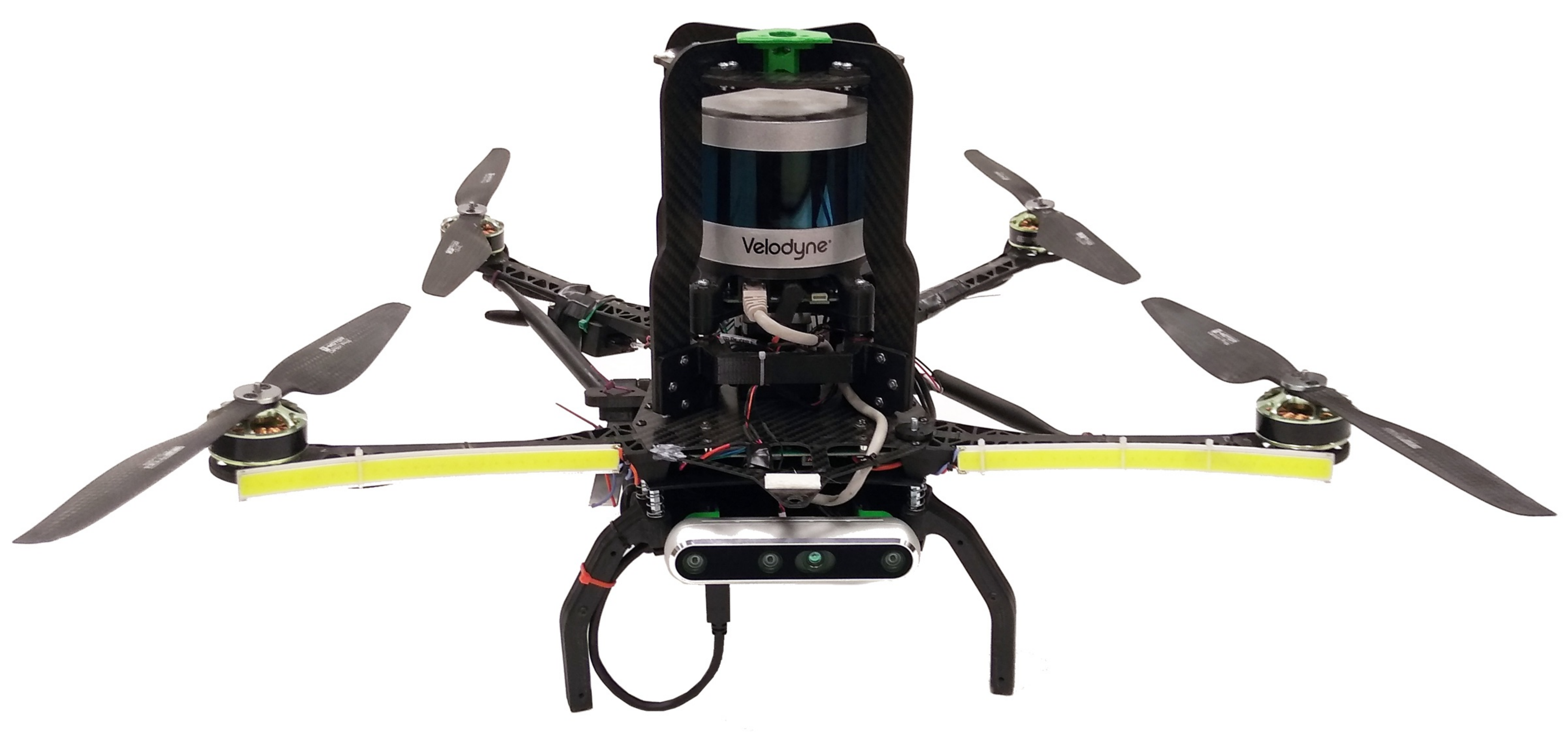}
    \caption{The custom built quadrotor Shafter.}
    \label{fig:shafter_platform}
\end{figure}

Shafter is a custom built quadrotor platform, as shown in Fig.~\ref{fig:shafter_platform}. The platform has maximum size diameter of $0.85$~m and mass around $3.5$~kg. It is equipped with the Intel NUC - NUC10i5FNKPA processor for all the onboard autonomy, while it also carries the Intel Neural Compute Stick 2 to offload the artifact detection modules. Additionally, the sensor payload includes the 3D LiDAR Velodyne VLP16 PuckLite used in the obstacle avoidance module, as well as in the localization module, combined with Inertial measurements, the single beam LiDAR LiDAR-Lite v3 facing towards the ground for relative altitude measurements and the RGB-D camera Intel Realsense D455 used in the object detection and heading regulation modules. The MAV is equipped with the low-level flight controller PixHawk 2.1 Black Cube, which provides the IMU measurements. Finally, the MAV is equipped with two $10$~W LED light bars in the front arms for additional illumination and one $10$~W LED light bars looking downwards. The platform has a flight endurance up to 8 minutes with the full payload. 

\subsubsection{Design of the Carrier Platform}
The mechanical unification of the two robotic modalities is realized based on the design and implementation of a structure mounted on the legged robot and is referred as the UAV carrier platform. The UAV carrier platform design fulfills requirements, such that the platform is able to allow for the take-off and landing of the aerial vehicle. At the same time, the UAV can be securely locked in the platform during the motion of the legged robot. 

The UAV carrier platform is designed with a passive drone landing alignment and an active landing gear locking. The core components are two rails, which have openings of 90~degrees in the $X$ axis and 45 degrees in the $Y$ axis. Based on lab verification experiments, the design is able to compensate misalignment errors at landing of 80~mm along the $Y$ axis and 31~mm along the $X$ respectively.

\begin{figure}[H]
\includegraphics[width=0.95\columnwidth]{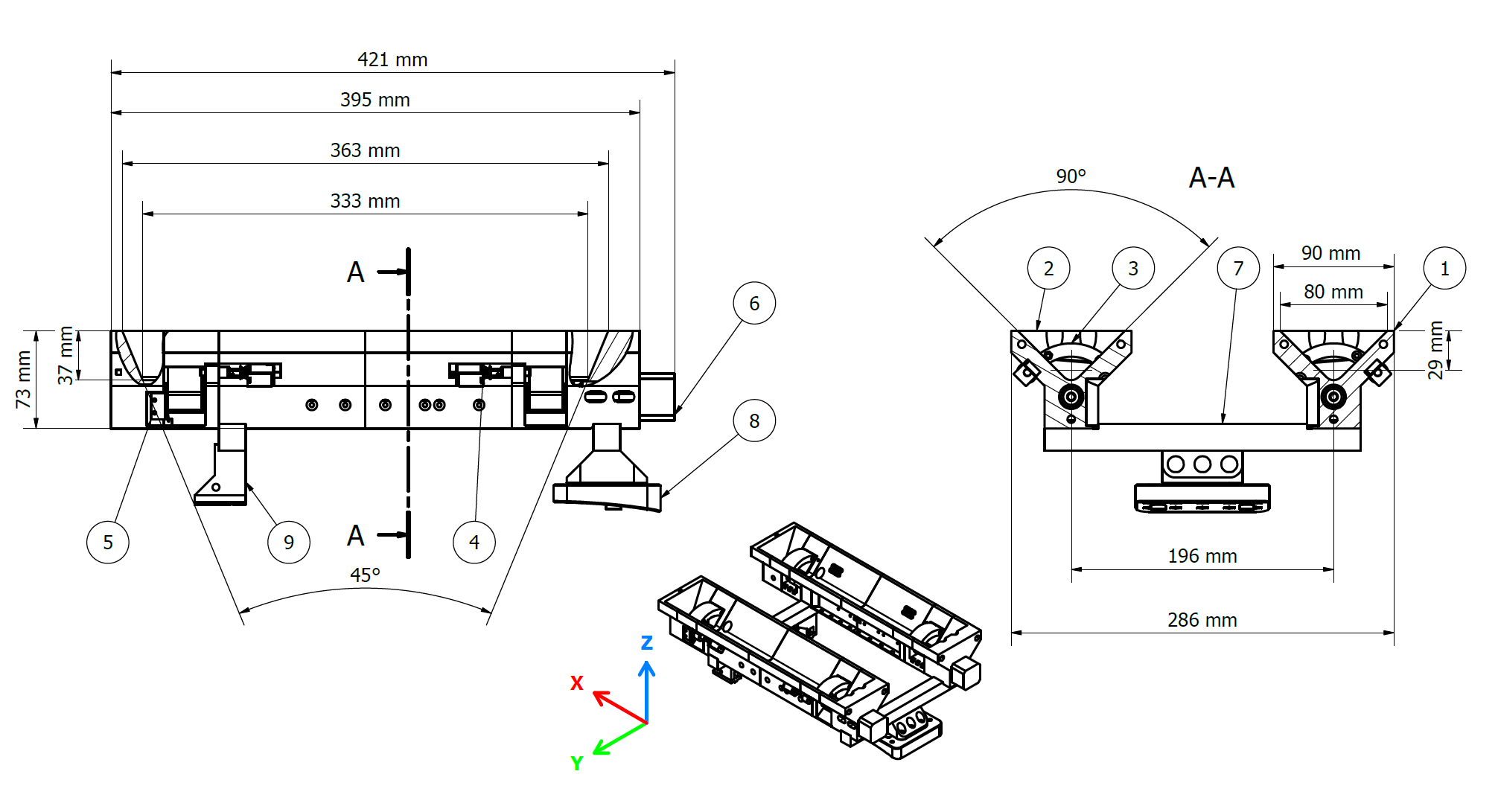}
    \caption{The 
    dimensions and components of the landing platform for the Shafter drone: 1---left V-rail, 2---right V-rail, 3---locking clamp, 4---door-lock-style solenoid, 5---end stop, 6---stepper motor NEMA14, 7---aluminum profile frame, 8---rear Spot mount plate, 9---front Spot T-slot mount.}
    \label{fig:Platform-mounts}
\end{figure}

The platform is attached to the BD Spot robot using the payload T-slot rails in the front and custom-made rear mount plate in the back, as shown in Fig.~\ref{fig:Platform-mounts}. A notable point in the design is the rear mount plate, which modifies the rear body panel screws, allowing secure connection as well as providing substantial clearance above the rear shoulders during the robot motion and it's transition to and from resting mode.
\vspace{-6pt}

The active locking is represented by a pair of rotating clamps in each rail, which are controlled by the NEMA14 stepper motor. Each clamp is secured in the locked state by a door-lock-style solenoid, while the clamps are closed and secured with the solenoids. The schematics of the landing platform are depicted in Fig.~\ref{fig:Platform-mounts}.
The locking and unlocking sequence software is operated within the Robot Operating System (ROS) framework.

The presented design platform was tailored for a UAV which has dimensions, excluding the 12-inch propellers, are $W \times L \times H$, 513~mm $\times$ 335~mm $\times$ 339~mm, and the total weight is 3615 grams. Nevertheless, this landing platform design can be generalized and used with other types of UAVs following appropriate integration steps, mainly by redesigning their landing gear. More details of the design, as well as other designs can be found on~\cite{haluvska2022unification}.

\subsubsection{UAV Autonomy}
The UAV autonomy is structured around the COMPRA (COMpact and Reactive Autonomy) framework~\cite{lindqvist2022compra}, a tailored-made navigation scheme for autonomous subterranean tunnel missions. The framework merits support rapid deployment, based on low-complexity and resilient reactive autonomy baseline navigation architectures. COMPRA uses the state-of-the-art LiDAR-Inertial Odometry (LIO-SAM)~\cite{shan2020lio} for 3D state estimation, using LiDAR pointcloud $\{P^a\}$ and IMU sensor data $\Pi^a$ as inputs, and produces the state vector as $X^a=[p_x,p_y,p_z,v_x,v_y,v_z,\phi,\theta, \psi, \omega_x, \omega_y, \omega_z]^\top$, e.g. position, velocity, angular, and angular velocity states. Moreover, the UAV body-frame state vector is defined as the rotation of position and velocity states using the yaw angle $\psi$ as $X^{a,\mathcal{B}}=[p^\mathcal{B},v^\mathcal{B},\phi,\theta, \psi, \omega]^\top$. Additionally, the downward-facing single-beam LiDAR provides a local z-coordinate measurement (e.g. distance to the ground) as  $p^\mathbb{L}_z = R_{sbl}\cos{\theta}\cos{\phi}$, where $R_{sbl}$ is the range measurement. 

The flight control uses a cascaded scheme with a high-level nonlinear model predictive reference tracking controller (NMPC)~\cite{small2019aerial, lindqvist2020nonlinear} and the Pixhawk Cube\cite{meier2011pixhawk} for the low-level attitude control. The NMPC considers $x_{nmpc} = [p^\mathcal{B},v^\mathcal{B},\phi, \theta]$, and thus, operates in the UAV body frame. The generated control inputs are $u^a = [T_{\mathrm{ref}}, \theta_{ref}, \phi_{ref}]$ with $\phi_{\mathrm{ref}}\in \mathbb{R}$, $\theta_{\mathrm{ref}}\in \mathbb{R}$ and $T_{\mathrm{ref}}\geq 0$ to be the references in roll, pitch and total mass-less thrust generated by the four rotors together with a yaw-rate command $\Dot{\psi}_{ref}$ are sent to the low level attitude controller, with $T_{\mathrm{ref}}$ mapped to a control signal as $u_t \in [0,1]$~\cite{meier2015px4}. The yaw angle $\psi$ is controlled with a decoupled PD controller. 

The COMPRA exploration behaviour is based on two components: a reactive Artificial Potential Field (APF) formulation, based on instantaneous 3D LiDAR pointcloud raw data to keep a safe distance from obstacles or walls, and a heading regulation technique that aligns the body-frame with the tunnel axis based on clustering depth points of the onboard RGB-D sensor, denoted as the Deepest-Point Heading Regulation (DPHR). More details on the presented architectures can be found on~\cite{lindqvist2022compra}.

\subsubsection{Spot Autonomy}

Spot comes with a local autonomy package, that accepts velocity and angular velocity commands and performs local avoidance. Nevertheless, these built-in capabilities of Spot are not sufficient for long and complex autonomous missions or missions in unknown environments. The proposed autonomy archtecture, enables the navigation towards the deployment point ($p_{dep}$) based on an a priori known map $\mathbb{M}_{init}$. The employed mapping framework uses the SLAM package Cartographer~\cite{hess2016real}, due to its capability to map an area and relocalize on that map. Cartographer operates in two steps, the first is an offline map building step and the second step is an online relocalization and mapping algorithm. The fist step uses recorded IMU and LiDAR data (from a previous mission) to build $\mathbb{M}_{init}$ and store for future use. Cartographer's second stage loads $\mathbb{M}_{init}$, and relocalized within $\mathbb{M}_{init}$ which afterwards continuously expands/updates the map $\mathbb{M}$.

To navigate from current state $X^s$ to goal state $p_g^s$ the risk-aware grid search algorithm is used, denoted as D$^*_+$ (DSP). DSP extends the D$^*$-lite ~\cite{dslgridsearch} path planning algorithm by adding a consideration of unknown space, a risk-aware layer, and an update-able/expandable map.
The construction of DSP's internal gird map $\mathbb{G}$ is build from $\mathbb{M}$ in such a way that each cell $c$ in $\mathbb{G}$ has a traversal cost $\zeta$ corresponding to whether $c$ is free ($c_f$), occupied ($c_o$) or unknown ($c_u$).
DSP plans a path $\vec{P}_{X^{s}} \to p_g^s$ based on $\zeta \in \mathbb{G}$ so that $\sum{\forall \zeta \in \vec{P}_{X^s} \to p_g^s}$ is as small as possible.

\subsubsection{Mission Behavior on the Unification}
\begin{figure}[!htbp]
    \centering
\includegraphics[width=0.5\columnwidth]{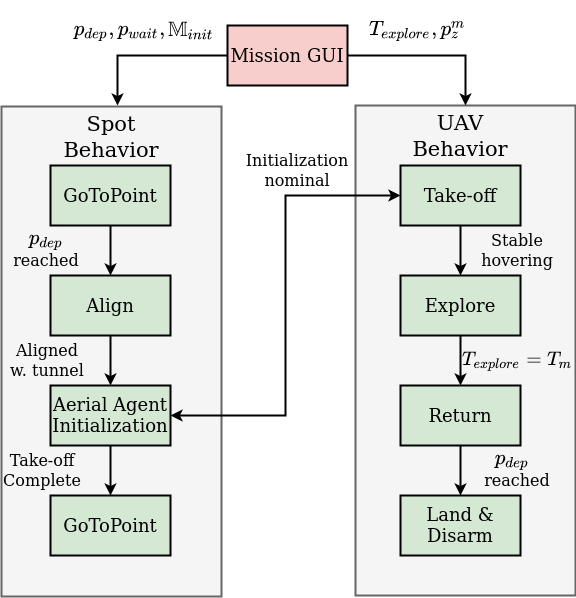}
    \caption{Behavior Lists for specific mission execution for the aerial-legged robot.}
    \label{fig:behavior_list}
\end{figure}

The software unification of the robotic system follows a baseline mission behavior list. Each block of the mission list is executed in order and considered completed by a condition (e.g. the mission time $T_m \geq T_{explore}$, or the robot reaching the desired deployment point $p_{dep}$). The designed behavior lists for baseline mission execution can be found in Fig.~\ref{fig:behavior_list}.

%% file: sections/11.field.tex
\section{Experiments} \label{sec:field}


\subsection{Field Tests and Demonstrations}
\begin{figure}[hbt]
    \centering
    \includegraphics[width=0.8\textwidth]{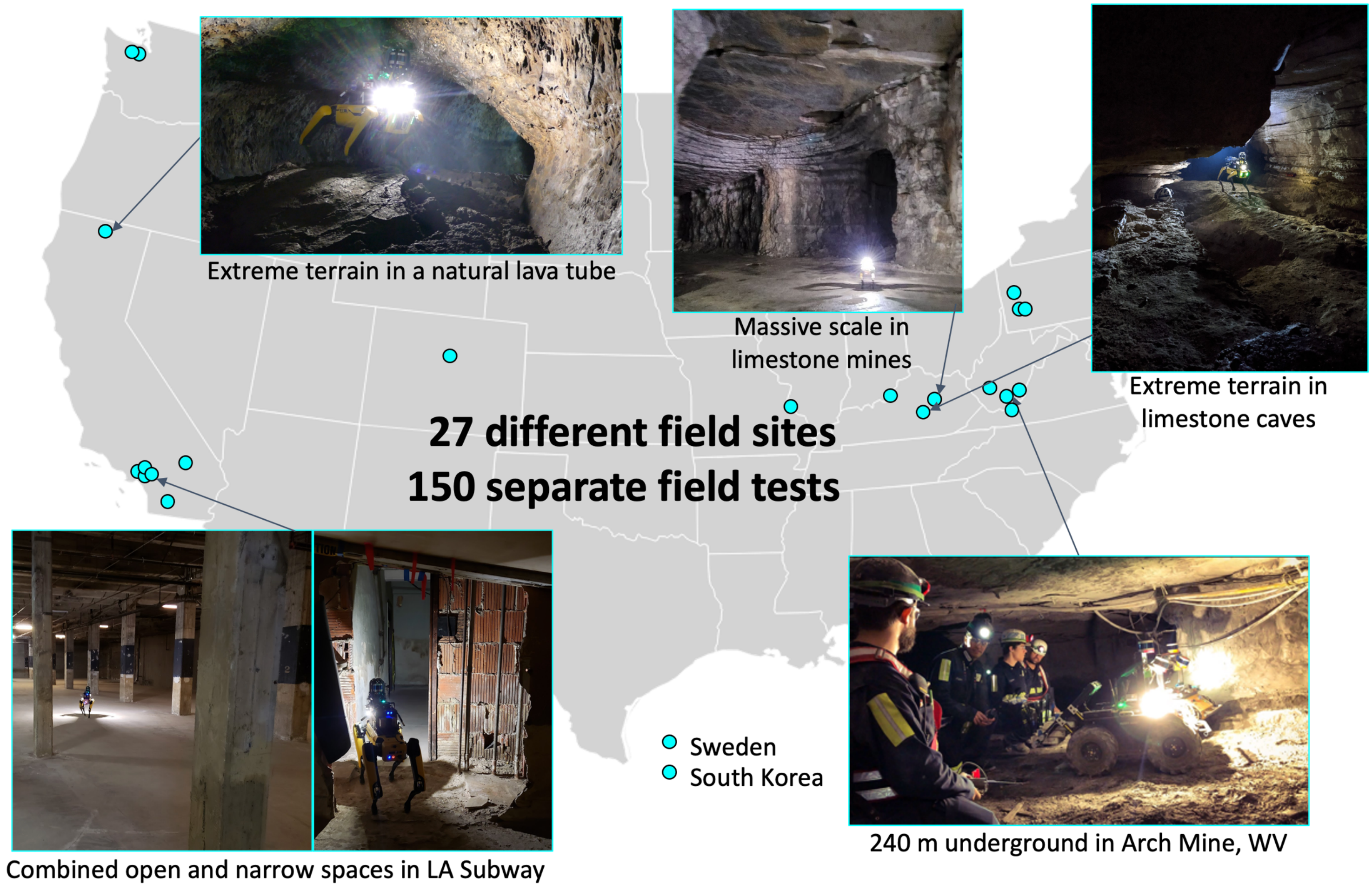}
    \caption{Outline of the different field test locations used by team CoSTAR across the DARPA Subterranean Challenge}
    \label{fig:field:test_locations}
\end{figure}

A core philosophy of team CoSTAR was to regularly and repeatedly test our systems to rapidly discover the main challenges, iterate, grow capability and build robustness. Overall, the team ran field tests in 27 different field sites with over 150 different field tests, all across the US, and the world (Fig.~\ref{fig:field:test_locations}). 

\begin{figure}[hbt]
    \centering
    \includegraphics[width=\textwidth]{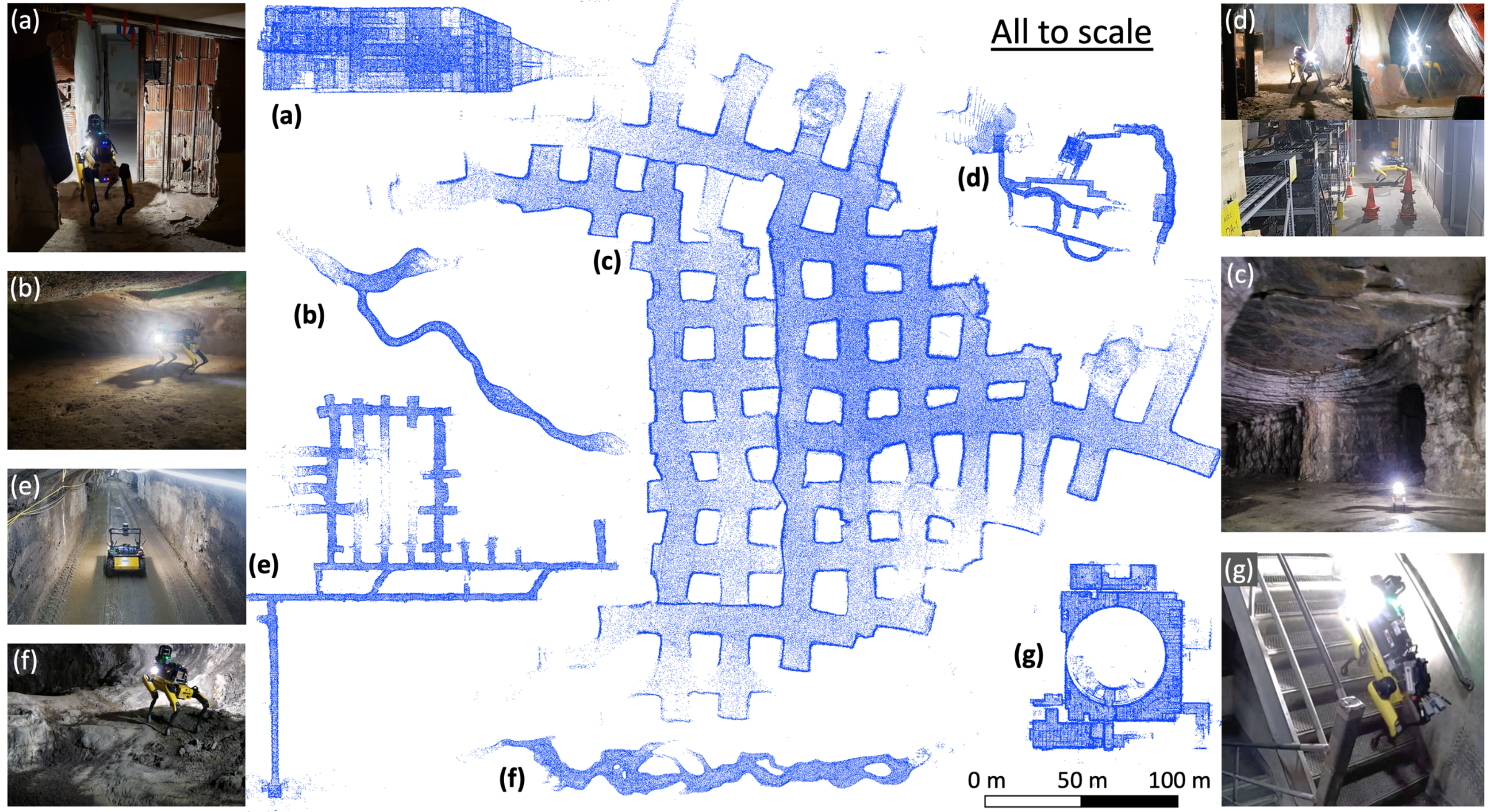}
    \caption{A comparison of the variety of different test environments used by Team CoSTAR, with maps and matched images showing the nature of the environment. a) LA Subway (multi-level offices and halls), b) Lava tube, c) Kentucky Underground (limestone mine), d) DARPA final competition (mixed), e) Bruceton Research Mine (coal mine), f) Valentine Cave (lava tube), g) Satsop Power Plant (factory).}
    \label{fig:field:maps_of_test_locations}
\end{figure}

Another goal of the field test campaign was to test the system in a variety of real-world environments. As is shown in Fig.~\ref{fig:field:maps_of_test_locations}, these different environments included substantial variations in: terrain, scales, cross-sections and density of decision points. This variety stresses the system in different ways to further expose the challenges to solve on the path to a robust, adaptable capability. 

\begin{figure}[hbt]
    \centering
    \includegraphics[width=\textwidth]{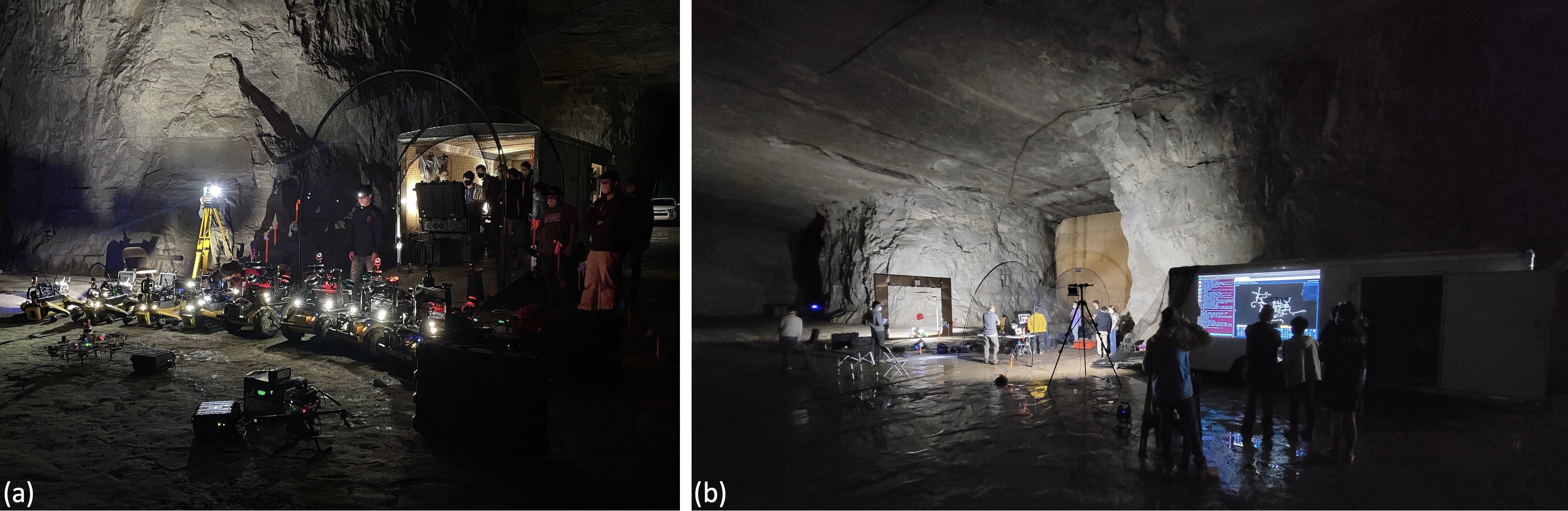}
    \caption{Example test setup for team CoSTAR. a) Full robotic line-up when simultaneously deploying 10 robots (4 spots, 4 huskies and 2 drones) with a single operator in a limestone mine. b) The larger team workspace, including base station projection for team members to observe, and calibration gate to simulate a DARPA course.}
    \label{fig:field:setup}
\end{figure}

A key part of the intensive field campaign was the support personnel, infrastructure and setup to enable efficient and safe testing of robots. An example of this setup is shown in Fig.~\ref{fig:field:setup}, with a complete competition-like setup (with artifact ground truth locations) in a limestone mine, netting to permit drone operations, and separated streaming of base station content to allow spectating in a safe manner. It is a testament to the robustness of the hardware, the efficiency of the robot setup, and the completeness of field equipment that the team was able to regularly deploy 10 robots in under 30 minutes. 

\begin{figure}[hbt]
    \centering
    \includegraphics[width=\textwidth]{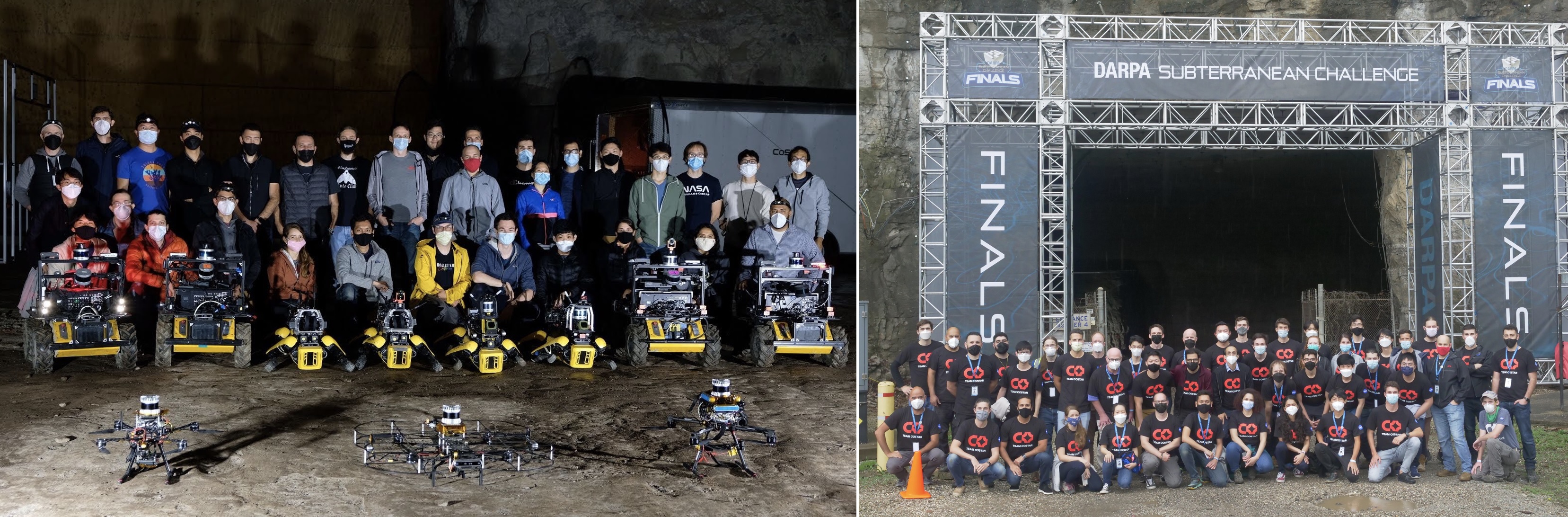}
    \caption{CoSTAR team members. Left: inside Kentucky Underground with the robot team. Right: outside the competition venue. }
    \label{fig:field:team}
\end{figure}

The scale of the field deployment enabled a large team to simultaneously test and develop every major subsystem, meaning we often had many people in the field at one time, for example in Fig.~\ref{fig:field:team}-left.

\subsection{SubT Challenge Final Event}





\begin{figure}[t]
    \centering
    \includegraphics[width=\textwidth,trim=1.0cm 6.5cm 1.0cm 2.0cm]{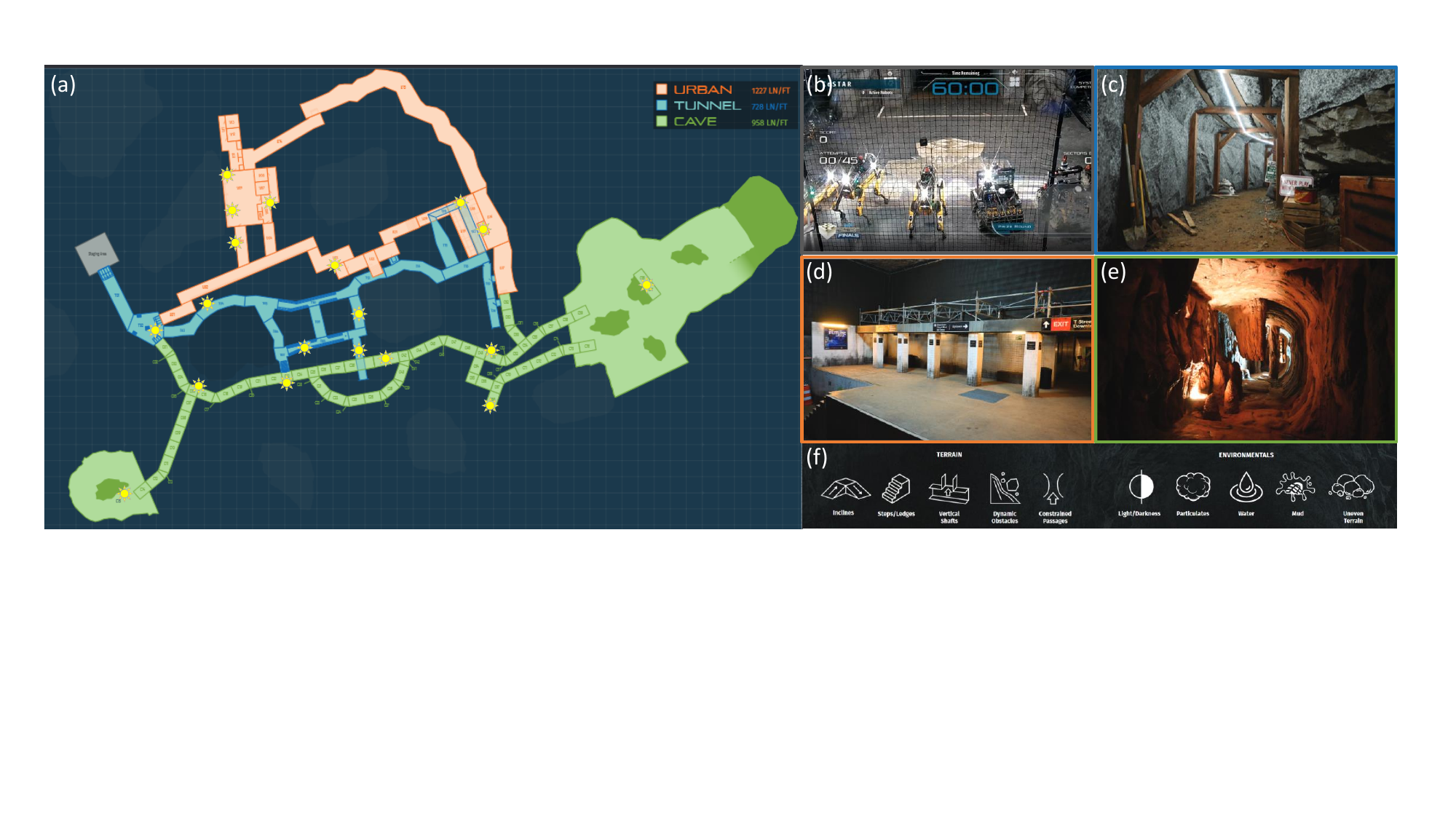}
    \caption{SubT Challenge final course configuration. a) The course map which consists tunnel, urban, and cave environment. The location of the \textit{artifacts} are annotated in yellow. b) Team CoSTAR's robots on the staging area (located on the top left of the map). c-e) Photos from different part of the course: tunnel, urban, and cave environment. f) Different type of terrain and environmental challenges in the course. The base map and photos are provided by DARPA. The annotations are added by the authors. }
    \label{fig:final_course}
\end{figure}

\ph{Competition Overview}  The competition in mixed environment (combined cave-mine-subway) took place in the Louisville Mega Cavern, Kentucky in September 2021. The teams participated in two 30-minute runs (Preliminary Run) and one 1-hour run (mixed Run). All runs share the same course carefully designed to combine various mobility challenges prevalent in subterranean environments (Fig.~\ref{fig:final_course}). The course consists of 121 sectors that possesses Tunnel, Urban, and Cave features, and spans around 900 m in length. The DARPA Final Course Callouts \cite{ackerman2022robots} describes the terrains, environmental components, and the difficulty rating on every sector.

\ph{Team CoSTAR Performance Statistics}
In the Final Event, Team CoSTAR deployed 8 robots in total: three wheeled robots (Clearpath Husky), three legged robots (Boston Dynamics Spot), and two drones (custom quad-copters). \autoref{tab:event_stats} summarizes Team CoSTAR's performance in Circuit and Final Events over the three-year effort. Nominally, we deploy a subset of robots depending on the complexity and challenges of the unseen course elements. During the Final Event, we adapted our deployment strategy to rely more on legged platforms to overcome narrow and cluttered environments with many small steps. The robot team traversed a combined distance of 1--2 km during each run. Note that the Preliminary Runs in the Final Event is only 30 minutes (half of typical runs), meaning that the driving efficiency was 1.5--2 times higher than the earlier circuit events. During the Prize Run, Team CoSTAR experienced various deployment failures, resulting in lower distance coverage and scoring.

\begin{table}[t]
    \centering
    \caption{Team CoSTAR's performance statistics for Circuit and Final Events}
    \label{tab:event_stats}
    \begin{tabular}{|l|p{1.3cm}p{1.3cm}|p{1.3cm}p{1.3cm}|p{2.6cm}|p{1.4cm}p{1.4cm}p{1.4cm}|}
        \hline
         & \multicolumn{2}{l|}{\textbf{Tunnel}} & \multicolumn{2}{l|}{\textbf{Urban}} & \textbf{Cave} & \multicolumn{3}{l|}{\textbf{Mixed Env}} \\
        Course & SR & EX & Alpha & Beta & Lavabeds & Pre-1 & Pre-2 & Prize \\
        \hline
        Num Robots & 4 & 4 & 4 & 4 & 1 & 7 & 5 & 7 \\
        Total Distance & 2.57 km & 2.54 km & 2.67 km & 2.32 km & 400 m & 2.29 km & 1.63 km & 1.05 km \\
        Score & 7 (35\%) & 4 (20\%) & 7 (35\%) & 9 (45\%) & 5 (100\%) & 1 (5\%) & 11 (55\%) & 13 (33\%) \\
        Rank & \multicolumn{2}{l|}{2nd} & \multicolumn{2}{l|}{1st} & Self-organized & 7th & 1st & 5th \\
        \hline
    \end{tabular}
\end{table}

\begin{figure}[t]
    \centering
    \includegraphics[width=\textwidth]{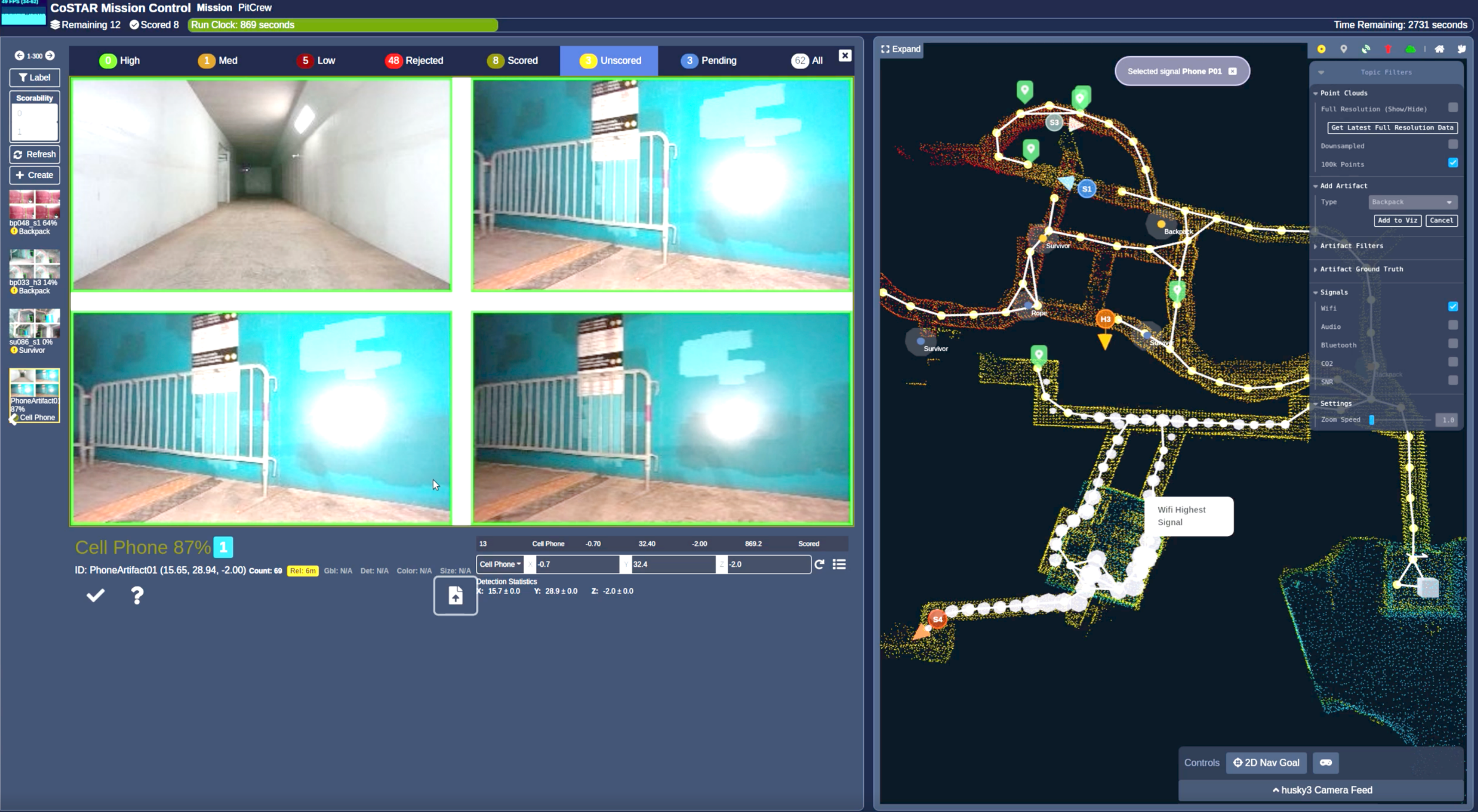}
    \caption{Live operation view during Preliminary Run. Left: The images taken around the area where strong WiFi signals are observed (a clue for Cell Phone artifacts). Right: Pointcloud/topological map built by the robot team with signal strength overlaid.}
    \label{fig:pre2_live_view}
\end{figure}

\ph{Preliminary Run 2 Result}
During the mixed environment event, Preliminary Run 2 was the only run that our robot team was deployed nominally. We will discuss the Preliminary Run 1 and the Prize Run, further below. Here we discuss the Preliminary Run 2 for the quantitative results. In Run 2, We deployed three Spots, one Husky, and one drone. The robot team traveled 1.63 km and scored 11 points (55\% of all artifacts) in 30 minutes. Fig.~\ref{fig:pre2_live_view} shows the live operation view used by the base station supervisor during the run. The left pane contains the artifact information. Artifact reports are sorted and categorized based on the scorability so that the supervisor can easily access important information under the high operational pressure. The right pane shows the 3D map constructed by the robot team. The geometry of the map is represented in two ways: pointcloud map and a topology map (IRM). Additional information is overlaid onto the map, such as WiFi signal strength (white bubbles) and communication checkpoints (green markers). At this moment, the supervisor is focusing on reviewing artifact reports, while the robots are exploring the environment on their own, in a distributed and autonomous mode, (Spot 4 is in the Urban area, Husky 3 is in the Tunnel area, and both Spots 1 and 3 are in the Cave area). Spot 2 has not been deployed yet. Spot 4 went out of communication after 600 s into the run, and fell 80 s after that with 22 artifact messages still in its communication buffer. Later, Spot 3 approached the fallen robot to retrieve the data, and was able to transfer 10 messages to the base station. \autoref{tab:event_robot_result} reports the breakdown of individual robot performance. 

\ph{Issue Breakdown in other runs}
Due to the strict COVID-19 restrictions in 2020 and 2021, and not having access to our laboratory and robots, our robot team suffered from a few deployment issues in the Final Event caused by insufficient hardware integration. This led to deployment issues in Preliminary Run 1 and the Prize Run. In the Preliminary Run 1, the artifact scoring system was disabled due to the camera hardware failure. Hence, while robots were successfully exploring the environment, no artifacts were being recorded. In the final round, our robots' internal locomotion system (not the autonomy payload) experienced unexpected device failures, preventing the robot team to be deployed nominally.



\begin{table}[t]
    \centering
    \caption{Individual robot performance during Preliminary Run 2 of Final Event}
    \label{tab:event_robot_result}
    \begin{tabular}{|l|p{1.4cm}p{2.7cm}|}
        \hline
        \textbf{Robot} & \textbf{Dist.} & \textbf{Cause} \\
        \hline
        Spot 1 & 533 m & end-of-mission \\
        Spot 3 & 584 m & end-of-mission \\
        Spot 4 & 278 m & Fall \\
        Husky 1 & 0 & not deployed \\
        Husky 2 & 0 & not deployed \\
        Husky 3 & 180 m & Stuck \\
        Drone 1 & 60 m & Stuck \\
        Drone 2 & 0 & not deployed \\
        \hline
    \end{tabular}
\end{table}


%% file: sections/12.conclusions.tex
\section{Conclusion} \label{sec:discussion}
In this paper, we provide the extensions to various modules of the original NeBula autonomy solution \cite{agha2021nebula} developed by the TEAM CoSTAR (Collaborative SubTerranean Autonomous Robots), participating in the the DARPA Subterranean Challenge. These extensions aim at increasing the size of the environment NeBula system can handle, as well as increasing the number of robots that NeBula can effectively manage in the environment. We discussed these extensions and improvements to various modules, including: (i) large-scale geometric and semantic environment mapping; (ii) localization system; (iii) terrain traversability analysis; (iv) large-scale global planning; (v) multi-robot networking; (vi) communication-aware mission planning; and (vii) multi-modal ground-aerial exploration solutions. We presented these solutions and their deployment results in the context of multi-robot deployment in large underground environments, specifically in large-scale limestone mine exploration scenarios and DARPA Subterranean challenge. The presented NeBula solution has led to 2nd, 1st, and 5th place in the three phases of the DARPA Subterranean Challenge. We believe these competitions and the robotic solutions from various teams, including the presented NeBula autonomy solution in this paper, will pave the way to more reliable deployment and more frequent adoption of multi-robot systems in challenging environments, ranging from terrestrial search-and-rescue to planetary surface exploration. 

%% file: sections/A.acronym_glossary.tex
\section{Glossary: Acronyms} \label{sec:acronyms}
\begin{acronym}
\setlength{\parskip}{0ex}
\setlength{\itemsep}{0ex}
\acro{ACHORD}{Autonomous and Collaborative High-bandwidth} \acro{BPMN}{Process Model and Notation}
\acro{CHORD}{Collaborative High-bandwidth Operations with Radio Droppables}
\acro{DDS}{Data Distribution Service}
\acro{JPLMM}{Jet Propulsion Lab Multi-Master}
\acro{MIMO}{Multi Input Multi Output}
\acro{NeBula}{Networked Belief-aware Perceptual Autonomy}
\acro{NSP}{NeBula Sensor Package}
\acro{NCC}{NeBula Computing Core}
\acro{NPDB}{NeBula Power and Diagnostics Board}
\acro{NCDS}{NeBula Communications Deployment System}
\acro{PropEM-L}{Propagation Environment Modelling and Learning}
\acro{QoS}{Quality of Service}
\acro{ROS}{Robot Operating System}
\acro{SNR}{Signal to Noise Ratio}
\acro{TCP}{Transmission Control Protocol}
\acro{UDP}{User Datagram Protocol}
\acro{HeRO}{Heterogeneous Robust Odometry}
\acro{LIO}{LiDAR-Inertial Odometry} 
\acro{VIO}{Visual-Inertial Odometry}
\acro{TIO}{Thermal-Inertial Odometry}
\acro{KIO}{Kinematic-Inertial Odometry} 
\acro{CIO}{Contact-Inertial Odometry}
\acro{RIO}{RaDAR-Inertial Odometry}
\acro{EKF}{Extended Kalman Filter}
\acro{AKF}{Adaptive Kalman Filter}
\acro{AMCCKF}{Adaptive Maximum Correntropy Criterion Kalman}
\acro{OF}{Optical Flow}
\acro{LAMP}{Large-scale Autonomous Mapping and Positioning}
\acro{PGO}{Pose-Graph Optimization}
\acro{ICM}{Incremental Consistency Maximization}
\acro{GNC}{Graduated Non-Convexity}
\acro{SAC-IA}{SAmple Consensus Initial Alignment}
\acro{GICP}{Generalized Iterative Closest Points}
\acro{ATE}{Absolute Trajectory Error}
\acro{RSSI}{Returned Signal Strength Indicator}
\acro{KLIO}{Keyframe-centric Lidar Inertial Odometry}
\acro{RTH}{Returning To Home}
\acro{DLO}{Direct Lidar Odometry}
\acro{APF}{Artificial Potential Field}
\acro{NMPC}{Nonlinear Model Predictive Controller}
\acro{DPHR}{Deepest-Point Heading Regulation}
\acro{NBV}{Next-Best-View}
\acro{HFE}{Hierarchical Frontier-based Exploration}
\acro{STEP}{Stochastic Traversability Evaluation and Planning}
\acro{POMDP}{Partially Observable Markov Decision Process}
\end{acronym}